\documentclass[11pt,letterpaper]{mystyle}
\usepackage{fancyhdr}

\usepackage[utf8]{inputenc} 
\usepackage[T1]{fontenc}
\usepackage[numbers]{natbib}
\usepackage{adjustbox}
\usepackage{breakurl}    
\usepackage{hyperref}
\usepackage[edges]{forest}
\usepackage{subcaption}
\usepackage{soul}
\usepackage{multirow}
\usepackage[utf8]{inputenc} 
\usepackage{booktabs}       
\usepackage{amsfonts}       
\usepackage{nicefrac}      
\usepackage{microtype}     
\usepackage{xcolor}        
\usepackage{colortbl}
\usepackage{enumitem}
\usepackage{amssymb}
\usepackage{wrapfig}
\usepackage{courier}
\usepackage{twemojis}
\usepackage{bxcoloremoji}
\usepackage{CJKutf8}
\usepackage{tabularx}

\usepackage{array}
\usepackage{graphicx}

\newcommand{\github}{\raisebox{-1.5pt}{\includegraphics[height=1.05em]{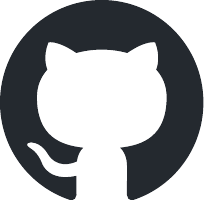}}}

\fancypagestyle{headstyle}{
    \fancyhead[L]{
    }

}

\definecolor{hidden-red}{RGB}{205, 44, 36}
\definecolor{hidden-blue}{RGB}{194,232,247}
\definecolor{hidden-orange}{RGB}{243,202,120}
\definecolor{hidden-green}{RGB}{34,139,34}
\definecolor{hidden-pink}{RGB}{255,245,247}
\definecolor{hidden-black}{RGB}{20,68,106}
\definecolor{purple}{RGB}{144,153,196}
\definecolor{yellow}{RGB}{255,228,123}
\definecolor{hidden-yellow}{RGB}{255,248,203}
\definecolor{tkcolor}{RGB}{224,223,255}
\definecolor{darkblue}{rgb}{0, 0.40, 0.75}
\newcommand{\eg}{\textit{e.g.,}}
\hypersetup{colorlinks=true, citecolor=darkblue, linkcolor=darkblue, urlcolor=darkblue}
\tcbset{
  aibox/.style={
    width=\linewidth,
    top=8pt,
    bottom=4pt,
    colback=blue!6!white,
    colframe=black,
    colbacktitle=black,
    enhanced,
    center,
    attach boxed title to top left={yshift=-0.1in,xshift=0.15in},
    boxed title style={boxrule=0pt,colframe=white,},
  }
}

\newtcolorbox{AIbox}[2][]{aibox,title=#2,#1}

\tcbset{
  takeawaybox/.style={
    width=\linewidth,
    top=8pt,
    bottom=4pt,
    colback=hidden-yellow,
    colframe=black,
    colbacktitle=black,
    enhanced,
    center,
    attach boxed title to top left={yshift=-0.1in,xshift=0.15in},
    boxed title style={boxrule=0pt,colframe=white,},
  }
}

\newtcolorbox{TakeawayBox}[2][]{takeawaybox,title=#2,#1}

\title{A Survey of Context Engineering for Large Language Models}

\author{
   Lingrui Mei$^{1,6,\dag}$ \quad Jiayu Yao$^{1,6,\dag}$ \quad Yuyao Ge$^{1,6,\dag}$ \quad Yiwei Wang$^{2}$ \quad Baolong Bi$^{1,6,\dag}$ \\
   \vspace{-5pt}
   \textbf{Yujun Cai}$^{3}$ \quad \textbf{Jiazhi Liu}$^{1}$ \quad \textbf{Mingyu Li}${^1}$ \quad \textbf{Zhong-Zhi Li}$^{6}$ \quad \textbf{Duzhen Zhang}$^{6}$\\
   \vspace{-4pt}
   \textbf{Chenlin Zhou}$^4$ \quad \textbf{Jiayi Mao}$^{5}$ \quad \textbf{Tianze Xia}$^{6}$   \quad \textbf{Jiafeng Guo}$^{1,6,\dag}$ \quad \textbf{Shenghua Liu}$^{1,6,\dag, \coloremojicode{2709}}$  \\
\normalfont{$^1$ Institute of Computing Technology, Chinese Academy of Sciences,\vspace{-5pt}\\
$^2$ University of California, Merced,
$^3$ The University of Queensland, \vspace{-5pt}\\
$^4$ Peking University,
$^5$ Tsinghua University, \vspace{-5pt}\\
$^6$ University of Chinese Academy of Sciences
}}

\begin{document}

\begin{abstract}
  \vspace{5mm}
  \textbf{\large Abstract:}
  The performance of Large Language Models (LLMs) is fundamentally determined by the contextual information provided during inference. This survey introduces \textbf{Context Engineering}, a formal discipline that transcends simple prompt design to encompass the systematic optimization of information payloads for LLMs. We present a comprehensive taxonomy decomposing Context Engineering into its foundational \textbf{Components} and the sophisticated \textbf{Implementations} that integrate them into intelligent systems.
  We first examine the foundational \textbf{Components}: (1) \textbf{Context Retrieval and Generation}, encompassing prompt-based generation and external knowledge acquisition; (2) \textbf{Context Processing}, addressing long sequence processing, self-refinement, and structured information integration; and (3) \textbf{Context Management}, covering memory hierarchies, compression, and optimization. We then explore how these components are architecturally integrated to create sophisticated \textbf{System Implementations}: (1) \textbf{Retrieval-Augmented Generation (RAG)}, including modular, agentic, and graph-enhanced architectures; (2) \textbf{Memory Systems}, enabling persistent interactions; (3) \textbf{Tool-Integrated Reasoning}, for function calling and environmental interaction; and (4) \textbf{Multi-Agent Systems}, coordinating communication and orchestration.
  Through this systematic analysis of over 1400 research papers, our survey not only establishes a technical roadmap for the field but also reveals a critical research gap: a fundamental asymmetry exists between model capabilities. While current models, augmented by advanced context engineering, demonstrate remarkable proficiency in \textit{understanding} complex contexts, they exhibit pronounced limitations in \textit{generating} equally sophisticated, long-form outputs. Addressing this gap is a defining priority for future research. Ultimately, this survey provides a unified framework for both researchers and engineers advancing context-aware AI.

  \vspace{5mm}

  $^{\dag}$ \textit{Also affiliated with: (1)Key Laboratory of Network Data Science and Technology,
  ICT, CAS; (2)State Key Laboratory of AI Safety}
  \vspace{1 mm}

  $^{\coloremojicode{2709}}$ \textit{Corresponding Author}

  \vspace{4mm}
  \vspace{4mm}

  \coloremojicode{1F4C5} \textbf{Date}: July 21, 2025


  \github{} \textbf{Code Repository}: \href{https://github.com/Meirtz/Awesome-Context-Engineering}{https://github.com/Meirtz/Awesome-Context-Engineering}

  \coloremojicode{1F4E7} \textbf{Contact}: \href{mailto:meilingrui25b@ict.ac.cn}{meilingrui25b@ict.ac.cn}, \href{mailto:liushenghua@ict.ac.cn}{liushenghua@ict.ac.cn}

\end{abstract}
\maketitle

\vspace{3mm}
\pagestyle{headstyle}
\thispagestyle{empty}
\newpage
\tableofcontents

\section{Introduction}
\label{sec:introduction}

The advent of LLMs has marked a paradigm shift in artificial intelligence, demonstrating unprecedented capabilities in natural language understanding, generation, and reasoning \citep{brown2020language, vaswani2017attention, huang2023advancing}. However, the performance and efficacy of these models are fundamentally governed by the \textit{context} they receive. This context—ranging from simple instructional prompts to sophisticated external knowledge bases—serves as the primary mechanism through which their behavior is steered, their knowledge is augmented, and their capabilities are unleashed. As LLMs have evolved from basic instruction-following systems into the core reasoning engines of complex applications, the methods for designing and managing their informational payloads have correspondingly evolved into the formal discipline of \textbf{Context Engineering} \citep{amatriain2024prompt, ye2023prompt, velsquezhenao2023prompt}.

The landscape of context engineering has expanded at an explosive rate, resulting in a proliferation of specialized yet fragmented research domains. We conceptualize this landscape as being composed of foundational \textit{components} and their subsequent \textit{implementations}. The foundational components represent the systematic pipeline of context engineering through three critical phases: \textbf{Context Retrieval and Generation}, encompassing prompt-based generation and external knowledge acquisition \citep{amatriain2024prompt, lewis2020retrieval, baek2023knowledge}; \textbf{Context Processing}, involving long sequence processing, self-refinement mechanisms, and structured information integration \citep{dao2022flashattention, madaan2023self, jiang2023structgpt}; and \textbf{Context Management}, addressing memory hierarchies, compression techniques, and optimization strategies \citep{zhong2023memorybank, wang2024adapting, packer2023memgpt}.

These foundational components serve as the building blocks for more complex, application-oriented implementations that bridge LLMs to external realities. These systems include \textbf{Advanced Retrieval-Augmented Generation (RAG)}, which has evolved into modular and agentic architectures for dynamic knowledge injection \citep{lewis2020retrieval, gao2024modular, singh2025agentic, gao2023retrieval}; explicit \textbf{Memory Systems} that mimic human cognitive faculties for persistent information retention \citep{xing2025structured, shan2025cognitive, zhong2023memorybank}; and the entire ecosystem of \textbf{Intelligent Agent Systems}. This latter category represents the pinnacle of context engineering, where agents leverage \textbf{Function Calling} and \textbf{Tool-Integrated Reasoning} to interact with the world \citep{schick2023toolformer, qian2025toolrl, lin2024training}, and rely on sophisticated \textbf{Agent Communication} protocols and \textbf{Context Orchestration} to achieve complex goals in multi-agent configurations \citep{guo2024large, ehtesham2025survey, rizk2020conversational, chang2025sagallm}.

While each of these domains has generated substantial innovation, they are predominantly studied in isolation. This fragmented development obscures the fundamental connections between techniques and creates significant barriers for researchers seeking to understand the broader landscape and practitioners aiming to leverage these methods effectively. The field urgently requires a unified framework that systematically organizes these diverse techniques, clarifies their underlying principles, and illuminates their interdependencies.

To address this critical gap, this survey provides the first comprehensive and systematic review of Context Engineering for LLMs. Our primary contribution is a novel, structured taxonomy that classifies the multifaceted techniques used to design, manage, and optimize context. This taxonomy organizes the field into coherent categories, distinguishing between foundational \textit{Components} and their integration into sophisticated \textit{System Implementations}. Through this framework, we: (1) provide a clear and structured overview of the state-of-the-art across each domain; (2) analyze the core mechanisms, strengths, and limitations of different approaches; and (3) identify overarching challenges and chart promising directions for future research. This work serves as both a technical roadmap for navigating the complex landscape of context engineering and a foundation for fostering deeper understanding and catalyzing future innovation.

The remainder of this paper is organized as follows. After discussing related work and formally defining Context Engineering, we first examine the \textbf{Foundational Components} of the field, covering Context Retrieval and Generation, Context Processing, and Context Management. We then explore their \textbf{System Implementations}, including Retrieval-Augmented Generation, Memory Systems, Tool-Integrated Reasoning, and Multi-Agent Systems. Finally, we discuss evaluation methodologies, future research directions, and conclude the survey. Figure \ref{fig:context-engineering-taxonomy} provides a comprehensive overview of our taxonomy, illustrating the hierarchical organization of techniques and their relationships within the Context Engineering landscape.

\tikzstyle{my-box}=[
rectangle,
draw=hidden-black,
rounded corners,
text opacity=1,
minimum height=1.5em,
minimum width=5em,
inner sep=2pt,
align=center,
fill opacity=.5,
]
\tikzstyle{leaf3}=[
my-box,
minimum height=1.5em,
fill=yellow!32,
text=black,
align=left,
font=\normalsize,
inner xsep=5pt,
inner ysep=4pt,
align=left,
text width=45em,
]
\tikzstyle{leaf6}=[
my-box,
minimum height=1.5em,
fill=purple!30,
text=black,
align=left,
font=\normalsize,
inner xsep=5pt,
inner ysep=4pt,
]
\tikzstyle{leaf4}=[
my-box,
minimum height=1.5em,
fill=hidden-blue!57,
text=black,
align=left,
font=\normalsize,
inner xsep=5pt,
inner ysep=4pt,
]
\tikzstyle{leaf2}=[
my-box,
minimum height=1.5em,
fill=hidden-green!20,
text=black,
align=left,
font=\normalsize,
inner xsep=5pt,
inner ysep=4pt,
]
\tikzstyle{leaf}=[
my-box,
minimum height=1.5em,
fill=hidden-red!20,
text=black,
align=left,
font=\normalsize,
inner xsep=5pt,
inner ysep=4pt,
]
\tikzstyle{leaf5}=[
my-box,
minimum height=1.5em,
fill=darkblue!15,
text=black,
align=left,
font=\normalsize,
inner xsep=5pt,
inner ysep=4pt,
]

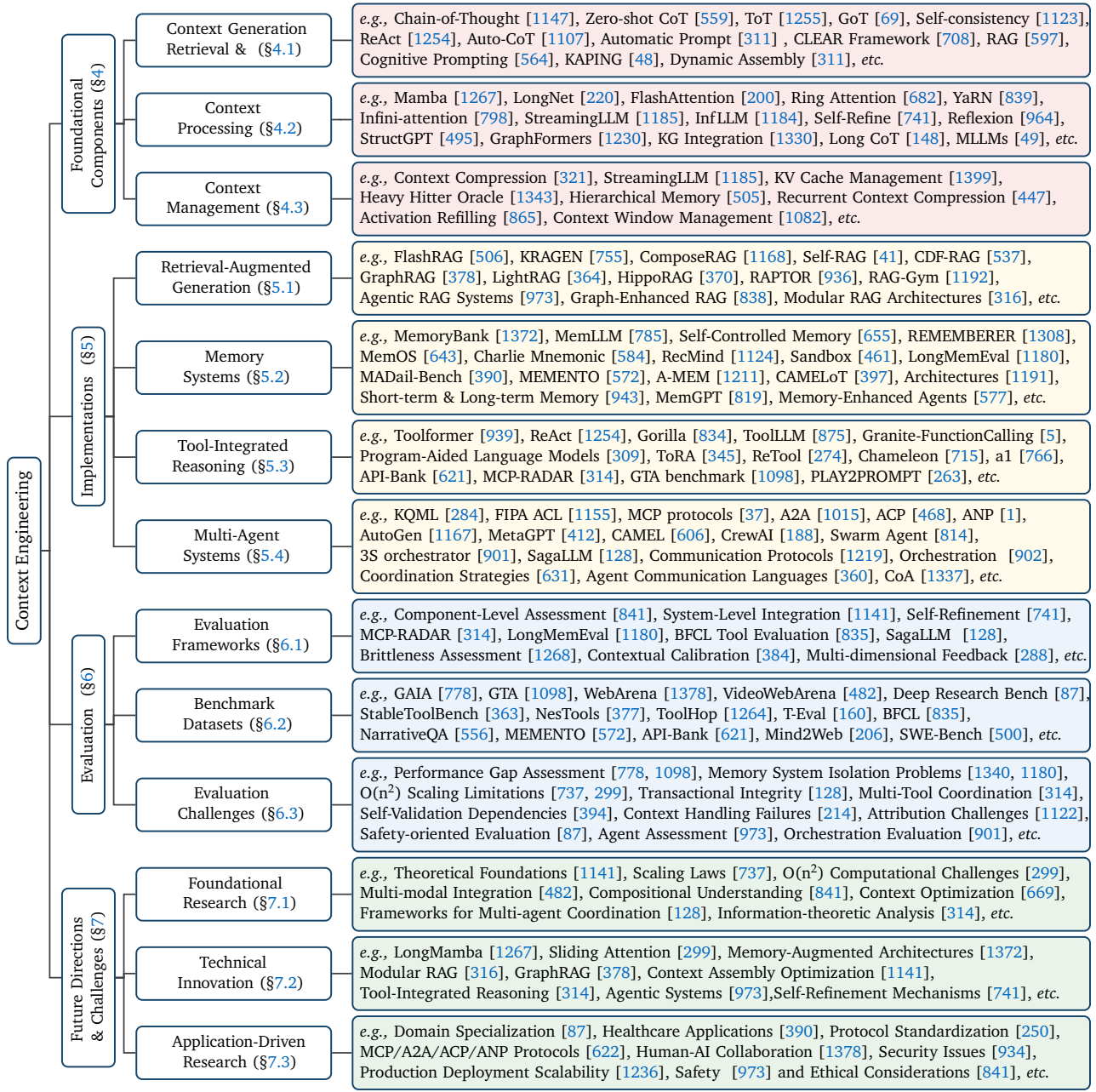
\begin{figure*}[!t]
        \vspace{-2mm}
        \centering
        \resizebox{0.96\textwidth}{!}{
                \begin{forest}
                        forked edges,
                        for tree={
                        grow=east,
                        reversed=true,
                        anchor=base west,
                        parent anchor=east,
                        child anchor=west,
                        base=left,
                        font=\large,
                        rectangle,
                        draw=hidden-black,
                        rounded corners,
                        align=left,
                        minimum width=4em,
                        edge+={darkgray, line width=1pt},
                        s sep=3pt,
                        inner xsep=2pt,
                        inner ysep=4pt,
                        line width=1.1pt,
                        ver/.style={rotate=90, child anchor=north, parent anchor=south, anchor=center},
                        },
                        where level=1{text width=10.5em,font=\normalsize,}{},
                        where level=2{text width=11.5em,font=\normalsize,}{},
                        where level=3{text width=12em,font=\normalsize,}{},
                        where level=4{text width=50em,font=\normalsize,}{},
                        [\ \ Context Engineering\ \ \ , ver
                        [\ \ \ \ \ \ \ Foundational \\ \ \ \ \ Components~(\S\ref{sec:foundational_components}),ver
                        [\ \ \ \ \  Context Generation \\ \ \ \ \ \ Retrieval \& ~(\S\ref{subsec:context_acquisition_generation})
                        [\eg ~Chain-of-Thought~\citep{wei2022chain}{,} Zero-shot CoT~\citep{kojima2022large}{,} ToT~\citep{yao2023tree}{,} GoT~\citep{besta2023graph}{,} Self-consistency~\citep{wang2022self}{,} \\ReAct~\citep{yao2022react}{,} Auto-CoT~\citep{wang2023discussion}{,} Automatic Prompt~\citep{gao2021making}{ ,} CLEAR Framework~\citep{lo2023art}{,} RAG~\citep{lewis2020retrieval}{,} \\Cognitive Prompting~\citep{kramer2024unlocking}{,} KAPING~\citep{baek2023knowledge}{,} Dynamic Assembly~\citep{gao2021making}{,} \textit{etc.}, leaf, text width=44em]
                        ]
                        [\ \ \ \ \ \ \ \  \ \ \ \ Context \\ \ \ \ \ \ \ \ Processing~(\S\ref{subsec:context_processing_transformation})
                        [\eg ~Mamba~\citep{ye2025longmamba}{,} LongNet~\citep{ding2023longnet}{,} FlashAttention~\citep{dao2022flashattention}{,} Ring Attention~\citep{liu2023ring}{,} YaRN~\citep{peng2023yarn}{,} \\Infini-attention~\citep{munkhdalai2024leave}{,} StreamingLLM~\citep{xiao2023efficient}{,} InfLLM~\citep{xiao2024infllm}{,} Self-Refine~\citep{madaan2023self}{,} Reflexion~\citep{shinn2023reflexion}{,} \\StructGPT~\citep{jiang2023structgpt}{,} GraphFormers~\citep{yang2021graphformers}{,} KG Integration~\citep{zhang2022greaselm}{,}  Long CoT~\citep{chen2025towards}{,} MLLMs~\citep{bai2024survey}{,} \textit{etc.}, leaf, text width=44em] 
                        ]
                        [\ \ \ \ \ \ \ \  \ \ \ \ Context \\ \ \ \ \  \ Management~(\S\ref{subsec:context_management})
                        [\eg ~Context Compression~\citep{ge2023context}{,} StreamingLLM~\citep{xiao2023efficient}{,} KV Cache Management~\citep{zhu2025towards}{,} \\Heavy Hitter Oracle~\citep{zhang2023heavy}{,} Hierarchical Memory~\citep{jin2024llm}{,} Recurrent Context Compression~\citep{huang2024recurrent}{,}  \\Activation Refilling~\citep{qian2024boosting}{,} Context Window Management~\citep{wang2024adapting}{,}  \textit{etc.}, leaf, text width=44em]
                        ]
                        ]
                        [ \ Implementations ~(\S\ref{sec:system_implementations}), ver
                                [\ \ \ \ Retrieval-Augmented\\ \ \ \ \ \ \ Generation~(\S\ref{subsec:advanced_rag})
                                        [\eg ~FlashRAG~\citep{jin2024flashrag}{,} KRAGEN~\citep{matsumoto2024kragen}{,} ComposeRAG~\citep{wu2025composerag}{,} Self-RAG~\citep{asai2023self}{,} CDF-RAG~\citep{khatibi2025cdf}{,} \\GraphRAG~\citep{han2025retrievalaugmentedgenerationgraphsgraphrag}{,} LightRAG~\citep{guo2024lightrag}{,} HippoRAG~\citep{gutierrez2024hipporag}{,} RAPTOR~\citep{sarthi2024raptor}{,} RAG-Gym~\citep{xiong2025rag}{,}\\Agentic RAG Systems~\citep{singh2025agentic}{,} Graph-Enhanced RAG~\citep{Peng2024GraphRG}{,} Modular RAG Architectures~\citep{gao2024modular}{,}  \textit{etc.}, leaf3, text width=44em]
                                ]
                                [\ \ \ \ \ \ \ \  \ \ \ \ Memory \\ \ \ \ \ \ \ \ \ Systems~(\S\ref{subsec:memory_systems})
                                        [\eg ~MemoryBank~\citep{zhong2023memorybank}{,} MemLLM~\citep{modarressi2024memllm}{,} Self-Controlled Memory~\citep{liang2023scm}{,} REMEMBERER~\citep{zhang2023large}{,} \\MemOS~\citep{li2025memos}{,} Charlie Mnemonic~\citep{lee2024towards}{,} RecMind~\citep{wang2024recmind}{,} Sandbox~\citep{huang2023memory}{,} LongMemEval~\citep{xia2025minerva}{,} \\MADail-Bench~\citep{he2024madial}{,} MEMENTO~\citep{kwon2025embodied}{,} A-MEM~\citep{xu2025mem}{,} CAMELoT~\citep{he2024camelot}{,} Architectures~\citep{xing2025structured}{,}  \\Short-term \& Long-term Memory~\citep{shan2025cognitive}{,} MemGPT~\citep{packer2023memgpt}{,} Memory-Enhanced Agents~\citep{langgraph_memory_2025}{,} \textit{etc.}, leaf3, text width=44em]
                                ]
                                [\ \ \ \ \ \ \ Tool-Integrated \\ \ \ \ \ \ \ Reasoning~(\S\ref{subsec:tool_augmented_systems})
                                        [\eg ~Toolformer~\citep{schick2023toolformer}{,} ReAct~\citep{yao2022react}{,} Gorilla~\citep{patil2023gorilla}{,} ToolLLM~\citep{qin2023toolllm}{,} Granite-FunctionCalling~\citep{abdelaziz2024granite}{,} \\Program-Aided Language Models~\citep{gao2022pal}{,} ToRA~\citep{gou2023tora}{,} ReTool~\citep{feng2025retool}{,} Chameleon~\citep{lu2023chameleon}{,} a1~\citep{mei2025a1steeptesttimescaling}{,} \\API-Bank~\citep{li2023api}{,} MCP-RADAR~\citep{gao2025mcp}{,} GTA benchmark~\citep{wang2024gta}{,} PLAY2PROMPT~\citep{fang2025zero}{,} \textit{etc.}, leaf3, text width=44em]
                                ]
                                [\ \ \ \ \ \ \ \ \ \ Multi-Agent \\ \ \ \ \ \ \ \ \ Systems~(\S\ref{subsec:multi_agent_systems})
                                        [\eg ~KQML~\citep{finin1994kqml}{,} FIPA ACL~\citep{wiki:acl}{,} MCP protocols~\citep{anthropic:mcp}{,} A2A~\citep{google:a2a}{,} ACP~\citep{ibm:acp}{,} ANP~\citep{anp:comm}{,} \\AutoGen~\citep{wu2023autogen}{,} MetaGPT~\citep{hong2023metagpt}{,} CAMEL~\citep{li2023camel}{,} CrewAI~\citep{crewai}{,} Swarm Agent~\citep{openai:swarm}{,} \\ 3S orchestrator~\citep{rizk2020unified}{,} SagaLLM~\citep{chang2025sagallm}{,} Communication Protocols~\citep{yan2025beyond}{,} Orchestration ~\citep{rizk2020conversational}{,} \\ Coordination Strategies~\citep{li2024survey}{,} Agent Communication Languages~\citep{guo2024large}{,} CoA~\citep{zhang2024chain}{,} \textit{etc.}, leaf3, text width=44em]
                                ]
                        ]
                        [\ \ \ \ \ Evaluation ~(\S\ref{sec:evaluation}), ver
                                [\ \ \ \ \ \ \ \ \ \ Evaluation \\ \ \ \ \ \ \ Frameworks~(\S\ref{subsec:evaluation_frameworks})
                                        [\eg ~Component-Level Assessment~\citep{peng2024survey}{,} System-Level Integration~\citep{wang2024what}{,} Self-Refinement~\citep{madaan2023self}{,}\\ MCP-RADAR~\citep{gao2025mcp}{,} LongMemEval~\citep{xia2025minerva}{,} BFCL Tool Evaluation~\citep{patil2025bfcl}{,} SagaLLM ~\citep{chang2025sagallm}{,} \\Brittleness Assessment~\citep{yehudai2025survey}{,} Contextual Calibration~\citep{hartmann2024survey}{,} Multi-dimensional Feedback~\citep{fortunato2019generalization}{,} \textit{etc.}, leaf5, text width=44em]
                                ]
                                [\ \ \ \ \ \ \ \ \ \ Benchmark \\ \ \ \ \ \ \ \ \ Datasets~(\S\ref{subsec:benchmark_datasets})
                                        [\eg ~GAIA~\citep{mialon2023gaia}{,} GTA~\citep{wang2024gta}{,} WebArena~\citep{zhou2023webarena}{,} VideoWebArena~\citep{jang2024videowebarena}{,} Deep Research Bench~\citep{bosse2025deep}{,} \\StableToolBench~\citep{guo2024stabletoolbench}{,} NesTools~\citep{han2024nestools}{,} ToolHop~\citep{ye2025toolhop}{,} T-Eval~\citep{chen2023t}{,} BFCL~\citep{patil2025bfcl}{,} \\NarrativeQA~\citep{kocisk2017narrativeqa}{,} MEMENTO~\citep{kwon2025embodied}{,} API-Bank~\citep{li2023api}{,} Mind2Web~\citep{deng2023towards}{,} SWE-Bench~\citep{jimenez2023swe}{,} \textit{etc.}, leaf5, text width=44em]
                                ]
                                [\ \ \ \ \ \ \ \ \ \ Evaluation \\ \ \ \ \ \ \ \ Challenges~(\S\ref{subsec:evaluation_challenges})
                                        [\eg ~Performance Gap Assessment~\citep{mialon2023gaia,wang2024gta}{,} Memory System Isolation Problems~\citep{zhang2024memsim,xia2025minerva}{,} \\O(n²) Scaling Limitations~\citep{ma2024megalodon,fu2025sliding}{,} Transactional Integrity~\citep{chang2025sagallm}{,} Multi-Tool Coordination~\citep{gao2025mcp}{,} \\Self-Validation Dependencies~\citep{he2025sentinelagent}{,} Context Handling Failures~\citep{deshpande2025trail}{,} Attribution Challenges~\citep{wang2024executable}{,} \\Safety-oriented Evaluation~\citep{bosse2025deep}{,} Agent Assessment~\citep{singh2025agentic}{,} Orchestration Evaluation~\citep{rizk2020unified}{,} \textit{etc.}, leaf5, text width=44em]
                                ]
                        ]
                        [\ \ \ \ Future Directions \\ \ \ \ \ \& Challenges~(\S\ref{sec:future_directions}), ver
                                [\ \ \ \ \ \ \ \ \ Foundational \\ \ \ \ \ \ \ \ \ Research~(\S\ref{subsec:foundational_challenges})
                                        [\eg ~Theoretical Foundations~\citep{wang2024what}{,} Scaling Laws~\citep{ma2024megalodon}{,} O(n²) Computational Challenges~\citep{fu2025sliding}{,} \\Multi-modal Integration~\citep{jang2024videowebarena}{,} Compositional Understanding~\citep{peng2024survey}{,} Context Optimization~\citep{lin2024training}{,}  \\Frameworks for Multi-agent Coordination~\citep{chang2025sagallm}{,} Information-theoretic Analysis~\citep{gao2025mcp}{,}  \textit{etc.}, leaf2, text width=44em]
                                ]
                                [\ \ \ \ \ \ \ \ \ \ \ Technical \\ \ \ \ \ \ \ \ Innovation~(\S\ref{subsec:technical_opportunities})
                                        [\eg ~LongMamba~\citep{ye2025longmamba}{,} Sliding Attention~\citep{fu2025sliding}{,} Memory-Augmented Architectures~\citep{zhong2023memorybank}{,} \\Modular RAG~\citep{gao2024modular}{,} GraphRAG~\citep{han2025retrievalaugmentedgenerationgraphsgraphrag}{,} Context Assembly Optimization~\citep{wang2024what}{,}  \\Tool-Integrated Reasoning~\citep{gao2025mcp}{,} Agentic Systems~\citep{singh2025agentic}{,}Self-Refinement Mechanisms~\citep{madaan2023self}{,}  \textit{etc.}, leaf2, text width=44em]
                                ]
                                [\ \ \ \ \ \ Application-Driven \\ \ \ \ \ \ \ \ \ Research~(\S\ref{subsec:application_directions})
                                        [\eg ~Domain Specialization~\citep{bosse2025deep}{,} Healthcare Applications~\citep{he2024madial}{,} Protocol Standardization~\citep{ehtesham2025survey}{,} \\MCP/A2A/ACP/ANP Protocols~\citep{li2025from}{,} Human-AI Collaboration~\citep{zhou2023webarena}{,} Security Issues~\citep{sarkar2025survey}{,} \\Production Deployment Scalability~\citep{yang2025lserve}{,} Safety ~\citep{singh2025agentic} and Ethical Considerations~\citep{peng2024survey}{,} \textit{etc.}, leaf2, text width=44em]
                                ]
                        ]
                        ]
                \end{forest}
        }
        \caption{The taxonomy of Context Engineering in Large Language Models is categorized into foundational components, system implementations, evaluation methodologies, and future directions. Each area encompasses specific techniques and frameworks that collectively advance the systematic optimization of information payloads for LLMs.}
        \label{fig:context-engineering-taxonomy}
\end{figure*}

\section{Related Work}
\label{sec:related_work}

The rapid maturation of LLMs has spurred a significant body of survey literature aiming to map its multifaceted landscape. This existing work, while valuable, has largely focused on specific vertical domains within the broader field of what we define as Context Engineering. Our survey seeks to complement these efforts by providing a horizontal, unifying taxonomy that distinguishes between foundational components and their integration into complex systems, thereby bridging these specialized areas.

\paragraph{Foundational Components} Numerous surveys have addressed the foundational \textbf{Components} of context engineering that form the core technical capabilities for effective context manipulation. The challenge of \textbf{Context Retrieval and Generation} encompasses both prompt engineering methodologies and external knowledge acquisition techniques. Surveys on prompt engineering have cataloged the vast array of techniques for guiding LLM behavior, from basic few-shot methods to advanced, structured reasoning frameworks \citep{amatriain2024prompt, fan2024survey, zhang2025survey}. External knowledge retrieval and integration techniques, particularly through knowledge graphs and structured data sources, are reviewed in works that survey representation techniques, integration paradigms, and applications in enhancing the factual grounding of LLMs \citep{ji2020survey, hu2022survey, pan2023unifying, ren2024survey}.

The domain of \textbf{Context Processing} addresses the technical challenges of handling long sequences, self-refinement mechanisms, and structured information integration. Long context processing is addressed in surveys analyzing techniques for extending context windows, optimizing attention mechanisms, and managing memory efficiently \citep{pawar2024what, li2024prompt, zeng2024cap, feng2024long}. The internal cognitive processes of LLMs are increasingly surveyed, with works on self-contextualizing techniques and self-improvement paradigms gaining prominence \citep{zhang2024survey, dong2024survey, wu2025from, shan2025cognitive}.

Finally, \textbf{Context Management} literature focuses on memory hierarchies, compression techniques, and optimization strategies that enable effective information organization and retrieval within computational constraints. While comprehensive surveys specifically dedicated to context management as a unified domain remain limited, related work on memory systems and context compression techniques provides foundational insights into these critical capabilities.

\paragraph{System Implementation} In parallel, the literature has extensively covered the \textbf{System Implementations} that integrate foundational components into sophisticated architectures addressing real-world application requirements. The domain of \textbf{RAG} has received substantial attention, with foundational surveys tracing its development and impact on mitigating hallucinations \citep{gao2023retrieval, fan2024survey, wang2024rag}. More recent work has surveyed the evolution towards modular, agentic, and graph-enhanced RAG architectures \citep{cheng2025survey, li2025survey, cao2024lego, gao2024modular, zhu2025graph}.

\textbf{Memory Systems} that enable persistent interactions and cognitive architectures have been explored through surveys focusing on memory-enhanced agents and their applications. The broader category of \textbf{LLM-based Agents} serves as a foundational area, with comprehensive overviews of autonomous agents, their architecture, planning, and methodologies \citep{wang2023survey, luo2025large, ferrag2025from, plaat2025agentic, zhao2023depth, jin2024from, yu2025survey}.

\textbf{Tool-Integrated Reasoning} encompassing function calling mechanisms and agent-environment interaction are well-documented, exploring the evolution from single-tool systems to complex orchestration frameworks \citep{lin2024training, qian2025toolrl, mialon2023augmented, qin2023toolllm}. The evolution towards \textbf{Multi-Agent Systems (MAS)} represents another focal point, with surveys detailing MAS workflows, infrastructure, communication protocols, and coordination mechanisms \citep{li2024survey, guo2024large, ehtesham2025survey, yang2025survey, aratchige2025llms, jin2025comprehensive, cui2022survey, huh2023multi}.

\paragraph{Evaluation} The critical aspect of \textbf{evaluating} these complex systems has been thoroughly reviewed, with works analyzing benchmarks and methodologies for assessing component-level and system-level capabilities and performance \citep{yehudai2025survey, hartmann2024survey, peng2024survey, gao2025mcp}. This evaluation literature spans both foundational component assessment and integrated system evaluation paradigms.

\paragraph{Our Contribution} While these surveys provide indispensable, in-depth analyses of their respective domains, they inherently present a fragmented view of the field. The connections between RAG as a form of external memory, tool use as a method for context acquisition, and prompt engineering as the language for orchestrating these components are often left implicit. Our work distinguishes itself by proposing \textit{Context Engineering} as a unifying abstraction that explicitly separates foundational components from their integration in complex implementations. By organizing these disparate fields into a single, coherent taxonomy, this survey aims to elucidate the fundamental relationships between them, providing a holistic map of how context is generated, processed, managed, and utilized to steer the next generation of intelligent systems.

\begin{figure}[h]
  \centering
  \includegraphics[width=\textwidth]{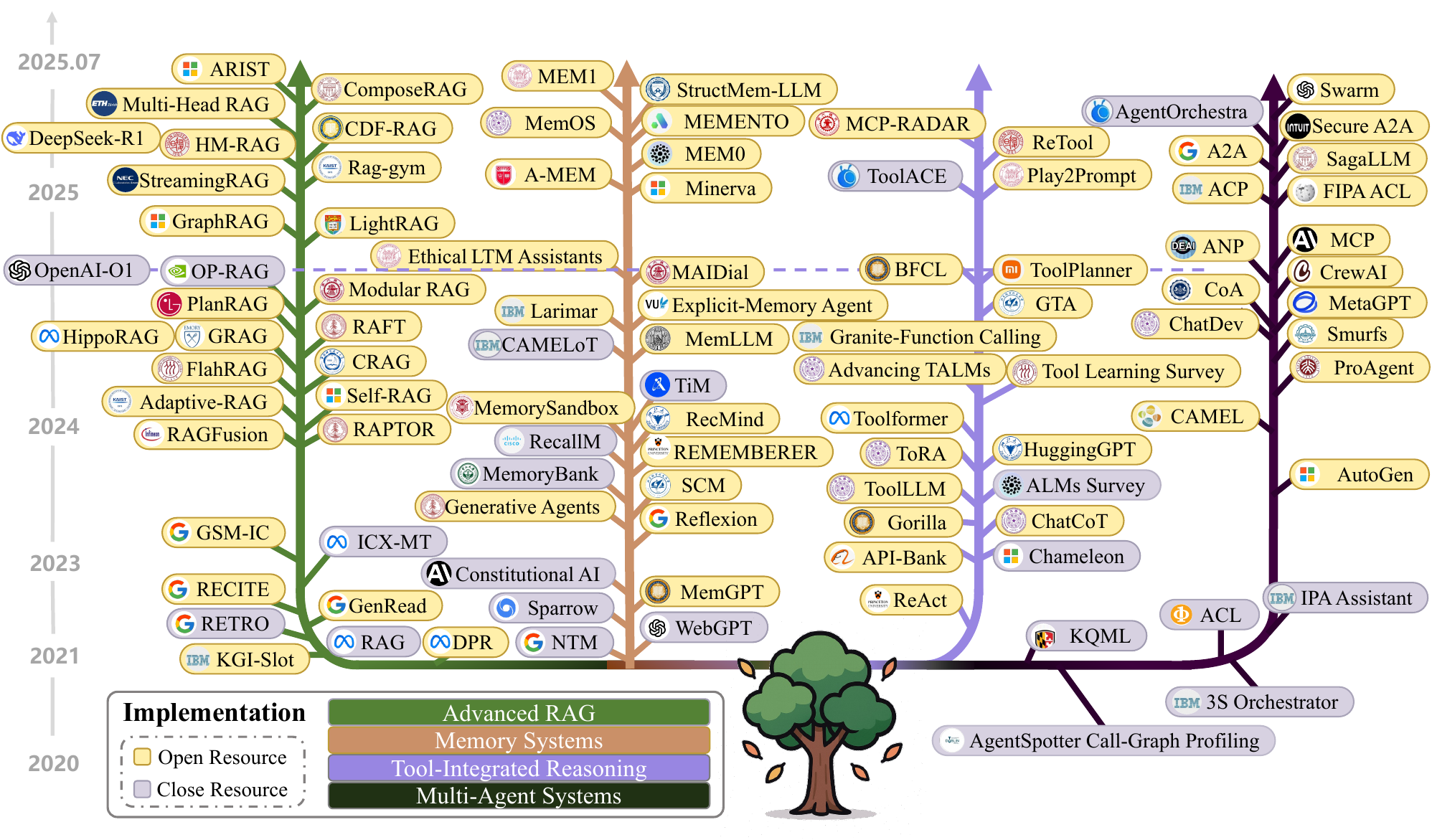}
  \caption{Context Engineering Evolution Timeline: A comprehensive visualization of the development trajectory of Context Engineering implementations from 2020 to 2025, showing the evolution from foundational RAG systems to sophisticated multi-agent architectures and tool-integrated reasoning systems.}
  \label{fig:context_engineering_timeline}
\end{figure}

\section{Why Context Engineering?}
\label{sec:why_ce}

As Large Language Models (LLMs) evolve from simple instruction-following systems into the core reasoning engines of complex, multi-faceted applications, the methods used to interact with them must also evolve. The term ``prompt engineering,'' while foundational, is no longer sufficient to capture the full scope of designing, managing, and optimizing the information payloads required by modern AI systems. These systems do not operate on a single, static string of text; they leverage a dynamic, structured, and multifaceted information stream. To address this, we introduce and formalize the discipline of \textbf{Context Engineering}.

\subsection{Definition of Context Engineering}
\label{subsec:context_engineering_definition}

To formally define Context Engineering, we begin with the standard probabilistic model of an autoregressive LLM. The model, parameterized by $\theta$, generates an output sequence $Y = (y_1, \dots, y_T)$ given an input context $C$ by maximizing the conditional probability:
\begin{equation}
    P_{\theta}(Y | C) = \prod_{t=1}^{T} P_{\theta}(y_t | y_{<t}, C)
    \label{eq:llm_generation}
\end{equation}
Historically, in the paradigm of prompt engineering, the context $C$ was treated as a monolithic, static string of text, i.e., $C = \text{prompt}$. This view is insufficient for modern systems.

Context Engineering re-conceptualizes the context $C$ as a dynamically structured set of informational components, $c_1, c_2, \dots, c_n$. These components are sourced, filtered, and formatted by a set of functions, and finally orchestrated by a high-level assembly function, $\mathcal{A}$:
\begin{equation}
    C = \mathcal{A}(c_1, c_2, \dots, c_n)
    \label{eq:context_assembly}
\end{equation}
The components $c_i$ are not arbitrary; they map directly to the core technical domains of this survey:
\begin{itemize}[noitemsep, topsep=0pt]
    \item $c_{\text{instr}}$: System instructions and rules (\textbf{Context Retrieval and Generation}, Sec. \ref{subsec:context_acquisition_generation}).
    \item $c_{\text{know}}$: External knowledge, retrieved via functions like RAG or from integrated knowledge graphs (\textbf{RAG}, Sec. \ref{subsec:advanced_rag}; \textbf{Context Processing}, Sec. \ref{subsec:context_processing_transformation}).
    \item $c_{\text{tools}}$: Definitions and signatures of available external tools (\textbf{Function Calling} \& \textbf{Tool-Integrated Reasoning}, Sec. \ref{subsec:tool_augmented_systems}).
    \item $c_{\text{mem}}$: Persistent information from prior interactions (\textbf{Memory Systems}, Sec. \ref{subsec:memory_systems}; \textbf{Context Management}, Sec. \ref{subsec:context_management}).
    \item $c_{\text{state}}$: The dynamic state of the user, world, or multi-agent system (\textbf{Multi-Agent Systems} \& \textbf{Orchestration}, Sec. \ref{subsec:multi_agent_systems}).
    \item $c_{\text{query}}$: The user's immediate request.
\end{itemize}

\paragraph{The Optimization Problem of Context Engineering.}
From this perspective, Context Engineering is the formal optimization problem of finding the ideal set of context-generating functions (which we denote collectively as $\mathcal{F} = \{\mathcal{A}, \text{Retrieve}, \text{Select}, \dots\}$) that maximizes the expected quality of the LLM's output. Given a distribution of tasks $\mathcal{T}$, the objective is:
\begin{equation}
    \mathcal{F}^* = \arg\max_{\mathcal{F}} \mathbb{E}_{\tau \sim \mathcal{T}} [\text{Reward}(P_{\theta}(Y | C_{\mathcal{F}}(\tau)), Y^*_{\tau})]
    \label{eq:ce_optimization}
\end{equation}
where $\tau$ is a specific task instance, $C_{\mathcal{F}}(\tau)$ is the context generated by the functions in $\mathcal{F}$ for that task, and $Y^*_{\tau}$ is the ground-truth or ideal output. This optimization is subject to hard constraints, most notably the model's context length limit, $|C| \leq L_{\text{max}}$.

\paragraph{Mathematical Principles and Theoretical Frameworks.}
This formalization reveals deeper mathematical principles. The assembly function $\mathcal{A}$ is a form of \textbf{Dynamic Context Orchestration}, a pipeline of formatting and concatenation operations, $\mathcal{A} = \text{Concat} \circ (\text{Format}_1, \dots, \text{Format}_n)$, where each function must be optimized for the LLM's architectural biases (e.g., attention patterns).

The retrieval of knowledge, $c_{\text{know}} = \text{Retrieve}(\dots)$, can be framed as an \textbf{Information-Theoretic Optimality} problem. The goal is to select knowledge that maximizes the mutual information with the target answer $Y^*$, given the query $c_{\text{query}}$:
\begin{equation}
    \text{Retrieve}^* = \arg\max_{\text{Retrieve}} I(Y^*; c_{\text{know}} | c_{\text{query}})
    \label{eq:info_theoretic_retrieval}
\end{equation}
This ensures that the retrieved context is not just semantically similar, but maximally informative for solving the task.

Furthermore, the entire process can be viewed through the lens of \textbf{Bayesian Context Inference}. Instead of deterministically constructing the context, we infer the optimal context posterior $P(C | c_{\text{query}}, \text{History}, \text{World})$. Using Bayes' theorem, this posterior is proportional to the likelihood of the query given the context and the prior probability of the context's relevance:
\begin{equation}
    P(C | c_{\text{query}}, \dots) \propto P(c_{\text{query}} | C) \cdot P(C | \text{History}, \text{World})
    \label{eq:bayesian_inference}
\end{equation}
The decision-theoretic objective is then to find the context $C^*$ that maximizes the expected reward over the distribution of possible answers:
\begin{equation}
    C^* = \arg\max_{C} \int P(Y | C, c_{\text{query}}) \cdot \text{Reward}(Y, Y^*) \,dY \cdot P(C | c_{\text{query}}, \dots)
    \label{eq:decision_theoretic_objective}
\end{equation}
This Bayesian formulation provides a principled way to handle uncertainty, perform adaptive retrieval by updating priors, and maintain belief states over context in multi-step reasoning tasks.

\paragraph{Comparison of Paradigms}

The formalization of Context Engineering highlights its fundamental distinctions from traditional prompt engineering. The following table summarizes the key differences.

\begin{table*}
\centering
\resizebox{\textwidth}{!}{
    \begin{tabular}{l|p{8cm}|p{12cm}}
        \toprule
\rowcolor{gray!20}
\textbf{Dimension} & \textbf{Prompt Engineering} & \textbf{Context Engineering} \\
        \midrule
        \textbf{Model} & $C = \text{prompt}$ (static string) & $C = \mathcal{A}(c_1, c_2, \dots, c_n)$ (dynamic, structured assembly) \\
        \midrule
        \textbf{Target} & $\arg\max_{\text{prompt}} P_{\theta}(Y | \text{prompt})$ & $\mathcal{F}^* = \arg\max_{\mathcal{F}} \mathbb{E}_{\tau \sim \mathcal{T}} [\text{Reward}(P_{\theta}(Y | C_{\mathcal{F}}(\tau)), Y^*_{\tau})]$ \\
        \midrule
\textbf{Complexity} & Manual or automated search over a string space. & System-level optimization of $\mathcal{F} = \{\mathcal{A}, \text{Retrieve}, \text{Select}, \dots\}$. \\
        \midrule
        \textbf{Information} & Information content is fixed within the prompt. & Aims to maximize task-relevant information under constraint $|C| \leq L_{\text{max}}$. \\
        \midrule
        \textbf{State} & Primarily stateless. & Inherently stateful, with explicit components for $c_{\text{mem}}$ and $c_{\text{state}}$. \\
        \midrule
\textbf{Scalability} & Brittleness increases with length and complexity. & Manages complexity through modular composition. \\
        \midrule
\textbf{Error Analysis} & Manual inspection and iterative refinement. & Systematic evaluation and debugging of individual context functions. \\
        \bottomrule
\end{tabular}
}
\caption{Comparison of Prompt Engineering and Context Engineering Paradigms.}
\label{tab:pe_vs_ce}
\end{table*}

In summary, Context Engineering provides the formal, systematic framework required to build, understand, and optimize the sophisticated, context-aware AI systems that are coming to define the future of the field. It shifts the focus from the ``art'' of prompt design to the ``science'' of information logistics and system optimization.

\paragraph{Context Scaling}

Context scaling encompasses two fundamental dimensions that collectively define the scope and sophistication of contextual information processing. The first dimension, \textbf{length scaling}, addresses the computational and architectural challenges of processing ultra-long sequences, extending context windows from thousands to millions of tokens while maintaining coherent understanding across extended narratives, documents, and interactions. This involves sophisticated attention mechanisms, memory management techniques, and architectural innovations that enable models to maintain contextual coherence over vastly extended input sequences.

The second, equally critical dimension is \textbf{multi-modal and structural scaling}, which expands context beyond simple text to encompass multi-dimensional, dynamic, cross-modal information structures. This includes temporal context (understanding time-dependent relationships and sequences), spatial context (interpreting location-based and geometric relationships), participant states (tracking multiple entities and their evolving conditions), intentional context (understanding goals, motivations, and implicit objectives), and cultural context (interpreting communication within specific social and cultural frameworks).

Modern context engineering must address both dimensions simultaneously, as real-world applications require models to process not only lengthy textual information but also diverse data types including structured knowledge graphs, multimodal inputs (text, images, audio, video), temporal sequences, and implicit contextual cues that humans naturally understand. This multi-dimensional approach to context scaling represents a fundamental shift from parameter scaling toward developing systems capable of understanding complex, ambiguous contexts that mirror the nuanced nature of human intelligence in facing a complex world \citep{kr36_context_scaling_2025}.

\subsection{Why Context Engineering}
\label{subsec:why_context_engineering}

\subsubsection{Current Limitations}
\label{subsubsec:current_limitations}

Large Language Models face critical technical barriers necessitating sophisticated context engineering approaches. The self-attention mechanism imposes quadratic computational and memory overhead as sequence length increases, creating substantial obstacles to processing extended contexts and significantly impacting real-world applications such as chatbots and code comprehension models \citep{tan2024lloco, song2024hierarchical}. Commercial deployment compounds these challenges through repeated context processing that introduces additional latency and token-based pricing costs \citep{tan2024lloco}.

Beyond computational constraints, LLMs demonstrate concerning reliability issues including frequent hallucinations, unfaithfulness to input context, problematic sensitivity to input variations, and responses that appear syntactically correct while lacking semantic depth or coherence \citep{shi2023trusting, yuan2025exploiting, katrix2025context}.

The prompt engineering process presents methodological challenges through approximation-driven and subjective approaches that focus narrowly on task-specific optimization while neglecting individual LLM behavior \citep{nema2025modp}. Despite these challenges, prompt engineering remains critical for effective LLM utilization through precise and contextually rich prompts that reduce ambiguity and enhance response consistency \citep{singh2024leveraging}.

\subsubsection{Performance Enhancement}
\label{subsubsec:performance_enhancement}

Context engineering delivers substantial performance improvements through techniques like retrieval-augmented generation and superposition prompting, achieving documented improvements including 18-fold enhancement in text navigation accuracy, 94\% success rates, and significant gains from careful prompt construction and automatic optimization across specialized domains \citep{feldman2024ragged, merth2024superposition, liu2024enhancing}.

Structured prompting techniques, particularly chain-of-thought approaches, enable complex reasoning through intermediate steps while enhancing element-aware summarization capabilities that integrate fine-grained details from source documents \citep{wei2022chain, matsuo2024schemato, wang2023element}. Few-shot learning implementations through carefully selected demonstration examples yield substantial performance gains, including 9.90\% improvements in BLEU-4 scores for code summarization and 175.96\% in exact match metrics for bug fixing \citep{gao2023what}.

Domain-specific context engineering proves especially valuable in specialized applications, with execution-aware debugging frameworks achieving up to 9.8\% performance improvements on code generation benchmarks and hardware design applications benefiting from specialized testbench generation and security property verification \citep{zhong2024debug, qiu2024autobench, ayalasomayajula2024lasp}. These targeted approaches bridge the gap between general-purpose model training and specialized domain requirements.

\subsubsection{Resource Optimization}
\label{subsubsec:resource_optimization}

Context engineering provides efficient alternatives to resource-intensive traditional approaches by enabling intelligent content filtering and direct knowledge transmission through carefully crafted prompts \citep{li2023unlocking, liu2024layoutcopilot}. LLMs can generate expected responses even when relevant information is deleted from input context, leveraging contextual clues and prior knowledge to optimize context length usage while maintaining response quality, particularly valuable in domains with significant data acquisition challenges \citep{li2023unlocking, liu2024layoutcopilot}.

Specialized optimization techniques further enhance efficiency gains through context awareness and responsibility tuning that significantly reduce token consumption, dynamic context optimization employing precise token-level content selection, and attention steering mechanisms for long-context inference \citep{hu2025optimizing, shen2025qwenlong, gu2024llmsteer}. These approaches maximize information density while reducing processing overhead and maintaining performance quality \citep{shen2025qwenlong, gu2024llmsteer}.

\subsubsection{Future Potential}
\label{subsubsec:future_potential}

Context engineering enables flexible adaptation mechanisms through in-context learning that allows models to adapt to new tasks without explicit retraining, with context window size directly influencing available examples for task adaptation \citep{li2024extending}. Advanced techniques integrate compression and selection mechanisms for efficient model editing while maintaining contextual coherence \citep{li2025incomes}. This adaptability proves especially valuable in low-resource scenarios, enabling effective utilization across various prompt engineering techniques including zero-shot approaches, few-shot examples, and role context without requiring domain-specific fine-tuning \citep{santos2025diverse, chang2023survey, wang2023prompt}.

Sophisticated context engineering techniques including in-context learning, chain-of-thought, tree-of-thought, and planning approaches establish foundations for nuanced language understanding and generation capabilities while optimizing retrieval and generation processes for robust, context-aware AI applications \citep{narayanan2024guard, some2025comprehensive}.

Future research directions indicate substantial potential for advancing context-sensitive applications through chain-of-thought augmentation with logit contrast mechanisms \citep{shim2024chain}, better leveraging different context types across domains, particularly in code intelligence tasks combining syntax, semantics, execution flow, and documentation \citep{wang2025towards}, and understanding optimal context utilization strategies as advanced language models continue demonstrating prompt engineering's persistent value \citep{wang2024advanced}. Evolution toward sophisticated filtering and selection mechanisms represents a critical pathway for addressing transformer architectures' scaling limitations while maintaining performance quality.

\section{Foundational Components}
\label{sec:foundational_components}

Context Engineering is built upon three fundamental components that collectively address the core challenges of information management in large language models: \textbf{Context Retrieval and Generation} sources appropriate contextual information through prompt engineering, external knowledge retrieval, and dynamic context assembly; \textbf{Context Processing} transforms and optimizes acquired information through long sequence processing, self-refinement mechanisms, and structured data integration; and \textbf{Context Management} tackles efficient organization and utilization of contextual information through addressing fundamental constraints, implementing sophisticated memory hierarchies, and developing compression techniques. These foundational components establish the theoretical and practical basis for all context engineering implementations, forming a comprehensive framework where each component addresses distinct aspects of the context engineering pipeline while maintaining synergistic relationships that enable comprehensive contextual optimization and effective context engineering strategies.

\begin{figure}[h]
  \centering
  \includegraphics[width=\textwidth]{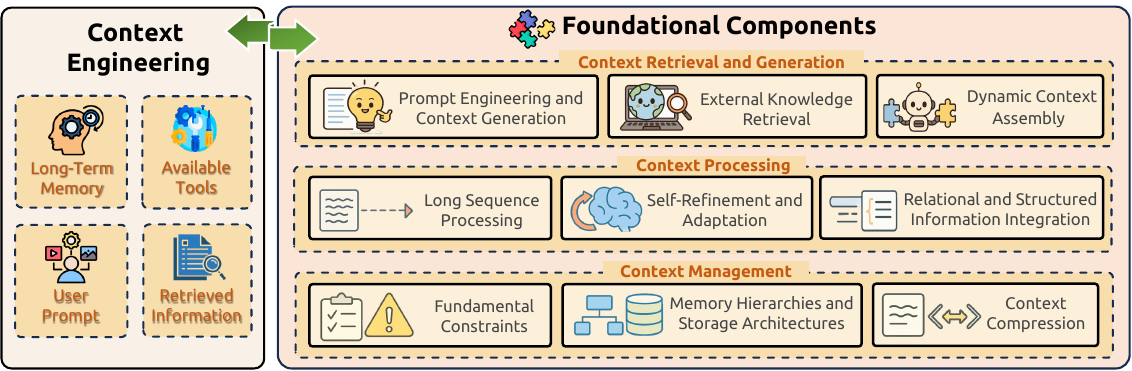}
  \caption{Context Engineering Framework: A comprehensive taxonomy of Context Engineering components including Context Retrieval and Generation, Context Processing, and Context Management, integrated into System Implementations such as RAG systems, memory architectures, tool-integrated reasoning, and multi-agent coordination mechanisms.}
  \label{fig:context_engineering_framework}
\end{figure}
\subsection{Context Retrieval and Generation}
\label{subsec:context_acquisition_generation}

Context Retrieval and Generation forms the foundational layer of context engineering, encompassing the systematic retrieval and construction of relevant information for LLMs. This component addresses the critical challenge of sourcing appropriate contextual information through three primary mechanisms: prompt-based generation that crafts effective instructions and reasoning frameworks, external knowledge retrieval that accesses dynamic information sources, and dynamic context assembly that orchestrates acquired components into coherent, task-optimized contexts.

\subsubsection{Prompt Engineering and Context Generation}
\label{subsubsec:prompt_based_generation}

Prompt engineering and context generation forms the foundational layer of context retrieval, encompassing strategic input design that combines art and science to craft effective instructions for LLMs. The CLEAR Framework—conciseness, logic, explicitness, adaptability, and reflectiveness—governs effective prompt construction, while core architecture integrates task instructions, contextual information, input data, and output indicators \citep{lo2023art, wang2024honeygpt, lamba2024role, deshmukh2025optimizing, amatriain2024prompt}.

\paragraph{Zero-Shot and Few-Shot Learning Paradigms} Zero-shot prompting enables task performance without prior examples, relying exclusively on instruction clarity and pre-trained knowledge \citep{zhong2024large, goknil2024privacy, kojima2022large, beri2024advanced, trad2024evaluating}. Few-shot prompting extends this capability by incorporating limited exemplars to guide model responses, demonstrating task execution through strategic example selection \citep{zhong2024large, hirunyasiri2023comparative, brown2020language, kirstain2021few, mosbach2023few, zhou2024teaching}. In-context learning facilitates adaptation to novel tasks without parameter updates by leveraging demonstration examples within prompts, with performance significantly influenced by example selection and ordering strategies \citep{gururajan2024aloe, brown2020language, zelikman2022star, tan2023make, salewski2023context, pond2025teaching, wei2023symbol, gu2024vocabulary, lee2025exploring}.

\paragraph{Chain-of-Thought Foundations} Chain-of-Thought (CoT) prompting decomposes complex problems into intermediate reasoning steps, mirroring human cognition \citep{wei2022chain, hirunyasiri2023comparative, goknil2024privacy, shao2023synthetic, li2023structured}. Zero-shot CoT uses trigger phrases like ``Let's think step by step,'' improving MultiArith accuracy from 17.7\% to 78.7\% \citep{kojima2022large, wang2023discussion, jade2025chatgpt, lin2023just}, with Automatic Prompt Engineer refinements yielding additional gains \citep{yang2023large, kepel2024autonomous}.

Tree-of-Thoughts (ToT) organizes reasoning as hierarchical structures with exploration, lookahead, and backtracking capabilities, increasing Game of 24 success rates from 4\% to 74\% \citep{yao2023tree, ding2023efficiency, kramer2025conceptual, li2024humaneval}. Graph-of-Thoughts (GoT) models reasoning as arbitrary graphs with thoughts as vertices and dependencies as edges, improving quality by 62\% and reducing costs by 31\% compared to ToT \citep{besta2023graph, park2025model, zhou2022least}.

\paragraph{Cognitive Architecture Integration} Cognitive prompting implements structured human-like operations including goal clarification, decomposition, filtering, abstraction, and pattern recognition, enabling systematic multi-step task resolution through deterministic, self-adaptive, and hybrid variants \citep{kramer2024unlocking, kramer2025conceptual, xu2025mindgym, wu2025effectively}. Guilford's Structure of Intellect model provides psychological foundations for categorizing cognitive operations such as pattern recognition, memory retrieval, and evaluation, enhancing reasoning clarity, coherence, and adaptability \citep{kramer2025cognitive, dai2024sap}. Advanced implementations incorporate cognitive tools as modular reasoning operations, with GPT-4.1 performance on AIME2024 increasing from 26.7\% to 43.3\% through structured cognitive operation sequences \citep{ebouky2025eliciting, tang2024fsponer}.

\begin{table*}[ht]
    \centering
    \resizebox{\textwidth}{!}{
        \begin{tabular}{l|p{20cm}}
                \toprule
                \rowcolor{gray!20}
                \textbf{Method} & \textbf{Description} \\
                \midrule
                \rowcolor{white}
                \textbf{Self-Refine} \citep{madaan2023self, sahoo2024systematic} & Enables LLMs to improve outputs through iterative feedback and refinement cycles using the same model as the generator, feedback provider, and refiner, without supervised training. \\
                \midrule
                \rowcolor{gray!10}
                \textbf{Multi-Aspect Feedback} \citep{nathani2023maf} & Integrates multiple feedback modules (frozen LMs and external tools), each focusing on specific error categories to enable more comprehensive, independent evaluation. \\
                \midrule
                \rowcolor{white}
                \textbf{N-CRITICS} \citep{mousavi2023critics} & Implements an ensemble of critics that evaluate an initial output. Compiled feedback from the generating LLM and other models guides refinement until a stopping criterion is met. \\
                \midrule
                \rowcolor{gray!10}
                \textbf{ISR-LLM} \citep{zhou2023isr} & Improves LLM-based planning by translating natural language to formal specifications, creating an initial plan, and then systematically refining it with a validator. \\
                \midrule
                \rowcolor{white}
                \textbf{SELF} \citep{lu2023self} & Teaches LLMs meta-skills (self-feedback, self-refinement) with limited examples, then has the model continuously self-evolve by generating and filtering its own training data. \\
                \midrule
                \rowcolor{gray!10}
                \textbf{ProMiSe} \citep{ramji2024self} & Addresses self-refinement in smaller LMs using principle-guided iterative refinement, combining proxy metric thresholds with few-shot refinement and rejection sampling. \\
                \midrule
                \rowcolor{white}
                \textbf{A2R} \citep{lee2024ask} & Augments LLMs through Metric-based Iterative Feedback Learning, using explicit evaluation across multiple dimensions (e.g., correctness) to generate feedback and refine outputs. \\
                \midrule
                \rowcolor{gray!10}
                \textbf{Experience Refinement} \citep{qian2024iterative} & Enables LLM agents to refine experiences during task execution by learning from recent (successive) or all previous (cumulative) experiences, prioritizing high-quality ones. \\
                \midrule
                \rowcolor{white}
                \textbf{I-SHEEP} \citep{liang2024sheep} & Allows LLMs to continuously self-align from scratch by generating, assessing, filtering, and training on high-quality synthetic datasets without external guidance. \\
                \midrule
                \rowcolor{gray!10}
                \textbf{CaP} \citep{yu2024teaching} & Uses external tools to refine chain-of-thought (CoT) responses, addressing the limitation of models that get stuck in non-correcting reasoning loops. \\
                \midrule
                \rowcolor{white}
                \textbf{Agent-R} \citep{yuan2025agent} & Enables language agents to reflect ``on the fly'' through iterative self-training, using Monte Carlo Tree Search (MCTS) to construct training data that corrects erroneous paths. \\
                \midrule
                \rowcolor{gray!10}
                \textbf{GenDiE} \citep{li2025generate} & Enhances context faithfulness with sentence-level optimization, combining generative and discriminative training to give LLMs self-generation and self-scoring capabilities. \\
                \midrule
                \rowcolor{white}
                \textbf{Self-Developing} \citep{ishibashi2024can} & Enables LLMs to autonomously discover, implement, and refine their own improvement algorithms by generating them as code, evaluating them, and using DPO to recursively improve. \\
                \midrule
                \rowcolor{gray!10}
                \textbf{SR-NLE} \citep{wang2025self} & Improves the faithfulness of post-hoc natural language explanations via an iterative critique and refinement process using self-feedback and feature attribution. \\
                \bottomrule
            \end{tabular}
    }
    \caption{Self-refinement methods in large language models and their key characteristics.}
    \label{tab:self_refinement_methods}
\end{table*}

\subsubsection{External Knowledge Retrieval}
\label{subsubsec:external_knowledge_retrieval}

External knowledge retrieval represents a critical component of context retrieval, addressing fundamental limitations of parametric knowledge through dynamic access to external information sources including databases, knowledge graphs, and document collections.

\paragraph{Retrieval-Augmented Generation Fundamentals} RAG combines parametric knowledge stored in model parameters with non-parametric information retrieved from external sources, enabling access to current, domain-specific knowledge while maintaining parameter efficiency \citep{lewis2020retrieval, gao2023retrieval, fan2024survey}. FlashRAG provides comprehensive evaluation and modular implementation of RAG systems, while frameworks like KRAGEN and ComposeRAG demonstrate advanced retrieval strategies with substantial performance improvements across diverse benchmarks \citep{jin2024flashrag, matsumoto2024kragen, wu2025composerag}.

Self-RAG introduces adaptive retrieval mechanisms where models dynamically decide when to retrieve information and generate special tokens to control retrieval timing and quality assessment \citep{asai2023self}. Advanced implementations include RAPTOR for hierarchical document processing, HippoRAG for memory-inspired retrieval architectures, and Graph-Enhanced RAG systems that leverage structured knowledge representations for improved information access \citep{sarthi2024raptor, gutierrez2024hipporag, guo2024lightrag}.

\paragraph{Knowledge Graph Integration and Structured Retrieval} Knowledge graph integration addresses structured information retrieval through frameworks like KAPING, which retrieves relevant facts based on semantic similarities and prepends them to prompts without requiring model training \citep{baek2023knowledge, liu2024dual}. KARPA provides training-free knowledge graph adaptation through pre-planning, semantic matching, and relation path reasoning, achieving state-of-the-art performance on knowledge graph question answering tasks \citep{fang2024karpa}.

Think-on-Graph enables sequential reasoning over knowledge graphs to locate relevant triples, conducting exploration to retrieve related information from external databases while generating multiple reasoning pathways \citep{sun2023think, luo2023reasoning}. StructGPT implements iterative reading-then-reasoning approaches that construct specialized functions to collect relevant evidence from structured data sources \citep{jiang2023structgpt}.

\paragraph{Agentic and Modular Retrieval Systems} Agentic RAG systems treat retrieval as dynamic operations where agents function as intelligent investigators analyzing content and cross-referencing information \citep{liang2025reasoning, cheng2025survey, singh2025agentic}. These systems incorporate sophisticated planning and reflection mechanisms requiring integration of task decomposition, multi-plan selection, and iterative refinement capabilities \citep{huang2025survey, xiong2025rag}.

Modular RAG architectures enable flexible composition of retrieval components through standardized interfaces and plug-and-play designs. Graph-Enhanced RAG systems leverage structured knowledge representations for improved information access, while Real-time RAG implementations address dynamic information requirements in streaming applications \citep{gao2024modular, zhu2025graph}.

\subsubsection{Dynamic Context Assembly}
\label{subsubsec:dynamic_context_assembly}

Dynamic context assembly represents the sophisticated orchestration of acquired information components into coherent, task-optimized contexts that maximize language model performance while respecting computational constraints.

\paragraph{Assembly Functions and Orchestration Mechanisms} The assembly function $\mathcal{A}$ encompasses template-based formatting, priority-based selection, and adaptive composition strategies that must adapt to varying task requirements, model capabilities, and resource constraints \citep{lo2023art, wang2024honeygpt, lamba2024role}. Contemporary orchestration mechanisms manage agent selection, context distribution, and interaction flow control in multi-agent systems, enabling effective cooperation through user input processing, contextual distribution, and optimal agent selection based on capability assessment \citep{rizk2020conversational, bandlamudi2023towards, choi2024malade}.

Advanced orchestration frameworks incorporate intent recognition, contextual memory maintenance, and task dispatching components for intelligent coordination across domain-specific agents. The Swarm Agent framework utilizes real-time outputs to direct tool invocations while addressing limitations in static tool registries and bespoke communication frameworks \citep{openai:swarm, fatouros2025towards, ehtesham2025survey}.

\paragraph{Multi-Component Integration Strategies} Context assembly must address cross-modal integration challenges, incorporating diverse data types including text, structured knowledge, temporal sequences, and external tool interfaces while maintaining coherent semantic relationships \citep{khan2024leveraging, yang2021graphformers, jin2022heterformer}. Verbalization techniques convert structured data including knowledge graph triples, table rows, and database records into natural language sentences, enabling seamless integration with existing language systems without architectural modifications \citep{agarwal2023kitlm, moiseev2022skill, vladika2023diversifying, agarwal2020large}.

Programming language representations of structured data, particularly Python implementations for knowledge graphs and SQL for databases, outperform traditional natural language representations in complex reasoning tasks by leveraging inherent structural properties \citep{wu2024thinking}. Multi-level structurization approaches reorganize input text into layered structures based on linguistic relationships, while structured data representations leverage existing LLMs to extract structured information and represent key elements as graphs, tables, or relational schemas \citep{liu2024enhancing, wang2024hpt, zhang2021bert}.

\paragraph{Automated Assembly Optimization} Automated prompt engineering addresses manual optimization limitations through systematic prompt generation and refinement algorithms. Automatic Prompt Engineer (APE) employs search algorithms for optimal prompt discovery, while LM-BFF introduces automated pipelines combining prompt-based fine-tuning with dynamic demonstration incorporation, achieving up to 30\% absolute improvement across NLP tasks \citep{gao2021making, hou2024enhancing, lester2021power}. Promptbreeder implements self-referential evolutionary systems where LLMs improve both task-prompts and mutation-prompts governing these improvements through natural selection analogies \citep{fernando2023promptbreeder, johnson2023fragility}.

Self-refine enables iterative output improvement through self-critique and revision across multiple iterations, with GPT-4 achieving approximately 20\% absolute performance improvement through this methodology \citep{madaan2023self, liu2024layoutcopilot}. Multi-agent collaborative frameworks simulate specialized team dynamics with agents assuming distinct roles (analysts, coders, testers), resulting in 29.9-47.1\% relative improvement in Pass@1 metrics compared to single-agent approaches \citep{hu2024nova, ye2025prompt}.

Tool integration frameworks combine Chain-of-Thought reasoning with external tool execution, automating intermediate reasoning step generation as executable programs strategically incorporating external data. LangChain provides comprehensive framework support for sequential processing chains, agent development, and web browsing capabilities, while specialized frameworks like Auto-GPT and Microsoft's AutoGen facilitate complex AI agent development through user-friendly interfaces \citep{singh2024exploring, wang2025toward, amatriain2024prompt, qin2023toolllm}.

\subsection{Context Processing}
\label{subsec:context_processing_transformation}

Context Processing focuses on transforming and optimizing acquired contextual information to maximize its utility for LLMs. This component addresses challenges in handling ultra-long sequence contexts, enables iterative self-refinement and adaptation mechanisms, and facilitates integration of multimodal, relational and structured information into coherent contextual representations.

\subsubsection{Long Context Processing}
\label{subsubsec:long_sequence_processing}

Ultra-long sequence context processing addresses fundamental computational challenges arising from transformer self-attention's O(n²) complexity, which creates significant bottlenecks as sequence lengths increase and substantially impacts real-world applications \citep{vaswani2017attention, ma2024megalodon, fu2025sliding, feng2024long, hosseini2024efficient}. Increasing Mistral-7B input from 4K to 128K tokens requires 122-fold computational increase, while memory constraints during prefilling and decoding stages create substantial resource demands, with Llama 3.1 8B requiring up to 16GB per 128K-token request \citep{tao2025saliency, yang2025lserve, hu2025raas}.

\paragraph{Architectural Innovations for Long Context} State Space Models (SSMs) maintain linear computational complexity and constant memory requirements through fixed-size hidden states, with models like Mamba offering efficient recurrent computation mechanisms that scale more effectively than traditional transformers \citep{ye2025longmamba, gu2022parameterization, gu2021efficiently}. Dilated attention approaches like LongNet employ exponentially expanding attentive fields as token distance grows, achieving linear computational complexity while maintaining logarithmic dependency between tokens, enabling processing of sequences exceeding one billion tokens \citep{ding2023longnet}.

Toeplitz Neural Networks (TNNs) model sequences with relative position encoded Toeplitz matrices, reducing space-time complexity to log-linear and enabling extrapolation from 512 training tokens to 14,000 inference tokens \citep{qin2023accelerating, qin2023toeplitz}. Linear attention mechanisms reduce complexity from O(N²) to O(N) by expressing self-attention as linear dot-products of kernel feature maps, achieving up to 4000× speedup when processing very long sequences \citep{katharopoulos2020transformers}. Alternative approaches like non-attention LLMs break quadratic barriers by employing recursive memory transformers and other architectural innovations \citep{kiruluta2025breaking}.

\paragraph{Position Interpolation and Context Extension} Position interpolation techniques enable models to process sequences beyond original context window limitations by intelligently rescaling position indices rather than extrapolating to unseen positions \citep{chen2023extending}. Neural Tangent Kernel (NTK) approaches provide mathematically grounded frameworks for context extension, with YaRN combining NTK interpolation with linear interpolation and attention distribution correction \citep{peng2023yarn, jacot2018neural, tancik2020fourier}.

LongRoPE achieves 2048K token context windows through two-stage approaches: first fine-tuning models to 256K length, then conducting positional interpolation to reach maximum context length \citep{ding2024longrope}. Position Sequence Tuning (PoSE) demonstrates impressive sequence length extensions up to 128K tokens by combining multiple positional interpolation strategies \citep{zhu2023pose}. Self-Extend techniques enable LLMs to process long contexts without fine-tuning by employing bi-level attention strategies—grouped attention and neighbor attention—to capture dependencies among distant and adjacent tokens \citep{jin2024llm}.

\paragraph{Optimization Techniques for Efficient Processing} Grouped-Query Attention (GQA) partitions query heads into groups that share key and value heads, striking a balance between multi-query attention and multi-head attention while reducing memory requirements during decoding \citep{ainslie2023gqa, zhao2024fully}. FlashAttention exploits asymmetric GPU memory hierarchy to achieve linear memory scaling instead of quadratic requirements, with FlashAttention-2 providing approximately twice the speed through reduced non-matrix multiplication operations and optimized work distribution \citep{dao2022flashattention, dao2023flashattention}.

Ring Attention with Blockwise Transformers enables handling extremely long sequences by distributing computation across multiple devices, leveraging blockwise computation while overlapping communication with attention computation \citep{liu2023ring}. Sparse attention techniques include Shifted sparse attention (S²-Attn) in LongLoRA and SinkLoRA with SF-Attn, which achieve 92\% of full attention perplexity improvement with significant computation savings \citep{zhang2024sinklora, yang2023longqlora}.

Efficient Selective Attention (ESA) proposes token-level selection of critical information through query and key vector compression into lower-dimensional representations, enabling processing of sequences up to 256K tokens \citep{wang2025unshackling}. BigBird combines local attention with global tokens that attend to entire sequences, plus random connections, enabling efficient processing of sequences up to 8× longer than previously possible \citep{zaheer2020big}.

\paragraph{Memory Management and Context Compression} Memory management strategies include Rolling Buffer Cache techniques that maintain fixed attention spans, reducing cache memory usage by approximately 8× on 32K token sequences \citep{zhao2024fully}. StreamingLLM enables processing infinitely long sequences without fine-tuning by retaining critical ``attention sink'' tokens together with recent KV cache entries, demonstrating up to 22.2× speedup over sliding window recomputation with sequences up to 4 million tokens \citep{xiao2023efficient}.

Infini-attention incorporates compressive memory into vanilla attention, combining masked local attention with long-term linear attention in single Transformer blocks, enabling processing of infinitely long inputs with bounded memory and computation \citep{munkhdalai2024leave}. Heavy Hitter Oracle (H$_2$O) presents efficient KV cache eviction policies based on observations that small token portions contribute most attention value, improving throughput by up to 29× while reducing latency by up to 1.9× \citep{zhang2023heavy}.

Context compression techniques like QwenLong-CPRS implement dynamic context optimization mechanisms enabling multi-granularity compression guided by natural language instructions \citep{shen2025qwenlong}. InfLLM stores distant contexts in additional memory units and employs efficient mechanisms to retrieve token-relevant units for attention computation, allowing models pre-trained on sequences of a few thousand tokens to effectively process sequences up to 1,024K tokens \citep{xiao2024infllm}.

\subsubsection{Contextual Self-Refinement and Adaptation}
\label{subsubsec:self_refinement_adaptation}

Self-refinement enables LLMs to improve outputs through cyclical feedback mechanisms mirroring human revision processes, leveraging self-evaluation through conversational self-interaction via prompt engineering distinct from reinforcement learning approaches \citep{madaan2023self, sahoo2024systematic, amatriain2024prompt, yan2023refining}.

\paragraph{Foundational Self-Refinement Frameworks} The Self-Refine framework uses the same model as generator, feedback provider, and refiner, demonstrating that identifying and fixing errors is often easier than producing perfect initial solutions \citep{madaan2023self, zhang2025survey, dong2024survey}. Reflexion maintains reflective text in episodic memory buffers for future decision-making through linguistic feedback \citep{shinn2023reflexion}, while structured guidance proves essential as simplistic prompting often fails to enable reliable self-correction \citep{liu2024intrinsic, lee2025explanation}.

Multi-Aspect Feedback integrates frozen language models and external tools focusing on specific error categories to enable more comprehensive, independent evaluation \citep{nathani2023maf}. The N-CRITICS framework implements ensemble-based evaluation where initial outputs are assessed by both generating LLMs and other models, with compiled feedback guiding refinement until task-specific stopping criteria are fulfilled \citep{mousavi2023critics}.

The A2R framework adopts explicit evaluation across multiple dimensions including correctness and citation quality, formulating natural language feedback for each aspect and iteratively refining outputs \citep{lee2024ask}. ISR-LLM improves LLM-based planning by translating natural language to formal specifications, creating an initial plan, and then systematically refining it with a validator \citep{zhou2023isr}.

\paragraph{Meta-Learning and Autonomous Evolution} SELF teaches LLMs meta-skills (self-feedback, self-refinement) with limited examples, then has the model continuously self-evolve by generating and filtering its own training data \citep{lu2023self}. Self-rewarding mechanisms enable models to improve autonomously through iterative self-judgment, where a single model adopts dual roles as performer and judge, maximizing rewards it assigns itself \citep{wu2024meta, yuan2024self}.

The Creator framework extends this paradigm by enabling LLMs to create and use their own tools through a four-module process encompassing creation, decision-making, execution, and recognition \citep{shen2024llm, qian2023creator}. The Self-Developing framework represents the most autonomous approach, enabling LLMs to discover, implement, and refine their own improvement algorithms through iterative cycles generating algorithmic candidates as executable code \citep{ishibashi2024can}.

In-context learning fundamentally represents a form of meta-learning where models learn optimization strategies during pre-training that generalize across diverse tasks, enabling rapid adaptation to novel challenges during inference \citep{contal2024ragsys, wu2023meta}. Meta-in-context learning demonstrates that in-context learning abilities can be recursively improved through in-context learning itself, adaptively reshaping model priors over expected tasks and modifying in-context learning strategies \citep{codaforno2023meta}.

\paragraph{Memory-Augmented Adaptation Frameworks} Memory augmentation represents a powerful approach for implementing meta-learning through frameworks like Memory of Amortized Contexts, which uses feature extraction and memory-augmentation to compress information from new documents into compact modulations stored in memory banks \citep{tack2024online}. Context-aware Meta-learned Loss Scaling addresses outdated knowledge challenges by meta-training small autoregressive models to dynamically reweight language modeling loss for each token during online fine-tuning \citep{hu2023meta}.

Decision-Pretrained Transformers demonstrate how transformers can be trained to perform in-context reinforcement learning, solving previously unseen RL problems by generalizing beyond pretraining distribution \citep{tajwar2025training, lee2023supervised}. Context-based meta-reinforcement learning methods enhance performance through direct supervision of context encoders, improving sample efficiency compared to end-to-end training approaches \citep{wang2021improving}.

\paragraph{Long Chain-of-Thought and Advanced Reasoning} Long Chain-of-Thought has emerged as a significant evolution characterized by substantially longer reasoning traces enabling thorough problem exploration, as implemented in advanced models including OpenAI-o1, DeepSeek-R1, QwQ, and Gemini 2.0 Flash Thinking \citep{chen2025towards, luo2025pruner, yan2025inftythink}. LongCoT effectiveness appears linked to context window capacity, with empirical evidence suggesting larger context windows often lead to stronger reasoning performance \citep{yang2025longer}.

Extended reasoning enables self-reflection and error correction mechanisms allowing models to identify and rectify mistakes during problem-solving processes \citep{zhang2024learn}. The effectiveness of increasing reasoning step length, even without adding new information, considerably enhances reasoning abilities across multiple datasets through test-time scaling \citep{zhao2025can}.

Optimization strategies address computational inefficiencies due to verbose reasoning traces through self-generated shorter reasoning paths via best-of-N sampling, adaptive reasoning modes including Zero-Thinking and Less-Thinking approaches, and explicit compact CoT methods reducing token usage while maintaining reasoning quality \citep{munkhbat2025self, zhao2025trade, liu2025efficient}. Auto Long-Short Reasoning enables dynamic adjustment of reasoning path length according to question complexity, helping models decide when longer chains are necessary \citep{luo2025auto}.

\subsubsection{Multimodal Context}

Multimodal Large Language Models (MLLMs) extend context engineering beyond text by integrating diverse data modalities including vision, audio, and 3D environments into unified contextual representations. This expansion introduces new challenges in modality fusion, cross-modal reasoning, and long-context processing while enabling sophisticated applications that leverage rich multimodal contextual understanding.

\paragraph{Multimodal Context Integration}
\paragraph{Foundational Techniques}
Multimodal MLLMs expand upon traditional LLMs by integrating data from diverse modalities like vision, audio, and 3D environments \citep{caffagni2024wiki, bai2024survey, shiri2024empirical}. A primary integration method converts visual inputs into discrete tokens concatenated with text tokens, conditioning the LLM's generative process on a combined representation \citep{zang2023contextual}. This is often facilitated by Visual Prompt Generators (VPGs) trained on image-caption pairs to map visual features into the LLM's embedding space \citep{li2023fine}. The dominant architectural paradigm connects specialized, external multimodal encoders—such as CLIP for vision or CLAP for audio—to the LLM backbone via alignment modules like Q-Former or simple MLPs \citep{alayrac2022flamingo, borsos2022audiolm, li2023blip, wang2024mio}, a modular design that allows for independent encoder updates without retraining the entire model \citep{li2025mindbridge}.

\paragraph{Advanced Integration Strategies}
More sophisticated approaches enable deeper modality fusion. Cross-modal attention mechanisms learn fine-grained dependencies between textual and visual tokens directly within the LLM’s embedding space, enhancing semantic understanding for tasks like image editing \citep{kumar2024multi, rombach2021high, brooks2022learning}. To manage lengthy inputs, hierarchical designs process modalities in stages to ensure scalability \citep{chen2024camml}, while the ``browse-and-concentrate'' paradigm fuses the contexts of multiple images before LLM ingestion to overcome the limitations of isolated processing \citep{wang2024browse}. Some research bypasses the adaptation of text-only LLMs, opting for unified training paradigms that jointly pre-train models on multimodal data and text corpora from the start to mitigate alignment challenges \citep{zhu2025exploring, yang2025mmada}. Other methods leverage text as a universal semantic space, using LLM in-context learning to improve generalization across diverse modality combinations \citep{tsai2024text}. For video, context integration techniques range from prompt tuning to adapter-based methods that transform video content into a sequence for reasoning \citep{wang2024contextual}. The development of these models is often constrained by the need for vast, high-quality multimodal data and significant computational resources \citep{zhang2024notellm, li2023blip, devlin2019bert}.

\paragraph{Core Challenges in Multimodal Context Processing}
\paragraph{Modality Bias and Reasoning Deficiencies}
A primary obstacle in MLLM development is modality bias, where models favor textual inputs, generating plausible but multimodally ungrounded responses by relying on learned linguistic patterns rather than integrated visual or auditory information \citep{zheng2025mllms, amara2024why, gat2021perceptual, zhang2025evaluating}. This issue is exacerbated by training methodologies; for instance, VPGs trained on simple image-captioning tasks learn to extract only salient features for captions, neglecting other visual details crucial for more complex, instruction-based tasks, which fundamentally limits deep multimodal understanding \citep{li2023fine, jin2023unified}. Consequently, MLLMs frequently struggle with fine-grained spatial or temporal reasoning, such as precise object localization or understanding detailed event sequences in videos \citep{tang2024empowering, shiri2024empirical}, particularly in complex domains like social media where interpreting the interplay of text and images to understand misinformation or sarcasm is difficult \citep{jin2024soc}. Effective multimodal reasoning requires not just comprehending each modality but also inferring their combined holistic meaning \citep{he2024poem}. Compounding these issues is our limited mechanistic understanding of MLLMs themselves; their internal workings are largely a black box, hindering the development of better architectures \citep{yu2024understanding}.

\paragraph{Advanced Contextual Capabilities and Future Directions}
\paragraph{In-Context and Long-Context Learning}
A key capability of MLLMs is in-context learning, where models adapt to new tasks from multimodal examples in the prompt without weight updates \citep{zong2024icl, zong2024icl_1, koh2023grounding}. Link-context learning (LCL) enhances this by providing demonstrations with explicit causal links, improving generalization \citep{tai2023link}. However, in-context learning is constrained by fixed context windows, as image tokens consume significant space, limiting many-shot learning \citep{huang2024multimodal}. Performance is also sensitive to input order and the relative importance of each modality varies by task \citep{tan2024order, xu2024from}. Processing long multimodal contexts, crucial for applications like video analysis, remains a major research frontier \citep{wang2024multimodal}. Innovations include adaptive hierarchical token compression for video \citep{wang2025empowering}, variable visual position encoding (V2PE) \citep{zhu2025exploring}, specialized modules like ContextQFormer for conversational memory \citep{lei2025contextqformer}, and dynamic, query-aware frame selection for video \citep{lee2025refocus}. MLLMs also show emergent communication efficiency over extended interactions, a phenomenon still under investigation \citep{hua2024talk}.

\paragraph{Emerging Applications}
The ability to process rich multimodal context is unlocking new applications. MLLMs are used for predictive reasoning, such as forecasting human activity from visual scenes \citep{zhu2023benchmarking}, and have demonstrated impressive perception and cognitive capabilities across various multimodal benchmarks \citep{fu2024video}. In VQA, context is leveraged for more precise answers, for instance, by prompting the MLLM to generate its own descriptive text context of an image \citep{zhao2023causal} or by integrating external knowledge via RAG \citep{su2024vqa, caffagni2024wiki}. Other applications include planning digital actions based on sensory inputs \citep{li2024omniactions}, enhancing surgical decision support through memory-augmented context comprehension \citep{hou2024memory}, and enabling nuanced video understanding by integrating visual information with speech and audio cues \citep{li2025watch, xu2021videoclip, acharya2018tallyqa}. Researchers have also extended MLLMs to emerging modalities like tactile information, event data, and graph structures \citep{zheng2025mllms, tang2023graphgpt, yan2023urbanclip}. The growing importance of these real-world use cases has spurred the development of comprehensive evaluation frameworks to assess contextual comprehension \citep{wang2025mucar}. These advancements enable applications previously impossible with text-only models, such as image captioning and sophisticated multimodal reasoning \citep{xia2025reimagining, liu2023visual, chen2025symbolic}.

\subsubsection{Relational and Structured Context}
\label{subsubsec:structured_information_integration}

Large language models face fundamental constraints processing relational and structured data including tables, databases, and knowledge graphs due to text-based input requirements and sequential architecture limitations \citep{jiang2023structgpt, badaro2023transformers, weber2024large}. Linearization often fails to preserve complex relationships and structural properties, with performance degrading when information is dispersed throughout contexts \citep{lee2024learning, lee2024learning_1, shao2024linearizing}.

\paragraph{Knowledge Graph Embeddings and Neural Integration} Advanced encoding strategies address structural limitations through knowledge graph embeddings that transform entities and relationships into numerical vectors, enabling efficient processing within language model architectures \citep{agarwal2023kitlm, yasunaga2021gnn, saxena2022sequence, xu2020understanding}. Graph neural networks capture complex relationships between entities, facilitating multi-hop reasoning across knowledge graph structures through specialized architectures like GraphFormers that nest GNN components alongside transformer blocks \citep{some2025comprehensive, hogan2020knowledge, yang2021graphformers, ji2020survey}.

GraphToken demonstrates substantial improvements by explicitly representing structural information, achieving up to 73 percentage points enhancement on graph reasoning tasks through parameter-efficient encoding functions \citep{perozzi2024let}. Heterformer and other hybrid GNN-LM architectures perform contextualized text encoding and heterogeneous structure encoding in unified models, addressing the computational challenges of scaling these integrated systems \citep{jin2022heterformer, ioannidis2022efficient, mavromatis2023train}.

\begin{table*}[ht]
    \centering
    \resizebox{\textwidth}{!}{
        \begin{tabular}{l|l|l|l}
            \toprule
            \rowcolor{gray!20}
            \textbf{Method} & \textbf{Approach} & \textbf{Performance} & \textbf{Key Innovation} \\
            \midrule
            \textbf{ODA} \citep{sun2024oda} & Observation-driven agent framework & 12.87\% and 8.9\% improvements & Recursive observation with action-reflection \\
            \midrule
            \textbf{RAG-KG} \citep{xu2024retrieval} & Historical issue KG construction & 77.6\% MRR, 0.32 BLEU improvement & Query parsing and sub-graph retrieval \\
            \midrule
            \textbf{KARPA} \citep{fang2024karpa} & Training-free KG adaptation & State-of-the-art KGQA performance & Pre-planning relation paths \\
            \midrule
            \textbf{Faithful Reasoning} \citep{luo2023reasoning} & Planning-retrieval-reasoning framework & N/A & LLM-KG synergy with relation paths \\
            \bottomrule
        \end{tabular}
    }
    \caption{Knowledge graph integration methods for enhanced reasoning in large language models.}
    \label{tab:kg_integration_methods}
    \end{table*}

\paragraph{Verbalization and Structured Data Representations} Verbalization techniques convert structured data including knowledge graph triples, table rows, and database records into natural language sentences, enabling seamless integration with existing language systems without architectural modifications \citep{agarwal2023kitlm, moiseev2022skill, vladika2023diversifying, agarwal2020large}. Multi-level structurization approaches reorganize input text into layered structures based on linguistic relationships, while structured data representations leverage existing LLMs to extract structured information and represent key elements as graphs, tables, or relational schemas \citep{liu2024enhancing, wang2024hpt, zhang2021bert, tay2017multi, li2023resdsql}.

Programming language representations of structured data, particularly Python implementations for knowledge graphs and SQL for databases, outperform traditional natural language representations in complex reasoning tasks by leveraging inherent structural properties \citep{wu2024thinking}. Resource-efficient approaches using structured matrix representations offer promising directions for reducing parameter counts while maintaining performance on structured data tasks \citep{grishina2025procrustesgpt}.

\paragraph{Integration Frameworks and Synergized Approaches} The integration of knowledge graphs with language models follows distinct paradigms characterized by different implementation strategies and performance trade-offs \citep{pan2023unifying, wei2022graph}. Pre-training integration methods like K-BERT inject knowledge graph triples during training to internalize factual knowledge, while inference-time approaches enable real-time knowledge access without requiring complete model retraining \citep{liu2019bert, yang2025pseudo, lu2021kelm}.

KG-enhanced LLMs incorporate structured knowledge to improve factual grounding through retrieval-based augmentation methods like KAPING, which retrieves relevant facts based on semantic similarities and prepends them to prompts without requiring model training \citep{baek2023knowledge, liu2024dual, lewis2020retrieval}. More sophisticated implementations embed KG-derived representations directly into model latent spaces through adapter modules and cross-attention mechanisms, with Text2Graph mappers providing linking between input text and KG embedding spaces \citep{chekalina2024addressing, vskrlj2025from, hu2022survey}.

Synergized approaches create unified systems where both technologies play equally important roles, addressing fundamental limitations through bidirectional reasoning driven by data and knowledge \citep{pan2023unifying, qi2024safety, wang2018improving}. GreaseLM facilitates deep interaction across all model layers, allowing language context representations to be grounded by structured world knowledge while linguistic nuances inform graph representations \citep{zhang2022greaselm}. QA-GNN implements bidirectional attention mechanisms connecting question-answering contexts and knowledge graphs through joint graph formation and mutual representation updates via graph-based message passing \citep{yasunaga2021gnn, some2025comprehensive}.

\paragraph{Applications and Performance Enhancement} Structured data integration significantly enhances LLM capabilities across multiple dimensions, with knowledge graphs providing structured information that reduces hallucinations by grounding responses in verifiable facts and improving factual accuracy through clearly defined information sources \citep{sun2024pyramid, zhao2025knowpath, dehal2025knowledge, kumar2025detecting}. Knowledge graphs enhance reasoning capabilities by providing structured entity relationships that enable complex multi-hop reasoning and logical inferences, with their rich repository of hierarchical knowledge significantly improving precision and reliability of inferences \citep{wu2024thinking, dernbach2024glam, tan2024struct}.

Real-world applications demonstrate substantial improvements across specialized domains. Healthcare systems combine structured medical knowledge with contextual understanding through Retrieval-Augmented Generation frameworks to improve disease progression modeling and clinical decision-making \citep{piya2025advancing, lee2025rag}. Scientific research platforms organize findings into structured knowledge supporting hypothesis generation and research gap identification, while business analytics systems balance rule-based precision with AI pattern recognition for more actionable insights \citep{zhang2024automated, vertsel2024hybrid}.

Question answering systems benefit from natural language interfaces over structured data sources, with integration creating more robust systems capable of handling multimodal queries and providing personalized responses that overcome static knowledge base limitations \citep{zhang2024trustuqa, wang2024application, safaei2024kglm, xu2024retrieval}. Research demonstrates that structured knowledge representations can improve summarization performance by 40\% and 14\% across public datasets compared to unstructured memory approaches, with Chain-of-Key strategies providing additional performance gains through dynamic structured memory updates \citep{hwang2024enhancing}.

\begin{table*}[ht]
    \centering
    \resizebox{\textwidth}{!}{
        \begin{tabular}{l|l|l|l|l}
            \toprule
            \rowcolor{gray!20}
            \textbf{Method} & \textbf{Data Type} & \textbf{Integration Method} & \textbf{Key Innovation} & \textbf{Task Scope} \\
            \midrule
            \textbf{K-LAMP} \citep{baek2023knowledge} & Knowledge graphs & Retrieval-based augmentation & KAPING framework & Zero-shot QA \\
            \midrule
            \textbf{Pan et al.} \citep{pan2023unifying} & Knowledge graphs & Pre-training \& inference integration & Synergized LLMs + KGs & Multi-domain reasoning \\
            \midrule
            \textbf{StructLM} \citep{zhuang2024structlm} & Tables, graphs, databases & Instruction tuning & 1.1M example dataset & 18 datasets, 8 SKG tasks \\
            \midrule
            \textbf{Shao et al.} \citep{shao2024linearizing} & Tables, databases, KGs & Linearization methods & Schema linking \& syntax prediction & Text-to-SQL tasks \\
            \bottomrule
        \end{tabular}
    }
    \caption{Representative approaches for structured data integration in large language models.}
    \label{tab:structured_integration_approaches}
    \end{table*}

\subsection{Context Management}
\label{subsec:context_management}

Context Management addresses the efficient organization, storage, and utilization of contextual information within LLMs. This component tackles fundamental constraints imposed by finite context windows, develops sophisticated memory hierarchies and storage architectures, and implements compression techniques to maximize information density while maintaining accessibility and coherence.

\subsubsection{Fundamental Constraints}

LLMs face fundamental constraints in context management stemming from finite context window sizes inherent in most architectures, which significantly reduce model efficacy on tasks requiring deep understanding of lengthy documents while imposing substantial computational demands that hinder applications requiring quick responses and high throughput \citep{wang2024adapting}. Although extending context windows enables models to handle entire documents and capture longer-range dependencies, traditional transformer architectures experience quadratic computational complexity growth as sequence length increases, making processing extremely long texts prohibitively expensive \citep{sun2023survey}. While innovative approaches like LongNet have reduced this complexity to linear, balancing window size and generalization capabilities remains challenging \citep{sun2023survey, ding2023longnet}.

Empirical evidence reveals the ``lost-in-the-middle'' phenomenon, where LLMs struggle to access information positioned in middle sections of long contexts, performing significantly better when relevant information appears at the beginning or end of inputs \citep{chang2025sagallm, liu2023lost, liang2025reasoning}. This positional bias severely impacts performance in extended chain-of-thought reasoning tasks where critical earlier results become susceptible to forgetting, with performance degrading drastically by as much as 73\% compared to performance with no prior context \citep{chang2025sagallm, wei2022chain, hankache2025evaluating}.

LLMs inherently process each interaction independently, lacking native mechanisms to maintain state across sequential exchanges and robust self-validation mechanisms, constraints stemming from fundamental limits identified in Gödel's incompleteness theorems \citep{chang2025sagallm, gdel1966formally}. This fundamental statelessness necessitates explicit management systems to maintain coherent operation sequences and ensure robust failure recovery mechanisms \citep{chang2025sagallm}. Context management faces opposing challenges of context window overflow, where models ``forget'' prior context due to exceeding window limits, and context collapse, where enlarged context windows or conversational memory cause models to fail in distinguishing between different conversational contexts \citep{sterken2025conversational}. Research demonstrates that claimed benefits of chain-of-thought prompting don't stem from genuine algorithmic learning but rather depend on problem-specific prompts, with benefits deteriorating as problem complexity increases \citep{stechly2024chain}. The computational overhead of long-context processing creates additional challenges in managing key-value caches which grow substantially with input length, creating bottlenecks in both latency and accuracy, while multi-turn and longitudinal interaction challenges further complicate context management as limited effective context hinders longitudinal knowledge accumulation and token demands of many-shot prompts constrain space available for system and user inputs while slowing inference \citep{ryu2024closer, luo2025large, he2025you}.

\subsubsection{Memory Hierarchies and Storage Architectures}

Modern LLM memory architectures employ sophisticated hierarchical designs organized into methodological approaches to overcome fixed context window limitations. OS-inspired hierarchical memory systems implement virtual memory management concepts, with MemGPT exemplifying this approach through systems that page information between limited context windows (main memory) and external storage, similar to traditional operating systems \citep{packer2023memgpt}. These architectures consist of main context containing system instructions, FIFO message queues, and writable scratchpads, alongside external context holding information accessible through explicit function calls, with memory management through function-calling capabilities enabling autonomous paging decisions \citep{pawar2024what}. PagedAttention, inspired by virtual memory and paging techniques in operating systems, manages key-value cache memory in LLMs \citep{barua2024exploring}.

Dynamic memory organizations implement innovative systems based on cognitive principles, with MemoryBank using Ebbinghaus Forgetting Curve theory to dynamically adjust memory strength according to time and significance \citep{xu2025mem, zhong2023memorybank}. ReadAgent employs episode pagination to segment content, memory gisting to create concise representations, and interactive look-up for information retrieval \citep{xu2025mem}. Compressor-retriever architectures support life-long context management by using base model forward functions to compress and retrieve context, ensuring end-to-end differentiability \citep{yang2024compressor}.

Architectural adaptations enhance model memory capabilities through internal modifications including augmented attention mechanisms, refined key-value cache mechanisms, and modified positional encodings \citep{chen2025survey, zheng2024dape}. Knowledge-organization methods structure memory into interconnected semantic networks enabling adaptive management and flexible retrieval, while retrieval mechanism-oriented approaches integrate semantic retrieval with memory forgetting mechanisms \citep{kang2025memory, zhong2023memorybank, huang2024emotional}.

System configurations balance efficiency and scalability through organizational approaches where centralized systems coordinate tasks efficiently but struggle with scalability as topics increase, leading to context overflow, while decentralized systems reduce context overflow but increase response time due to inter-agent querying \citep{helmi2025modeling}. Hybrid approaches balance shared knowledge with specialized processing for semi-autonomous operation, addressing challenges in balancing computational efficiency with contextual fidelity while mitigating memory saturation where excessive storage of past interactions leads to retrieval inefficiencies \citep{chen2025survey, helmi2025modeling}. Context Manager Components provide fundamental capabilities for snapshot creation, restoration of intermediate generation states, and overall context window management for LLMs \citep{mei2024aios}.

\subsubsection{Context Compression}

Context compression techniques enable LLMs to handle longer contexts efficiently by reducing computational and memory burden while preserving critical information. Autoencoder-based compression achieves significant context reduction through In-context Autoencoder (ICAE), which achieves 4× context compression by condensing long contexts into compact memory slots that LLMs can directly condition on, significantly enhancing models' ability to handle extended contexts with improved latency and memory usage during inference \citep{ge2023context}. Recurrent Context Compression (RCC) efficiently expands context window length within constrained storage space, addressing challenges of poor model responses when both instructions and context are compressed by implementing instruction reconstruction techniques \citep{huang2024recurrent}.

Memory-augmented approaches enhance context management through kNN-based memory caches that store key-value pairs of past inputs for later lookup, improving language modeling capabilities through retrieval-based mechanisms \citep{he2024camelot}. Contrastive learning approaches enhance memory retrieval accuracy, while side networks address memory staleness without requiring LLM fine-tuning, and consolidated representation methods dynamically update past token representations, enabling arbitrarily large context windows without being limited by fixed memory slots \citep{he2024camelot}.

Hierarchical caching systems implement sophisticated multi-layer approaches, with Activation Refilling (ACRE) employing Bi-layer KV Cache where layer-1 cache captures global information compactly and layer-2 cache provides detailed local information, dynamically refilling L1 cache with query-relevant entries from L2 cache to integrate broad understanding with specific details \citep{qian2024boosting}. Infinite-LLM addresses dynamic context length management through DistAttention for distributing attention computation across GPU clusters, liability mechanisms for borrowing memory across instances, and global planning coordination \citep{shan2025cognitive}. KCache optimizes inference by storing K Cache in high-bandwidth memory while keeping V Cache in CPU memory, selectively copying key information based on attention calculations \citep{shan2025cognitive}.

Multi-agent distributive processing represents an emerging approach using LLM-based multi-agent methods to handle massive inputs in distributed manner, addressing core bottlenecks in knowledge synchronization and reasoning processes when dealing with extensive external knowledge \citep{liu2025scaling}. Analysis of real-world key-value cache access patterns reveals high cache reusability in workloads like RAG and agents, highlighting the need for efficient distributed caching systems with optimized metadata management to reduce redundancy and improve speed \citep{zhu2025towards}. These compression techniques can be combined with other long-context modeling approaches to further enhance LLMs' capacity to process and utilize extended contexts efficiently while reducing computational overhead and preserving information integrity \citep{ge2023context}.

\begin{table*}[ht]
    \centering
    \footnotesize
    \renewcommand{\arraystretch}{1.3}
    \resizebox{\textwidth}{!}{
        \begin{tabular}{l|l|l|l|l|l}
            \toprule
            \rowcolor{gray!20}
            \textbf{Method} & \textbf{Strategy} & \textbf{Efficiency} & \textbf{Accuracy} & \textbf{Length Mgmt} & \textbf{Scalability} \\
            \midrule
            \textbf{O1-Pruner} \citep{luo2025pruner} & RL fine-tuning & N/A & +Acc, -Overhead & Auto pruning & +Efficiency \\
            \midrule
            \textbf{InftyThink} \citep{yan2025inftythink} & Iterative + summarization & Complexity reduction & +3-13\% & Iterative control & Scalable \\
            \midrule
            \textbf{Long-CoT Survey} \citep{chen2025towards} & Long CoT + reasoning & +Efficiency frameworks & +Complex domains & Deep exploration & Test-time scaling \\
            \midrule
            \textbf{PREMISE} \citep{yu2025premise} & Prompt opt + diagnostics & Gradient-inspired opt & Maintained/+Acc & -87.5\% tokens & Performance maintained \\
            \midrule
            \textbf{Prune-on-Logic} \citep{luo2025through} & Structure-aware pruning & Selective pruning & +Accuracy & Selective framework & Logic-based opt \\
            \bottomrule
        \end{tabular}
    }
    \renewcommand{\arraystretch}{1.0}
    \caption{Long-chain reasoning methods and their characteristics in large language models. O1-Pruner uses reinforcement learning-style fine-tuning to shorten reasoning chains while maintaining accuracy. InftyThink employs iterative reasoning with intermediate summarization to reduce computational complexity. Long-CoT Survey explores long chain-of-thought characteristics that enhance reasoning abilities through efficiency improvements and enhanced knowledge frameworks. PREMISE optimizes prompts with trace-level diagnostics using gradient-inspired optimization, achieving 87.5\% token reduction. Prune-on-Logic performs structure-aware pruning of logic graphs through selective removal of low-utility reasoning steps.}
    \label{tab:long_reasoning_methods}
    \end{table*}

\subsubsection{Applications}

Effective context management extends LLMs' capabilities beyond simple question-answering to enable sophisticated applications leveraging comprehensive contextual understanding across multiple domains. Document processing and analysis capabilities enable LLMs to handle entire documents or comprehend full articles rather than fragments, allowing for contextually relevant responses through comprehensive understanding of input material, particularly valuable for inherently long sequential data such as gene sequences, legal documents, and technical literature where maintaining coherence across extensive content is critical \citep{sun2023survey}.

Extended reasoning capabilities facilitated by context management techniques support complex reasoning requiring maintenance and building upon intermediate results across extended sequences. By capturing longer-range dependencies, these systems support multi-step problem solving where later reasoning depends on earlier calculations or deductions, enabling sophisticated applications in fields requiring extensive contextual awareness like complex decision support systems and scientific research assistance \citep{sun2023survey, chen2025survey}.

Collaborative and multi-agent systems benefit from effective context management in multi-turn dialogues or sequential tasks where maintaining consistent state and synchronizing internal information between collaborating models is essential \citep{chen2025survey_1}. These capabilities support applications including distributed task processing, collaborative content creation, and multi-agent problem-solving where contextual coherence across multiple interactions must be maintained \citep{chen2025survey_1}.

Enhanced conversational interfaces leverage robust context management to seamlessly handle extensive conversations without losing thread coherence, enabling more natural, persistent dialogues that closely resemble human conversations \citep{raiaan2024review}. Task-oriented LLM systems benefit from structured context management approaches, with sliding window storage implementing minimal context management systems that permanently append prompts and responses to context stores, and Retrieval-Augmented Generation systems supplementing LLMs with access to external sources of dynamic information \citep{dhamani2023tyranny, sarkar2025survey}. These capabilities support applications like personalized virtual assistants, long-term tutoring systems, and therapeutic conversational agents that maintain continuity across extended interactions \citep{raiaan2024review}.

Memory-augmented applications implement strategies enabling LLMs to persistently store, manage, and dynamically retrieve relevant contextual information, supporting applications requiring knowledge accumulation over time through building personalized user models via continuous interaction, implementing effective knowledge management across extended interactions, and supporting long-term planning scenarios depending on historical context \citep{chen2025survey}. Advanced memory frameworks like Contextually-Aware Intelligent Memory (CAIM) enhance long-term interactions by incorporating cognitive AI principles through modules that enable storage and retrieval of user-specific information while supporting contextual and time-based relevance filtering \citep{westhuer2025caim}. Memory management for LLM agents incorporates processes analogous to human memory reconsolidation, including deduplication, merging, and conflict resolution, with approaches like Reflective Memory Management combining prospective and retrospective reflection for dynamic summarization and retrieval optimization \citep{wu2025from, hatalis2024memory}. Case-based reasoning systems provide theoretical foundations for LLM agent memory through architectural components that enable cognitive integration and persistent context storage techniques that implement caching strategies for faster provisioning of necessary context \citep{hatalis2025review, hassani2019context}. The benefits extend beyond processing longer texts to fundamentally enhancing LLM interaction quality through improved comprehension, more relevant responses, and greater continuity across extended engagements, significantly expanding LLMs' utility and resolving limitations imposed by restricted context windows \citep{raiaan2024review}.

\section{System Implementations}
\label{sec:system_implementations}

Building upon the foundational components of Context Engineering, this section examines sophisticated system implementations that integrate these components into practical, intelligent architectures. These implementations represent the evolution from theoretical frameworks to deployable systems that leverage context engineering principles. We present four major categories of system implementations. \textbf{RAG} systems demonstrate external knowledge integration through modular architectures and graph-enhanced approaches. \textbf{Memory Systems} showcase persistent context management through sophisticated memory architectures enabling long-term learning. \textbf{Tool-Integrated Reasoning} transforms language models into world interactors through function calling and environment interaction. \textbf{Multi-Agent Systems} present coordinated approaches through communication protocols and orchestration mechanisms. Each implementation builds upon foundational components while addressing specific challenges in context utilization, demonstrating how theoretical principles translate into practical systems.

\subsection{Retrieval-Augmented Generation}
\label{subsec:advanced_rag}

Retrieval-Augmented Generation bridges the gap between parametric knowledge and dynamic information access by integrating external knowledge sources with language model generation. This implementation enables models to access current, domain-specific information through modular architectures, agentic frameworks, and graph-enhanced approaches that extend beyond static training data.

\begin{figure}[h]
  \centering
  \includegraphics[width=\textwidth]{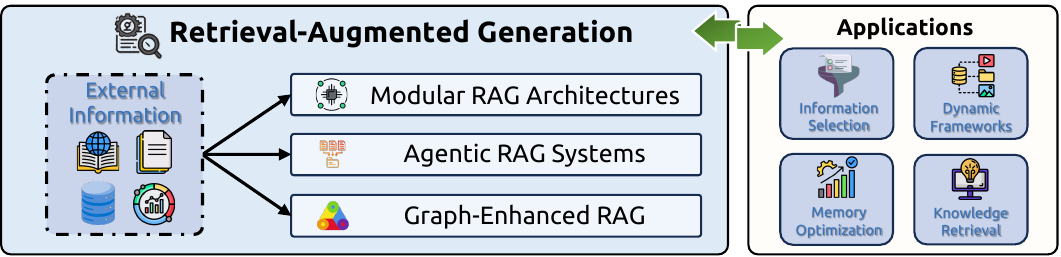}
  \caption{Retrieval-Augmented Generation Framework: Overview of RAG system architectures including Modular RAG, Agentic RAG Systems, and Graph-Enhanced RAG approaches for external context integration.}
  \label{fig:rag_framework}
\end{figure}

\subsubsection{Modular RAG Architectures}
\label{subsubsec:modular_rag}

Modular RAG shifts from linear retrieval-generation architectures toward reconfigurable frameworks with flexible component interaction \citep{gao2023retrieval, wang2024rag, lewis2020retrieval}. Unlike Naive RAG and Advanced RAG's query rewriting, Modular RAG introduces hierarchical architectures: top-level RAG stages, middle-level sub-modules, and bottom-level operational units \citep{gao2024modular, ma2023query}. This transcends linear structures through routing, scheduling, and fusion mechanisms enabling dynamic reconfiguration \citep{gao2024modular}.

The formal representation RAG = {R, G} operates through sophisticated module arrangements enabling Rewrite-Retrieve-Read models and Generate-Read approaches, incorporating adaptive search modules, RAGFusion for multi-query processing, routing modules for optimal data source selection, and hybrid retrieval strategies addressing retrieval accuracy and context relevance \citep{gao2023retrieval, jiang2023active, ru2024ragchecker, toshevska2025llm, rackauckas2024rag, brehme2025can}.

Contemporary frameworks demonstrate significant improvements in retrieval accuracy and trustworthiness \citep{zhou2024trustworthiness}. FlashRAG provides a modular toolkit with 5 core modules and 16 subcomponents enabling independent adjustment and pipeline combination \citep{jin2024flashrag}. KRAGEN enhances biomedical problem-solving by integrating knowledge graphs with vector databases, utilizing biomedical knowledge graph-optimized prompt generation to address hallucination in complex reasoning \citep{hemmat2024leveraging, matsumoto2024kragen, soman2023biomedical}. ComposeRAG implements atomic modules for Question Decomposition and Query Rewriting, incorporating self-reflection mechanisms for iterative refinement \citep{wu2025composerag}. This modularity facilitates integration with fine-tuning and reinforcement learning, enabling customization for specific applications and comprehensive toolkits supporting diverse NLP tasks \citep{gao2024modular, saberi2024context, abdallah2025rankify}.

\subsubsection{Agentic RAG Systems}

Agentic RAG embeds autonomous AI agents into the RAG pipeline, enabling dynamic, context-sensitive operations guided by continuous reasoning \citep{singh2025agentic, ferrag2025from}. These systems leverage reflection, planning, tool use, and multi-agent collaboration to manage retrieval strategies dynamically and adapt workflows to complex task requirements \citep{singh2025agentic}. RAG and agent workflows align through query rewriting corresponding to semantic comprehension, while retrieval phases correspond to planning and execution \citep{li2025survey}.

LLM-based autonomous agents extend basic language model capabilities through multimodal perception, tool utilization, and external memory integration \citep{wu2024retrieval, wang2023survey, schick2023toolformer, plaat2025agentic}. External long-term memory serves as a knowledge datastore enabling agents to incorporate and access information over extended periods \citep{wu2024retrieval, hatalis2024memory}. Unlike static approaches, Agentic RAG treats retrieval as dynamic operation where agents function as intelligent investigators analyzing content and cross-referencing information \citep{liang2025reasoning, cheng2025survey}.

Implementation paradigms encompass prompt-based methods requiring no additional training and training-based approaches optimizing models through reinforcement learning for strategic tool invocation \citep{liang2025reasoning, zhang2025process, singh2025agentic}. Advanced systems enable LLM agents to query vector databases, access SQL databases, or utilize APIs within single workflows, with methodological advances focusing on reasoning capabilities, tool integration, memory mechanisms, and instruction fine-tuning for autonomous decision-making \citep{loffredo2025agent, acharya2025agentic}.

Core capabilities include reasoning and planning components through task decomposition, multi-plan selection, and memory-augmented planning strategies enabling agents to break down complex tasks and select appropriate strategies \citep{huang2025survey, huang2025towards}. PlanRAG improves decision-making through plan-then-retrieve approaches, enabling agents to evaluate multiple information sources and optimize retrieval strategies, while SLA management frameworks address reconfigurable multi-agent architectures \citep{cheng2025survey, iannelli2024sla}. Tool utilization enables systems to employ diverse resources including search engines, calculators, and APIs, with frameworks like ReAct and Reflexion exemplifying how interleaving reasoning with actions enhances adaptability \citep{cheng2025survey, wu2024retrieval, shinn2023reflexion}. Memory mechanisms provide external long-term storage, while adaptive retrieval strategies enable autonomous analysis of complexity and context \citep{cheng2025survey, wang2025internet}.

Self-reflection and adaptation mechanisms enable Agentic RAG systems to operate in dynamic environments through iterative feedback loops refining operations based on previous interaction outcomes \citep{xiong2025rag, liu2025rag}. Advanced memory systems like MemoryBank implement update mechanisms inspired by the Ebbinghaus Forgetting Curve, enhancing agents' ability to retrieve and apply learnings from past interactions \citep{zhong2023memorybank, cheng2023lift}. CDF-RAG employs closed-loop processes combining causal graph retrieval with reinforcement learning-driven query refinement and hallucination correction \citep{khatibi2025cdf}. Self-RAG trains models that retrieve passages on demand while reflecting on retrievals and generations, using reflection tokens to control behavior during inference \citep{du2025survey, asai2023self}.

\subsubsection{Graph-Enhanced RAG}

Graph-based Retrieval-Augmented Generation shifts from document-oriented approaches toward structured knowledge representations capturing entity relationships, domain hierarchies, and semantic connections \citep{cao2024lego, zheng2023automating, guo2024lightrag, zhu2025graph}. This enables extraction of specific reasoning paths providing relevant information to language models while supporting multi-hop reasoning through structured pathway navigation \citep{cao2024lego}. Graph structures minimize context drift and hallucinations by leveraging interconnectivity for enhanced context-aware retrieval and logical coherence \citep{kamra2024enhancing, ocker2025grounded}.

Knowledge graphs serve as foundational representations encapsulating entities and interrelationships in structured formats enabling efficient querying and semantic relationship capture \citep{cheng2025survey, van2024strategist}. Graph-based knowledge representations categorize into knowledge-based GraphRAG using graphs as knowledge carriers, index-based GraphRAG employing graphs as indexing tools, and hybrid GraphRAG combining both approaches \citep{xu2025noderag}. Sophisticated implementations include GraphRAG's hierarchical indexing with community detection, PIKE's multi-level heterogeneous knowledge graphs organizing documents into three-layer hierarchies, and EMG-RAG's Editable Memory Graph architecture \citep{gao2025synergizing}.

Graph Neural Networks enhance RAG systems by addressing limitations in handling structured knowledge, with GNNs excelling at capturing entity associations and improving knowledge consistency \citep{dong2024advanced, cao2018question}. GNN-RAG implementations adopt lightweight architectures for effective knowledge graph element retrieval, improving graph structure capture before interfacing with language models \citep{zhou2025depth, cheng2025survey}. The integration process encompasses graph building through node and edge extraction, retrieval based on queries, and generation incorporating retrieved information \citep{zhou2025depth}.

Multi-hop reasoning capabilities enable graph-based systems to synthesize information across multiple connected knowledge graph nodes, facilitating complex query resolution requiring interconnected fact integration \citep{van2024strategist, cheng2025human}. These systems employ structured representations capturing semantic relationships between entities and domain hierarchies in ways that unstructured text cannot \citep{van2024strategist, cheng2025human}. Advanced frameworks like Hierarchical Lexical Graph preserve statement provenance while clustering topics for flexible retrieval and linking entities for graph-based traversal \citep{ghassel2025hierarchical}. Systems like GraphRAG, LightRAG, and derivatives implement dual-level retrieval, hierarchical indexing, and graph-enhanced strategies enabling robust multilevel reasoning \citep{xiang2025when, gao2025synergizing}.

Prominent architectures demonstrate diverse approaches to graph-enhanced retrieval, with optimization strategies showing significant improvements in retrieval effectiveness \citep{cahoon2025optimizing}. LightRAG integrates graph structures with vector representations through dual-level retrieval paradigms improving efficiency and content quality \citep{hong2025rag, luo2025hypergraphrag}. HippoRAG leverages Personalized PageRank over knowledge graphs achieving notable improvements in multi-hop question answering \citep{wang2025proprag, markovic2025optimizing, gutierrez2024hipporag}. HyperGraphRAG proposes hypergraph structured representations advancing beyond binary relations \citep{luo2025hypergraphrag}. RAPTOR provides hierarchical summary tree construction for recursive context generation, while PathRAG introduces pruning techniques for graph-based retrieval \citep{zhao2025streamlining, sarthi2024raptor, chen2025pathrag}. These structured approaches enable transparent reasoning pathways with explicit entity connections, reducing noise and improving semantic understanding while overcoming traditional RAG challenges \citep{xiang2025when, kamra2024enhancing}.

\subsubsection{Applications}

Real-time RAG systems address critical challenges in production environments where dynamic knowledge bases require continuous updates and low-latency responses \citep{zhao2024retrieval, khan2024developing}. Core challenges include efficient deployment and processing pipeline optimization, with existing frameworks lacking plug-and-play solutions necessitating system-level optimizations \citep{zhao2024retrieval}. Integration of streaming data introduces complications as traditional architectures demonstrate poor accuracy with frequently changing information and decreased efficiency as document volumes grow \citep{kang2024sakr}.

Dynamic retrieval mechanisms advance over static approaches by continuously updating strategies during generation, adjusting goals and semantic vector spaces in real-time based on generation states and identified knowledge gaps \citep{he2025context}. Current limitations in determining optimal retrieval timing and query formulation are addressed through Chain-of-Thought reasoning, iterative retrieval processes, decomposed prompting, and LLM-generated content for dynamic retrieval enabling adaptive information selection, with approaches extending to adaptive control mechanisms enhancing generation quality through reflective tags \citep{su2024dragin, khandelwal2019generalization, borgeaud2021improving, khot2022decomposed, yao2024adaptive}.

Low-latency retrieval approaches leverage graph-based methods demonstrating significant promise in speed-accuracy optimization, with dense passage retrieval techniques providing foundational improvements \citep{karpukhin2020dense}. LightRAG's dual-level retrieval system enhances information discovery while integrating graph structures with vector representations for efficient entity relationship retrieval, reducing response times while maintaining relevance \citep{guo2024lightrag}. Multi-stage retrieval pipelines optimize computational efficiency through techniques like graph-based reranking, enabling dynamic access to current information while reducing storage requirements \citep{some2025comprehensive}.

Scalability solutions incorporate distributed processing architectures with efficient data partitioning, query optimization, and fault tolerance mechanisms adapting to changing stream conditions \citep{tinati2015streaming, aniello2013adaptive}. Memory optimization through transformed heavy hitters streaming algorithms intelligently filters irrelevant documents while maintaining quality, particularly valuable for frequently changing content \citep{kang2024sakr}. Production frameworks demonstrate efficiency gains through modular RAG architectures supporting pre-retrieval processes like query expansion and post-retrieval refinements such as compression and selection, enabling fine-tuning of individual components \citep{wan2024generative}.

Incremental indexing and dynamic knowledge updates ensure systems adapt to new information without full retraining, particularly crucial in rapidly evolving domains like cybersecurity and climate finance applications \citep{paul2025llm, vaghefi2025climate}. Modern frameworks incorporate dynamic knowledge retrieval methods enabling continuous strategy adjustment based on evolving input and contextual information, enhancing interactivity and semantic understanding while increasing applicability across cross-domain integration \citep{he2025context}. Advanced agent-based approaches demonstrate sophisticated task allocation capabilities in complex environments, such as coordinated UAV operations requiring real-time decision-making, with applications extending to grounded planning for embodied agents \citep{zhang2025coordfield, song2022llm}. Dynamic Retrieval Augmented Generation frameworks like DRAGON-AI showcase specialized implementations for ontology generation, combining textual and logical components while incorporating self-memory mechanisms enabling iterative improvement \citep{toro2023dynamic}. These advances represent significant evolution toward seamlessly integrating real-time knowledge with flexible retrieval capabilities in dynamic environments.
\subsection{Memory Systems}
\label{subsec:memory_systems}

Memory Systems enable LLMs to transcend stateless interactions by implementing persistent information storage, retrieval, and utilization mechanisms. This implementation transforms models from pattern-matching processors into sophisticated agents capable of learning, adaptation, and long-term contextual understanding across extended interactions.

\begin{figure}[h]
  \centering
  \includegraphics[width=\textwidth]{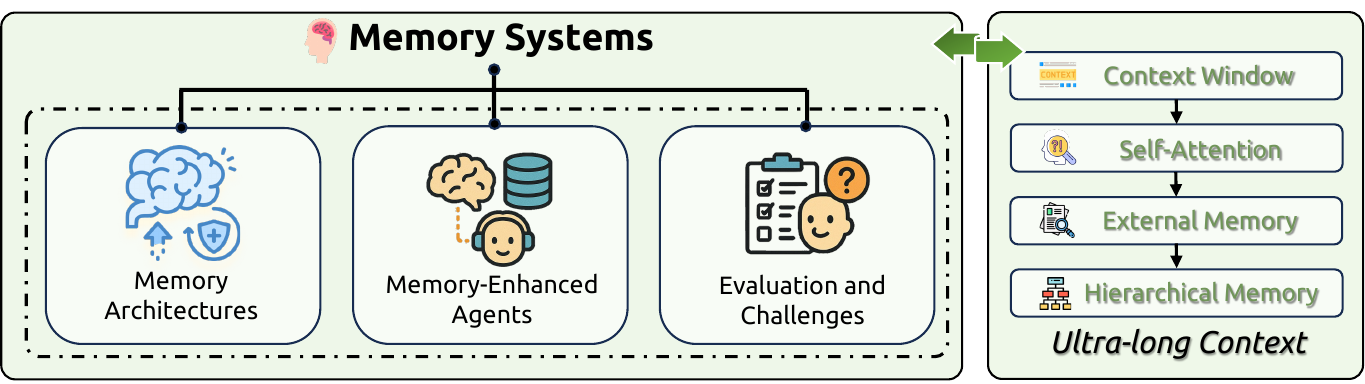}
  \caption{Memory Systems Framework: Overview of memory architectures, memory-enhanced agents, and evaluation challenges for ultra-long context processing in LLMs.}
  \label{fig:memory_framework}
\end{figure}

\subsubsection{Memory Architectures}
\label{subsubsec:memory_architectures}

Memory distinguishes sophisticated language systems from pattern-matching models, enabling information processing, storage, and utilization across natural language tasks \citep{xing2025structured, wu2025from, fusi2021memory}. LLMs face considerable memory system constraints despite breakthroughs in text generation and multi-turn conversations \citep{xing2025structured}. Neural memory mechanisms struggle with inadequate structured information storage and reliance on approximate vector similarity calculations rather than precise symbolic operations, challenging accurate storage and retrieval for multi-hop reasoning \citep{hu2023chatdb}. These limitations represent critical challenges for developing AI systems operating effectively in complex real-world applications \citep{kirkpatrick2016overcoming}.

\paragraph{Memory Classification Frameworks} LLM memory systems can be organized into multiple classification frameworks. The primary temporal classification divides memory into three categories: sensory memory (input prompts), short-term memory (immediate context processing), and long-term memory (external databases or dedicated structures) \citep{shan2025cognitive}. From a persistence perspective, short-term memory includes key-value caches and hidden states existing only within single sessions, while long-term memory encompasses text-based storage and knowledge embedded in model parameters, persisting across multiple interaction cycles \citep{shan2025cognitive, pan2025memorization}.

Implementation-based classifications identify parametric memory (knowledge encoded in model weights), ephemeral activation memory (context-limited runtime states), and plaintext memory accessed through Retrieval-Augmented Generation methods \citep{li2025memos}. Current implementations lack sophisticated lifecycle management and multi-modal integration, limiting long-term knowledge evolution. Feed-forward network layers serve as key-value tables storing memory, functioning as ``inner lexicon'' for word retrieval and creating mechanisms analogous to human associative memory \citep{kaplan2024from, geva2020transformer, geva2022transformer, meng2022locating, indiveri2015memory}. These classification schemes reflect attempts to develop LLM memory architectures paralleling human cognitive systems \citep{wu2025from}.

\paragraph{Short-Term Memory Mechanisms} Short-term memory in LLMs operates through the context window, serving as working memory maintaining immediate access to previously processed tokens \citep{yue2024fragrel}. This functionality is implemented through key-value caches storing token representations but disappearing when sessions terminate \citep{rezazadeh2024from}. Architectural variations demonstrate significant differences: transformer-based models implement working memory systems flexibly retrieving individual token representations across arbitrary delays, while LSTM architectures maintain coarser, rapidly-decaying semantic representations weighted toward earliest items \citep{armeni2022characterizing}.

Modern LLM short-term memory frequently manifests as in-context learning, reflecting models' ability to acquire and process information temporarily within context windows \citep{xie2024travelplanner, brown2020language}. This enables few-shot learning and task adaptation without parameter updates. Research identifies three primary memory configurations: full memory (utilizing entire context history), limited memory (using context subsets), and memory-less operation (without historical context) \citep{torre2023llmr}. Despite advances expanding context windows to millions of tokens, LLMs struggle with effective reasoning over extended contexts, particularly when relevant information appears in middle positions \citep{rezazadeh2024from, liu2023lost}.

\paragraph{Long-Term Memory Implementations} LLMs face significant challenges maintaining long-term memory due to context window limitations and catastrophic forgetting \citep{cao2023retentive}. External memory-based methods address these limitations by utilizing physical storage to cache historical information, allowing relevant history retrieval without maintaining all information within constrained context windows \citep{liu2023think, zhong2023memorybank}. These approaches contrast with internal memory-based methods focusing on reducing self-attention computational costs to expand sequence length \citep{liu2023think, fournier2021practical}.

Long-term memory implementations categorize into knowledge-organization methods (structuring memory into interconnected semantic networks), retrieval mechanism-oriented approaches (integrating semantic retrieval with forgetting curve mechanisms), and architecture-driven methods (implementing hierarchical structures with explicit read-write operations) \citep{kang2025memory, zhong2023memorybank, huang2024emotional}. Memory storage representations can be further divided into token-level memory (information stored as structured text for direct retrieval) and latent-space memory (utilizing high-dimensional vectors for abstract and compact information representation) \citep{yang2024language, wang2025extending}. Advanced approaches incorporate psychological principles, with MemoryBank implementing Ebbinghaus Forgetting Curve theory for selective memory preservation based on temporal factors \citep{zhong2023memorybank}, emotion-aware frameworks employing Mood-Dependent Memory theory \citep{huang2024emotional}, and memorization mechanisms balancing performance advantages with privacy concerns through extraction vulnerability analysis \citep{tirumala2022memorization, carlini2018secret, carlini2020extracting}.

\paragraph{Memory Access Patterns and Structures} LLMs exhibit characteristic memory access patterns with notable similarities to human cognitive processes, demonstrating clear primacy and recency effects when recalling information lists \citep{janik2023aspects}. Memory retrieval operates through sequential access (retrieving content in consecutive order) and random access (accessing information from arbitrary points without processing preceding content) \citep{zhu2024beyond}. Memory persistence studies employ recognition experiments, recall experiments, and retention experiments to quantify information accessibility duration and retrieval conditions \citep{orhan2023recognition}, with cognitive psychology concepts like semantic and episodic memory integration improving LLM information synthesis capabilities \citep{du2024perltqa}.

Memory organization encompasses diverse structural approaches including textual-form storage (complete and recent agent-environment interactions, retrieved historical interactions, external knowledge), knowledge representation structures (chunks, knowledge triples, atomic facts, summaries, mixed approaches), hierarchical systems with library-enhanced reasoning components, and functional patterns organized by tasks, temporal relevance, or semantic relationships \citep{zhang2024survey, zeng2024structural, tang2025chemagent}. Core memory operations include encoding (transforming textual information into latent space embeddings), retrieval (accessing relevant information based on semantic relevance, importance, and recency), reflection (extracting higher-level insights), summarization (condensing texts while highlighting critical points), utilization (integrating memory components for unified outputs), forgetting (selective information discarding), truncation (formatting within token limitations), and judgment (assessing information importance for storage prioritization) \citep{zhang2025memengine}. These structures offer varying trade-offs between comprehensiveness, retrieval efficiency, and computational requirements.

\newcommand{\cmark}{$\checkmark$}
\newcommand{\xmark}{$\times$}

\begin{table*}[ht]
    \centering
    \footnotesize
    \label{tab:memory_forms_summary}
    \begin{tabular}{l|cccc|cc}
        \toprule
        \multirow{2}{*}{\textbf{Model}} & \multicolumn{4}{c|}{\textbf{Textual Form}} & \multicolumn{2}{c}{\textbf{Parametric Form}} \\
        \cmidrule(lr){2-5} \cmidrule(lr){6-7}
        & \textbf{Complete} & \textbf{Recent} & \textbf{Retrieved} & \textbf{External} & \textbf{Fine-tuning} & \textbf{Editing} \\
        \midrule
        
        \rowcolor{gray!20} \multicolumn{7}{c}{\textbf{Core Memory Systems}} \\
        MemoryBank \citep{zhong2023memory} & \xmark & \xmark & \cmark & \xmark & \xmark & \xmark \\
        RET-LLM \citep{modarressi2024retllm} & \xmark & \xmark & \cmark & \xmark & \xmark & \xmark \\
        ChatDB \citep{hu2023chatdb} & \xmark & \xmark & \cmark & \xmark & \xmark & \xmark \\
        TiM \citep{liu2023thinkinmemory} & \xmark & \xmark & \cmark & \xmark & \xmark & \xmark \\
        Voyager \citep{wang2023voyager} & \xmark & \xmark & \cmark & \xmark & \xmark & \xmark \\
        MemGPT \citep{packer2024memgpt} & \xmark & \cmark & \cmark & \xmark & \xmark & \xmark \\
        RecMind \citep{wang2024recmind} & \cmark & \xmark & \xmark & \xmark & \xmark & \xmark \\
        Retroformer \citep{yao2024retroformer} & \cmark & \xmark & \xmark & \cmark & \cmark & \xmark \\
        ExpeL \citep{zhao2024expel} & \cmark & \xmark & \cmark & \cmark & \xmark & \xmark \\
        Synapse \citep{zheng2024synapse} & \xmark & \xmark & \cmark & \xmark & \xmark & \xmark \\
        
        \midrule
        \rowcolor{gray!20} \multicolumn{7}{c}{\textbf{Agent-Based Systems}} \\
        ChatDev \citep{qian2024chatdev} & \cmark & \xmark & \xmark & \xmark & \xmark & \xmark \\
        InteRecAgent \citep{huang2024recommender} & \xmark & \cmark & \cmark & \cmark & \xmark & \xmark \\
        TPTU \citep{ruan2023tptu, kong2023tptuv2} & \cmark & \xmark & \xmark & \cmark & \xmark & \xmark \\
        MetaGPT \citep{hong2024metagpt} & \cmark & \xmark & \xmark & \xmark & \xmark & \xmark \\
        S³ \citep{gao2025s3} & \xmark & \xmark & \cmark & \xmark & \xmark & \xmark \\
        Mem0 \citep{chhikara2025mem0} & \xmark & \xmark & \cmark & \xmark & \xmark & \xmark \\
          
        \midrule
        \rowcolor{gray!20} \multicolumn{7}{c}{\textbf{Advanced Memory Architectures}} \\
        Larimar \citep{das2024larimar} & \xmark & \cmark & \cmark & \xmark & \xmark & \cmark \\
        EM-LLM \citep{fountas2024humanlike} & \xmark & \cmark & \cmark & \xmark & \xmark & \xmark \\
        Controllable Working Memory \citep{li2022largelanguagemodelscontrollable} & \cmark & \cmark & \cmark & \xmark & \cmark & \xmark \\
        Working Memory Hub \citep{guo2024empower} & \cmark & \cmark & \cmark & \cmark & \xmark & \xmark \\
        
        \midrule
        \rowcolor{gray!20} \multicolumn{7}{c}{\textbf{Recent and Emerging Systems}} \\
        LLM-based Opinion Dynamics \citep{chuang2024simulating} & \xmark & \xmark & \cmark & \xmark & \xmark & \xmark \\
        Memory Sandbox \citep{huang2023memorysandbox} & \xmark & \xmark & \cmark & \xmark & \xmark & \cmark \\
        A-MEM \citep{xu2025amemagenticmemoryllm} & \xmark & \xmark & \cmark & \xmark & \xmark & \cmark \\
        MemEngine \citep{zhang2025memengine} & \xmark & \xmark & \cmark & \cmark & \xmark & \xmark \\
        HIAGENT \citep{hu2024hiagent} & \xmark & \cmark & \cmark & \xmark & \xmark & \xmark \\
        MemInsight \citep{salama2025meminsight} & \xmark & \xmark & \cmark & \cmark & \xmark & \xmark \\
        Memory Sharing (MS) \citep{gao2024memory} & \xmark & \xmark & \cmark & \cmark & \xmark & \xmark \\
        MemoRAG \citep{qian2025memorag} & \cmark & \xmark & \cmark & \cmark & \cmark & \xmark \\
        Echo \citep{liu2025echo} & \cmark & \cmark & \cmark & \cmark & \cmark & \xmark \\
        
        \bottomrule
    \end{tabular}
    \caption{Extended from \citep{zhang2024survey}: Memory implementation patterns. \cmark = Adopted, \xmark = Not Adopted}
    
\end{table*}

\subsubsection{Memory-Enhanced Agents}

Memory systems fundamentally transform LLMs from stateless pattern processors into sophisticated agents capable of persistent learning and adaptation across extended interactions \citep{yehudai2025survey}. Memory-enhanced agents leverage both short-term memory (facilitating real-time responses and immediate context awareness) and long-term memory (supporting deeper understanding and knowledge application over extended periods) to adapt to changing environments, learn from experiences, and make informed decisions requiring persistent information access \citep{yehudai2025survey}.

\paragraph{Agent Architecture Integration} Contemporary LLM agents employ memory systems analogous to computer memory hierarchies, with short-term memory functioning as primary storage for contextual understanding within context windows, while long-term memory serves as persistent storage for extended information retention \citep{mi2025building}. From object-oriented perspectives, AI systems generate personal memories related to individual users and system memories containing intermediate task results \citep{wu2025from}. Structured frameworks like MemOS classify memory into Parametric Memory (knowledge encoded in model weights), Activation Memory, and Plaintext Memory, with parametric memory representing long-term knowledge embedded within feedforward and attention layers enabling zero-shot generation \citep{li2025memos}.

Memory integration frameworks have evolved to address LLM limitations through sophisticated architectures. The Self-Controlled Memory (SCM) framework enhances long-term memory through LLM-based agent backbones, memory streams, and memory controllers managing updates and utilization \citep{liang2023scm}. The REMEMBERER framework equips LLMs with experience memory exploiting past episodes across task goals, enabling success/failure learning without parameter fine-tuning through verbal reinforcement and self-reflective feedback mechanisms \citep{zhang2023large}. Advanced systems like MemLLM implement structured read-write memory modules addressing challenges in memorizing rare events, updating information, and preventing hallucinations \citep{modarressi2024memllm}. Autonomous agents leveraging LLMs rely on four essential components—perception, memory, planning, and action—working together to enable environmental perception, interaction recall, and real-time planning and execution \citep{li2024surveying, aratchige2025llms}.

\paragraph{Real-World Applications} Memory-enhanced LLM agents have demonstrated transformative impact across diverse application domains. In conversational AI, memory systems enable more natural, human-like interactions by recalling past experiences and user preferences to deliver personalized, context-aware responses. Commercial implementations include Charlie Mnemonic (combining Long-Term, Short-Term, and episodic memory using GPT-4), Google Gemini (leveraging long-term memory for personalized experiences across Google's ecosystem), and ChatGPT Memory (remembering conversations across sessions) \citep{lee2024towards}. User simulation applications employ LLM-powered conversational agents mimicking human behavior for cost-effective dialogue system evaluation, adapting flexibly across open-domain dialogues, task-oriented interactions, and conversational recommendation \citep{deng2024large}, with systems like Memory Sandbox enabling user control over conversational memories through data object manipulation \citep{huang2023memory}.

Task-oriented agents utilize memory to perform complex autonomous operations with minimal human intervention, employing LLMs as controllers extended through multimodal perception, tool utilization, and external memory \citep{wu2024retrieval}. Applications span recommendation systems (RecMind providing personalized recommendations through planning and external knowledge, InteRecAgent employing LLMs with recommender models as tools), autonomous driving (DiLu instilling human-like knowledge through reasoning, reflection, and memory), scientific research (ChemCrow automating chemical synthesis design and execution), and social simulation (generative agents exhibiting believable behavior through memory storage and synthesis) \citep{tan2023user, liang2025llm, bran2023augmenting, park2023generative}. Proactive conversational agents address challenges in strategic dialogue scenarios requiring goal-oriented conversation steering through prompt-based policy planning methods and AI feedback generation based on dialogue history \citep{deng2024large, deng2023prompting}.

Personalized assistant applications leverage memory to maintain coherent long-term relationships with users, with memory components serving as structured repositories storing contextually relevant information including user preferences and historical interactions \citep{huang2025survey}. Domain-specific implementations include healthcare assistants employing memory coordination for medical interactions \citep{zhang2025personaagent, zhang2023llm}, recommendation agents leveraging external knowledge bases \citep{zhang2025personaagent, zhang2023generative}, educational agents providing context-aware support through memory-enabled progress tracking \citep{liang2025llm}, and specialized frameworks like MARK enhancing personalized AI assistants through user preference memory \citep{ganguli2025mark}.

\paragraph{Memory Technologies and Integration Methods} Memory technology evolution addresses fundamental context window limitations through RAG, which combines parametric and non-parametric memory for language generation using pre-trained seq2seq models and dense vector indices \citep{xue2024comfybench, lewis2020retrieval}. This approach enables access to information beyond parameter storage without requiring retraining, significantly extending knowledge capabilities. Advanced memory mechanisms including vector databases and retrieval-augmented generation enable vast information storage with quick relevant data access, incorporating short-term contextual memory and long-term external storage \citep{aratchige2025llms, guu2020realm, xiong2024converging, johnson2017billion}.

Non-parametric approaches maintain frozen LLM parameters while leveraging external resources like RAG to enrich task contexts \citep{schoepp2025evolving}. Systems like Reflexion implement verbal reinforcement through self-reflective feedback in episodic memory buffers, while REMEMBERER incorporates persistent experience memory enabling learning from past successes and failures. Advanced architectures like MemoryBank enable memory retrieval, continuous evolution through updates, and personality adaptation by integrating previous interaction information \citep{xu2025mem, zhong2023memorybank}.

Specialized memory architectures address particular agent requirements through sophisticated organization and retrieval mechanisms. While early systems required predefined storage structures and retrieval timing, newer systems like Mem0 incorporate graph databases following RAG principles for more effective memory organization and relevance-based retrieval \citep{xu2025mem}. Commercial and open-source implementations including OpenAI ChatGPT Memory, Apple Personal Context, mem0, and MemoryScope demonstrate widespread adoption of memory systems for enhanced personalization capabilities \citep{wu2025from}. Tool-augmentation paradigms validate effectiveness in complex task decomposition while leveraging world interaction tools, with memory-enhanced agents becoming central to modern AI systems performing complex tasks through natural language integration of planning, tool use, memory, and multi-step reasoning \citep{epperson2025interactive, guo2024large, wang2023survey, andreas2022language}.

\subsubsection{Evaluation and Challenges}

Memory evaluation frameworks have emerged as critical components for systematically assessing LLM agent capabilities across multiple dimensions, reflecting the multifaceted nature of memory in intelligent systems. These comprehensive evaluation approaches reveal significant challenges while pointing toward promising research directions that could unlock new capabilities for memory-enhanced agents.

\paragraph{Evaluation Frameworks and Metrics} Contemporary memory evaluation employs specialized metrics extending beyond traditional NLP performance indicators to capture nuanced memory functionality aspects \citep{zhang2024memsim}. Effectiveness metrics focus on factual information storage and utilization through accuracy measures (correctness of responses based on historical messages) and recall@5 indicators (percentage of relevant messages retrieved within top-5 results). Efficiency metrics examine temporal aspects through response time (duration for information retrieval and utilization) and adaptation time (period required for new information storage) \citep{zhang2024memsim}.

Extensive benchmarks such as LongMemEval assess five fundamental long-term memory capabilities: information extraction, temporal reasoning, multi-session reasoning, knowledge updates, and abstention through 500 carefully selected questions, demonstrating 30\% accuracy degradation in commercial assistants throughout prolonged interactions, while automated memory evaluation frameworks facilitate thorough assessment extending beyond passkey search methodologies \citep{xia2025minerva}. Dedicated frameworks target episodic memory via benchmarks assessing temporally-situated experiences, with research demonstrating that cutting-edge models including GPT-4, Claude variants, and Llama 3.1 encounter difficulties with episodic memory challenges involving interconnected events or intricate spatio-temporal associations even in comparatively brief contexts \citep{huet2025episodic}. Contemporary LLM benchmarks predominantly concentrate on assessing models' retention of factual information and semantic relationships while substantially overlooking episodic memory assessment—the capacity to contextualize memories with temporal and spatial occurrence details \citep{pink2024assessing}.

Task-specific evaluations encompass long-context passage retrieval (locating specific paragraphs within extended contexts), long-context summarization (developing comprehensive understanding for concise summaries), NarrativeQA (answering questions based on lengthy narratives), and specialized benchmarks like MADail-Bench evaluating both passive and proactive memory recall in conversational contexts with novel dimensions including memory injection, emotional support proficiency, and intimacy assessment \citep{zhang2024survey, zhu2025evolutionary, kocisk2017narrativeqa, he2024madial}. Additional task-specific frameworks include QMSum for meeting summarization, QuALITY for reading comprehension, DialSim for dialogue-based QA requiring spatiotemporal memory, and MEMENTO for personalized embodied agent evaluation using two-stage processes to assess memory utilization in physical environment tasks \citep{zhu2025evolutionary, kwon2025embodied}.

\paragraph{Current Limitations and Challenges} Memory evaluation faces substantial challenges limiting effective assessment of capabilities. Fundamental limitations include absence of consistent, rigorous methodologies for assessing memory performance, particularly regarding generalization beyond training data \citep{fortunato2019generalization}. The lack of standardized benchmarks specifically designed for long-term memory evaluation represents another significant obstacle, with existing frameworks often failing to capture the full spectrum of memory capabilities needed for human-like intelligence \citep{wan2025storybench}.

Architectural constraints significantly complicate evaluation efforts, as most contemporary LLM-based agents operate in fundamentally stateless manners, treating interactions independently without truly accumulating knowledge incrementally over time \citep{zheng2025lifelongagentbench, zheng2024towards}, despite advances in working memory through attentional tagging mechanisms enabling flexible memory representation control \citep{qin2022working}. This limitation prevents genuine lifelong learning assessment—a cornerstone of human-level intelligence involving continuous knowledge acquisition, retention, and reuse across diverse contexts and extended time horizons.

Methodological issues arise when isolating memory-specific performance from other intelligence aspects, challenging determination of whether failures stem from inadequate memory mechanisms or reasoning limitations \citep{fortunato2019generalization}. Dynamic memory usage in real-world applications poses evaluation challenges, as controlled laboratory tests inadequately capture memory system performance in complex scenarios where information relevance changes unpredictably \citep{wan2025storybench}.

\paragraph{Optimization Strategies and Future Research Directions} Memory optimization encompasses diverse techniques enhancing utilization while minimizing computational overhead and maximizing efficiency. Biologically-inspired forgetting mechanisms provide effective optimization approaches, with frameworks like MemoryBank implementing Ebbinghaus forgetting curves to selectively preserve and discard information based on temporal factors and significance \citep{zhong2023memorybank}. Reflection-based optimization through systems like Reflexion enables performance assessment through integrated evaluation and self-reflection, creating dual feedback systems refining memory and behavior through continuous learning \citep{gao2023large}.

Hierarchical memory structures optimize information organization through multi-level formats enabling efficient retrieval, demonstrated by Experience-based Hierarchical Control frameworks with rapid memory access modules \citep{qiao2025efficiently}, memory consolidation processes through bidirectional fast-slow variable interactions \citep{benna2015complex}, and Adaptive Cross-Attention Networks dynamically ranking memories based on query relevance \citep{hong2025enhancing}.

Future research directions encompass hybrid memory frameworks combining parametric precision with non-parametric efficiency \citep{schoepp2025evolving}, automated feedback mechanisms for scalable response evaluation \citep{rasal2024artificial}, multi-agent memory systems enabling collaborative learning through shared external memories \citep{gao2024memory}, enhanced metadata learning with knowledge graph integration \citep{ren2025towards, hatalis2024memory}, domain-specific memory architectures for specialized applications \citep{jin2023genegpt}, cognitive-inspired optimization incorporating memory consolidation during inactive periods \citep{mcclelland1995why}, and parameter-efficient memory updates through techniques like Low-Rank Adaptation for efficient knowledge integration \citep{hu2021lora, fan2025llm}. These developments promise advancing memory-enhanced LLM agents toward sophisticated, human-like cognitive capabilities while addressing computational and architectural limitations, with applications extending to long-term robotic planning, real-world decision-making systems, and collaborative AI assistants through streaming learning scenarios and continuous feedback integration \citep{wu2024streambench, zhao2023expel, you2024llm}.
\subsection{Tool-Integrated Reasoning}
\label{subsec:tool_augmented_systems}

Tool-Integrated Reasoning transforms language models from passive text generators into active world interactors capable of dynamic tool utilization and environmental manipulation. This implementation enables models to transcend their inherent limitations through function calling mechanisms, integrated reasoning frameworks, and sophisticated environment interaction capabilities.

\begin{figure}[h]
  \centering
  \includegraphics[width=\textwidth]{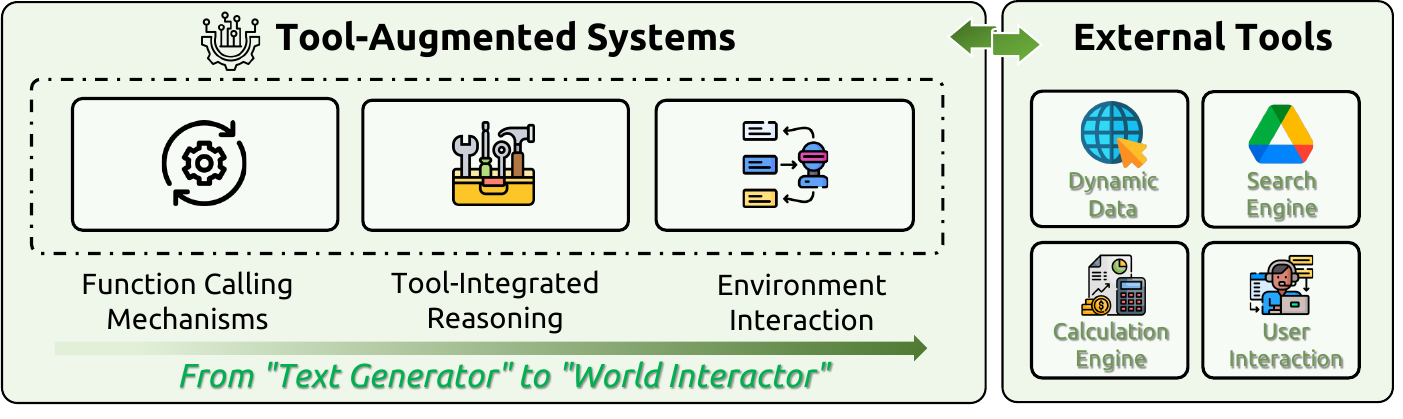}
  \caption{Tool-Augmented Systems Framework: Evolution from text generators to world interactors through function calling mechanisms, tool-integrated reasoning, and environment interaction capabilities.}
  \label{fig:tool_context_framework}
\end{figure}

\subsubsection{Function Calling Mechanisms}
\label{subsubsec:function_calling_mechanisms}

Function calling transforms LLMs from generative models into interactive agents through structured output generation leveraging functions' abstraction mechanism, enabling external tool manipulation and access to current, domain-specific information for complex problem-solving \citep{abdelaziz2024granite, lin2024training, ghica2009function, qu2024tool, basu2024nestful, kang2023quantitative, wang2024weaver}.

Evolution began with Toolformer's self-supervised approach demonstrating autonomous API learning, inspiring ReAct's ``thought-action-observation'' cycle, progressing through specialized models like Gorilla and comprehensive frameworks including ToolLLM, RestGPT, with OpenAI's JSON standardization, while advanced systems like Chameleon enabled multimodal question answering and TaskMatrix.AI managed AI models across domains \citep{schick2023toolformer, erdogan2024tinyagent, liang2025reasoning, kim2023llm, saha2024sequential, qin2023tool, qin2023toolllm, lu2023chameleon, liang2023taskmatrix, shen2023hugginggpt}.

Technical implementation involves fine-tuning (dominant method providing stable capabilities via extensive API training but requiring significant resources) and prompt engineering (flexible, resource-efficient but unstable), with approaches like ``Reverse Chain'' enabling API operation via prompts, addressing challenges in large tool management \citep{he2024achieving, abdelaziz2024granite, zhang2023reverse, moon2024efficient, chen2024facilitating, faghih2025gaming}.

Core process encompasses intent recognition, function selection, parameter-value-pair mapping, function execution, and response generation, with modern implementations utilizing structured LLM outputs for external program interaction, while tools include diverse interfaces (digital systems, scratch pads, user interactions, other LLMs, developer code), requiring complex navigation of tool selection, argument formulation, and result parsing \citep{yehudai2025survey, lin2024training, wang2024what, cui2025enhancing, shi2024tool, lee2024functionchat, ross2025when}.

\paragraph{Training Methodologies and Data Systems} Training methodologies evolved from basic prompt-based approaches to sophisticated multi-task learning frameworks, with fine-tuning on specialized datasets through systems like ToolLLM and Granite-20B-FunctionCalling, beginning with synthetic single-tool data followed by human annotations \citep{he2024achieving, abdelaziz2024granite, gunter2024apple, mialon2023augmented, yang2023teaching}.

Data generation strategies include Weaver's GPT-4-based environment synthesis, APIGen's hierarchical verification pipelines (format checking, function execution, semantic verification), generating 60,000+ high-quality entries across thousands of APIs \citep{wang2024weaver, xiao2025ultra, yehudai2025survey, wu2024seal, bensal2025reflect, zhuang2023toolqa, maria2025compass}.

Tool selection enhancement involves irrelevance-aware data augmentation, with Hammer's function masking techniques, oracle tool mixing for increased difficulty, tool intent detection synthesis for over-triggering mitigation, emphasizing high-quality data through stringent filtering and format verification \citep{lin2024hammer, acikgoz2025desideratum, gunter2024apple, iskander2024quality, zeng2025itool, ding2025toolcoder}.

Self-improvement paradigms reduce external supervision dependence through JOSH algorithm's sparse reward simulation environments and TTPA's token-level optimization with error-oriented scoring, demonstrating improvements while preserving general capabilities \citep{lattimer2024sparse, huang2025ttpa, gupta2024codenav, yin2025magnet}.

Sophisticated benchmarks include API-Bank (73 APIs, 314 dialogues), StableToolBench (API instability solutions), NesTools (nested tool evaluation), ToolHop (995 queries, 3,912 tools), addressing single-tool to multi-hop scenarios \citep{li2023api, guo2024stabletoolbench, han2024nestools, ye2025toolhop, paranjape2023art, styles2024workbench, yao2024bench, song2025callnavi}.

\subsubsection{Tool-Integrated Reasoning}
\label{subsubsec:tool_integrated_reasoning}

Tool-Integrated Reasoning (TIR) represents a paradigmatic advancement in Large Language Model capabilities, addressing fundamental limitations including outdated knowledge, calculation inaccuracy, and shallow reasoning by enabling dynamic interaction with external resources during the reasoning process \citep{qian2025toolrl}. Unlike traditional reasoning approaches that rely exclusively on internal model knowledge, TIR establishes a synergistic relationship where reasoning guides complex problem decomposition into manageable subtasks while specialized tools ensure accurate execution of each computational step \citep{mialon2023augmented}. This paradigm extends beyond conventional text-based reasoning by requiring models to autonomously select appropriate tools, interpret intermediate outputs, and adaptively refine their approach based on real-time feedback \citep{qian2025toolrl}.

The evolution of TIR methodologies encompasses three primary implementation categories addressing distinct aspects of tool utilization optimization. Prompting-based methods guide models through carefully crafted instructions without additional training, exemplified by approaches that decompose mathematical problems into executable code while delegating computation to Python interpreters \citep{chen2022program, li2023chain}. Supervised fine-tuning approaches teach tool usage through imitation learning, with systems like ToRA focusing on mathematical problem-solving by integrating natural language reasoning with computational libraries and symbolic solvers \citep{gou2023tora}. Reinforcement learning methods optimize tool-use behavior through outcome-driven rewards, though current implementations often prioritize final correctness without considering efficiency, potentially leading to cognitive offloading phenomena where models over-rely on external tools \citep{dong2025tool}.

In operational terms, TIR-based agents serve as intelligent orchestrators that systematically interweave cognitive processing with external resource engagement to achieve targeted outcomes \citep{wang2025toward}. This mechanism requires the harmonious integration of intrinsic reasoning capabilities and extrinsic tool utilization for progressive knowledge synthesis toward objective fulfillment, where the agent's execution pathway is formally characterized as a structured sequence of tool activations coupled with corresponding information assimilation events \citep{wang2025toward}. Emerging developments have established Agentic Reasoning architectures that amplify language model intelligence by incorporating autonomous tool-deploying agents, fluidly orchestrating web-based information retrieval, computational processing, and layered reasoning-memory integration to tackle sophisticated challenges necessitating comprehensive research and cascaded logical analysis \citep{wu2025agentic}.

\paragraph{Implementation Frameworks and Paradigms} Single-tool frameworks established foundational principles of tool-integrated reasoning through specialized implementations targeting specific computational domains. Program-Aided Language Models (PAL) pioneered problem decomposition strategies by generating executable code while delegating mathematical computations to Python interpreters \citep{gao2022pal}. ToolFormer demonstrated that language models could learn external API usage with minimal demonstrations, incorporating calculators, search engines, and diverse tools to enhance computational capabilities \citep{schick2023toolformer}. ToRA advanced mathematical reasoning by integrating natural language processing with computational libraries and symbolic solvers, while ReTool applied reinforcement learning to optimize code interpreter usage, demonstrating improvements in self-correction patterns \citep{gou2023tora, zhang2025computational, singh2025agentic}. Self-Edit utilizes execution results of generated code to improve code quality for competitive programming tasks, employing a fault-aware code editor to correct errors based on test case results \citep{zhang2023self}.

Multi-tool coordination systems address the complexity of orchestrating heterogeneous tools within integrated reasoning architectures. ReAct pioneered the interleaving of reasoning traces with task-specific actions, enabling models to think and act complementarily where reasoning supports plan tracking while actions interface with external information sources \citep{yao2022react}. Chameleon introduced plug-and-play compositional reasoning by synthesizing programs combining vision models, search engines, and Python functions with an LLM-based planner core \citep{lu2023chameleon}. AutoTools established automated frameworks transforming raw tool documentation into executable functions, reducing manual engineering requirements in tool integration \citep{hou2025model, shi2024tool}. Chain-of-Agents (CoA) trains models to decode reasoning chains with abstract placeholders, subsequently calling domain-specific tools to fill knowledge gaps \citep{li2025cort, zhang2024chain}.

Agent-based frameworks represent the most sophisticated evolution of TIR systems, moving beyond static prompting approaches to create autonomous and adaptive AI systems. Unlike conventional tool-use that follows rigid patterns, agent models learn to couple Chain-of-Thought (CoT) and Chain-of-Action (CoA) patterns into their core behavior, resulting in stronger logical coherence and natural transitions between reasoning and action \citep{zhang2025agent}. These systems build upon foundational agent architectures including reactive systems that map perceptions directly to actions, deliberative systems implementing Belief-Desire-Intention (BDI) models, and hybrid architectures combining multiple subsystems in hierarchical structures \citep{ma2024computational}.

\begin{table*}[ht]
\centering
\scriptsize

\label{tab:method-tools}
\renewcommand{\arraystretch}{1.2} 
\definecolor{lightgray}{RGB}{245,245,245} 
\resizebox{\textwidth}{!}{%
\begin{tabular}{l|cccccccc}
\toprule
\multirow{2}{*}{\textbf{Method}} & \multicolumn{8}{c}{\textbf{Tool Categories}} \\
\cmidrule(lr){2-9}
& \textbf{\begin{tabular}[c]{@{}c@{}}Search \&\\ Retrieval\end{tabular}} & \textbf{\begin{tabular}[c]{@{}c@{}}Computation \&\\ Code Execution\end{tabular}} & \textbf{\begin{tabular}[c]{@{}c@{}}Knowledge Base\\ \& QA\end{tabular}} & \textbf{\begin{tabular}[c]{@{}c@{}}APIs \&\\ External Services\end{tabular}} & \textbf{\begin{tabular}[c]{@{}c@{}}Multimodal\\ Tools\end{tabular}} & \textbf{\begin{tabular}[c]{@{}c@{}}Language\\ Processing\end{tabular}} & \textbf{\begin{tabular}[c]{@{}c@{}}Interactive\\ Environments\end{tabular}} & \textbf{\begin{tabular}[c]{@{}c@{}}Domain-Specific\\ Tools\end{tabular}} \\
\midrule
\rowcolor{white}ReAct \cite{yao2023react} & \cmark &  & \cmark &  &  &  & \cmark &  \\
\rowcolor{lightgray}Toolformer \cite{schick2023toolformer} & \cmark & \cmark & \cmark &  &  & \cmark &  & \cmark \\
\rowcolor{white}ToolkenGPT \cite{hao2023toolkengpt} & \cmark & \cmark & \cmark & \cmark &  &  & \cmark &  \\
\rowcolor{lightgray}ToolLLM \cite{qin2023toolllm} & \cmark & \cmark & \cmark & \cmark & \cmark & \cmark & \cmark & \cmark \\
\rowcolor{white}ToRA \cite{gou2023tora} &  & \cmark &  &  &  &  &  &  \\
\rowcolor{lightgray}PAL \cite{gao2023pal} &  & \cmark &  &  &  &  &  &  \\
\rowcolor{white}HuggingGPT \cite{shen2023hugginggpt} &  &  &  & \cmark & \cmark &  &  &  \\
\rowcolor{lightgray}GPT4Tools \cite{yang2023gpt4tools} &  &  &  &  & \cmark &  &  &  \\
\rowcolor{white}CRITIC \cite{gou2023critic} & \cmark & \cmark & \cmark &  &  &  &  &  \\
\rowcolor{lightgray}Chain of Code \cite{li2023chain} &  & \cmark &  &  &  &  &  &  \\
\rowcolor{white}TRICE \cite{qiao2023making} & \cmark & \cmark & \cmark &  &  & \cmark &  &  \\
\rowcolor{lightgray}TP-LLaMA \cite{chen2024advancing} & \cmark & \cmark & \cmark & \cmark & \cmark & \cmark & \cmark & \cmark \\
\rowcolor{white}AlignToolLLaMA \cite{chen2024towards} & \cmark & \cmark & \cmark & \cmark & \cmark & \cmark & \cmark & \cmark \\
\rowcolor{lightgray}ReTool \cite{feng2025retool} &  & \cmark &  &  &  &  &  &  \\
\rowcolor{white}Tool-Star \cite{dong2025toolstar} & \cmark & \cmark &  &  &  &  &  &  \\
\rowcolor{lightgray}ARTIST \cite{singh2025agentic} &  & \cmark &  &  &  &  &  &  \\
\rowcolor{white}Ego-R1 \cite{tian2025egor1} &  &  &  &  & \cmark &  &  &  \\
\rowcolor{lightgray}VTool-R1 \cite{wu2025vtoolr1} &  &  &  &  & \cmark &  &  &  \\
\rowcolor{white}KG-Agent \cite{jiang2024kgagent} &  &  & \cmark &  &  &  &  & \cmark \\
\rowcolor{lightgray}CACTUS \cite{mcnaughtoncactus} &  &  &  &  &  &  &  & \cmark \\
\rowcolor{white}MuMath-Code \cite{yin2024mumathcode} &  & \cmark &  &  &  &  &  &  \\
\rowcolor{lightgray}ToRL \cite{li2025torl} &  & \cmark &  &  &  &  &  &  \\
\rowcolor{white}MetaTool \cite{huang2023metatool} & \cmark & \cmark & \cmark & \cmark &  &  &  &  \\ 
\rowcolor{lightgray}ToolEyes \cite{ye2024tooleyes} &  &  &  & \cmark &  &  &  & \cmark \\
\rowcolor{white}Graph-CoT \cite{jin2024graph} &  &  & \cmark &  &  &  &  & \cmark \\
\rowcolor{lightgray}ToolRL \cite{qian2025toolrl} & \cmark & \cmark & \cmark & \cmark &  &  &  &  \\
\rowcolor{white}LATS \cite{zhou2023language} & \cmark &  &  &  &  &  &  & \cmark \\
\bottomrule
\end{tabular}%
}
\caption{Tool-augmented language model architectures: Comparison of multiple methods across 8 tool categories including search, computation, knowledge bases, APIs, multimodal, language tools, interactive environments, and domain-specific applications.}
\renewcommand{\arraystretch}{1.0} 
\end{table*}

\subsubsection{Agent-Environment Interaction}
\label{subsubsec:agent_environment_interaction}

Reinforcement learning approaches have emerged as superior alternatives to prompting-based methods and supervised fine-tuning for tool integration, enabling models to autonomously discover optimal tool usage strategies through exploration and outcome-driven rewards \citep{dong2025tool}. ReTool exemplifies this advancement by focusing on code interpreter optimization for mathematical reasoning, achieving 67.0\% accuracy on AIME2024 benchmarks after only 400 training steps, substantially outperforming text-based RL baselines reaching 40.0\% accuracy with extensive training \citep{feng2025retool}. This demonstrates that explicitly modeling tool use within decision processes enhances both reasoning capabilities and training efficiency.

Search-augmented reasoning systems represent innovative integrations of information retrieval directly into reasoning processes through specialized learning environments. The Search-R1 framework trains models to make dynamic decisions about when to search and what queries to generate during multi-step reasoning tasks, unlike traditional retrieval-augmented generation systems \citep{song2025searcher}. The architecture employs specialized token systems structuring reasoning and search processes, where models learn to generate reasoning steps interspersed with explicit search actions triggered through tokens that encapsulate generated queries \citep{liang2025reasoning}.

Multi-turn and customizable tool invocation frameworks address the complexity of coordinating multiple heterogeneous tools during reasoning processes. Recent developments include frameworks like VisTA that use reinforcement learning to enable visual agents to dynamically explore, select, and combine tools from diverse libraries based on empirical performance \citep{huang2025visualtoolagent}. ReVeal demonstrates self-evolving code agents via iterative generation-verification processes \citep{jin2025reveal}. In multimodal domains, systems like VideoAgent employ vision-language foundation models as tools for translating and retrieving visual information, achieving impressive performance on video understanding benchmarks \citep{wang2024videoagent, fan2024videoagent}.

\paragraph{Evaluation and Applications} Comprehensive evaluation of tool-integrated reasoning systems requires specialized benchmarks that measure tool-integrated capabilities rather than general model performance. MCP-RADAR provides a standardized evaluation framework employing strictly objective metrics derived from quantifiable performance data, with extensible design spanning software engineering, mathematical reasoning, and general problem-solving domains \citep{gao2025mcp}. The framework visualizes performance through radar charts highlighting model strengths and weaknesses across multiple dimensions, enabling systematic comparison of tool-integrated language models regardless of implementation mechanisms.

Real-world evaluation approaches reveal significant performance gaps between current systems and human-level capabilities, providing crucial insights into practical limitations and optimization opportunities. The General Tool Agents (GTA) benchmark addresses limitations in existing evaluations by featuring real human-written queries with implicit tool-use requirements, evaluation platforms with deployed tools across perception, operation, logic, and creativity categories, and authentic multimodal inputs including images and code snippets \citep{wang2024gta}. Results demonstrate substantial challenges for current LLMs, with GPT-4 completing less than 50

Function calling enabled sophisticated multi-agent systems where multiple LLM agents collaborate through coordinated tool use and task decomposition, with MAS leveraging collective intelligence through parallel processing, information sharing, and adaptive role assignment, while LLM integration enhanced capabilities in planning, specialization, and task decomposition through frameworks like DyLAN, MAD, and MetaGPT \citep{du2025survey, rosser2025agentbreeder, gronauer2021multi, chen2025from, li2024survey}. Advanced multi-agent function calling employs sophisticated orchestration mechanisms decomposing complex tasks into manageable subtasks, with fundamental approaches involving splitting reward machines into parallel execution units, each agent maintaining individual reward machines, local state spaces, and propositions, while adaptive orchestration enables dynamic agent selection based on context, responses, and status reports \citep{ardon2023learning, triedman2025multi, liu2024toolace_1, cao2025large}.
\subsection{Multi-Agent Systems}
\label{subsec:multi_agent_systems}

Multi-Agent Systems represent the pinnacle of collaborative intelligence, enabling multiple autonomous agents to coordinate and communicate for solving complex problems beyond individual agent capabilities. This implementation focuses on sophisticated communication protocols, orchestration mechanisms, and coordination strategies that enable seamless collaboration across diverse agent architectures.

\begin{figure}[h]
  \centering
  \includegraphics[width=\textwidth]{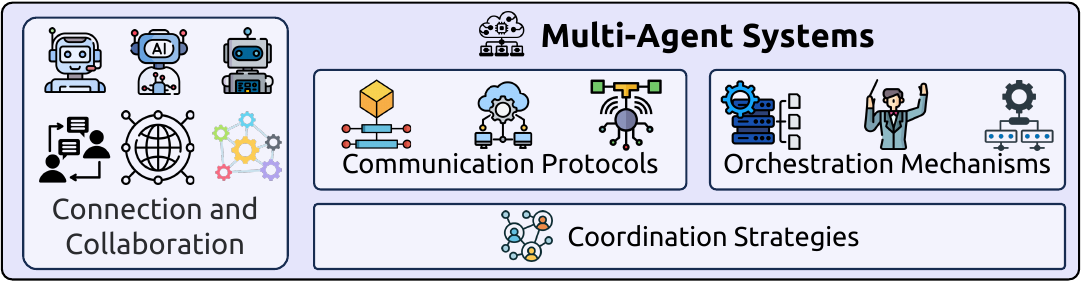}
  \caption{Multi-Agent Systems Framework: Overview of communication protocols, orchestration mechanisms, and coordination strategies for collaborative AI agent systems.}
  \label{fig:mas_framework}
\end{figure}

\subsubsection{Communication Protocols}
\label{subsubsec:communication_protocols}

Agent communication systems originate from the Knowledge Sharing Effort of the early 1990s, establishing foundational principles for autonomous entity coordination through standardized languages addressing interoperability challenges \citep{habler2025building, bravo2008aligning}. KQML emerged as the pioneering Agent Communication Language, introducing multi-layered architecture separating content, message, and communication layers while employing speech act theory \citep{habler2025building, bien2009call, lillis2017internalising, finin1994kqml}. FIPA ACL enhanced this foundation through semantic frameworks based on modal logic, feasibility preconditions, and rational effects \citep{wiki:acl, habler2025building, bien2009call}.

Interoperability requirements necessitate semantic-level communication capabilities enabling cross-platform agent understanding without extensive pre-communication setup, addressing increasing heterogeneity through ontology-based protocol formalization and Semantic Web technologies, while incorporating security mechanisms against communication vulnerabilities \citep{jeong2025study, berges2008semantic, huang2025agent, ji2004multi, mori2002stochastic, uytsel2001structured}.

\paragraph{Contemporary Protocol Ecosystem} Contemporary standardized protocols address fragmentation challenges hindering LLM agent collaboration \citep{yang2025survey, wang2025internet, hong2023metagpt}. MCP functions as ``USB-C for AI,'' standardizing agent-environment interactions through JSON-RPC client-server interfaces, enabling hundreds of servers across diverse domains while introducing security vulnerabilities \citep{sarkar2025survey, ehtesham2025survey, li2025from, fei2025mcp, ahmadi2025mcp, fang2025should, sanikommu2025model, wang2025towards, halloran2025mcp, xiong2025invisible, gan2025rag, szeider2024mcp, lumer2025scalemcp, feng2025get}.

A2A standardizes peer-to-peer communication through capability-based Agent Cards enabling task delegation and secure collaboration via JSON-based lifecycle models \citep{li2025from, ehtesham2025survey, sarkar2025survey}. ACP provides general-purpose RESTful HTTP communication supporting multipart messages and synchronous/asynchronous interactions with discovery, delegation, and orchestration features \citep{ferrag2025from, ehtesham2025survey}.

ANP extends interoperability to open internet through W3C decentralized identifiers and JSON-LD graphs, with emerging protocols AGNTCY and Agora diversifying standardization ecosystems \citep{ehtesham2025survey, liu2025acps, wang2025internet}. Progressive layering strategy: MCP provides tool access, ACP enables message exchange, A2A supports peer interaction, ANP extends network interoperability \citep{google:a2a, sarkar2025survey}.

\paragraph{LLM-Enhanced Communication Frameworks} LLMs transform agent communication through sophisticated natural language processing enabling unprecedented context sensitivity across academic and industrial applications spanning social science, natural science, and engineering domains \citep{jiang2023large, liu2024from, jin2024from, wang2023survey, xi2023rise, wang2024large, robinson2022leveraging, tulchinskii2024listening, qiu2025agentdistill}. Enhanced systems demonstrate cognitive synergy through specialized knowledge bases, planning, memory, and introspection capabilities, supporting cooperative, debate-oriented, and competitive communication paradigms \citep{jiang2023large, guo2024large}.

Communication structures encompass layered hierarchical organization, decentralized peer-to-peer networks, centralized coordination, and shared message pool architectures, complemented by sequential exchanges, universal language interfaces, and message-passing strategies \citep{guo2024large, yao2024comal, yan2025beyond, cheng2024exploring, helmi2025modeling, jiang2022learning, kim2024learning, lin2021common, nachmani2023spoken, shen2017natural}.

Framework implementations support comprehensive ecosystems: AutoGen enables dynamic response generation, MetaGPT provides shared message pools, CAMEL offers integrated orchestration, CrewAI facilitates adaptation, with reinforcement learning integration enhancing reward redesign, action selection, and policy interpretation \citep{crewai, aratchige2025llms, cao2024survey, sun2024llm, dong2024large, qin2022survey, sarkar2023testing, shi2020learning, yin2020tabert}. Human-agent communication introduces complex interaction landscapes through flexible participation and cognitive diversity, with agents inferring communicator properties and mirroring human communicative intentions \citep{zou2025survey, andreas2022language, lior2024computation}.

\subsubsection{Orchestration Mechanisms}
\label{subsubsec:orchestration_mechanisms}

Orchestration mechanisms constitute the critical coordination infrastructure for multi-agent systems, managing agent selection, context distribution, and interaction flow control \citep{rizk2020conversational}, enabling effective cooperation among human and non-human actors through user input processing, contextual distribution, and optimal agent selection based on capability assessment and response evaluation \citep{bandlamudi2023towards}, while managing message flow, ensuring task progression, and addressing task deviations \citep{choi2024malade}. Advanced orchestration frameworks incorporate intent recognition, contextual memory maintenance, and task dispatching components for intelligent coordination across domain-specific agents, with the Swarm Agent framework utilizing real-time outputs to direct tool invocations while addressing limitations in static tool registries and bespoke communication frameworks \citep{openai:swarm, fatouros2025towards, ehtesham2025survey}.

Contemporary orchestration strategies exhibit distinct operational paradigms: a priori orchestration determines agent selection through pre-execution analysis of user input and agent capabilities, while posterior orchestration distributes inputs to multiple agents simultaneously, utilizing confidence metrics and response quality assessment as demonstrated by the 3S orchestrator framework \citep{rizk2020unified}; function-based orchestration emphasizes agent selection from available pools, contextual information management, and conversation flow control \citep{bandlamudi2024building}; component-based orchestration employs dynamic planning processes where orchestrators arrange components in logical sequences based on user instructions, utilizing LLMs as component orchestration tools to generate workflows with embedded orchestration logic \citep{liu2025workteam}.

Emergent orchestration paradigms include puppeteer-style orchestration featuring centralized orchestrators that dynamically direct agents in response to evolving task states through reinforcement learning-based adaptive sequencing and prioritization, and serialized orchestration addressing collaboration topology complexity by unfolding collaboration graphs into reasoning sequences guided by topological traversal, enabling orchestrators to select single agents at each step based on global system state and task specifications \citep{dang2025multi}.

\paragraph{Context Management and Environmental Adaptation} Context serves as the foundational element guiding agent actions and interactions within orchestrated systems, supporting operational mode diversity while maintaining application individuality and task execution sequencing through global state maintenance that enables orchestration systems to track task execution progress across distributed nodes, providing agents with contextual awareness necessary for effective subtask performance within broader workflow contexts \citep{aminiranjbar2024dawn}. Session-based context refinement defines collaborative scope boundaries, facilitating event-driven orchestration where agents can enter and exit dynamically, create output streams, and contribute to shared session streams, with configurable sessions enabling agent inclusion based on user input or autonomous decision-making to create adaptable systems responsive to changing task requirements \citep{kandogan2025orchestrating}.

Well-designed interaction structures and task orchestration mechanisms underscore context's critical role in scalable multi-agent collaboration. Systems adapt communication patterns and agent roles to contextual requirements, supporting dynamic collaboration tailored to specific task demands through complex task decomposition and suitable agent assignment for subtask execution \citep{wang2025internet}. This contextual adaptation encompasses both organizational and operational dimensions, enabling systems to maintain coherence while accommodating environmental variability and evolving user requirements.

\subsubsection{Coordination Strategies}
\label{subsubsec:coordination_strategies}

Multi-agent orchestration encounters significant challenges in maintaining transactional integrity across complex workflows, with contemporary frameworks including LangGraph, AutoGen, and CAMEL demonstrating insufficient transaction support: LangGraph provides basic state management while lacking atomicity guarantees and systematic compensation mechanisms, AutoGen prioritizes flexible agent interactions without adequate compensatory action management potentially resulting in inconsistent system states following partial failures, and validation limitations emerge as many frameworks rely exclusively on large language models' inherent self-validation capabilities without implementing independent validation procedures, exposing systems to reasoning errors, hallucinations, and inter-agent inconsistencies \citep{chang2025sagallm}.

Context handling failures compound these challenges as agents struggle with long-term context maintenance encompassing both episodic and semantic information \citep{deshpande2025trail, wang2024executable}, while central orchestrator topologies introduce non-deterministic, runtime-dependent execution paths that enhance adaptability while complicating anomaly detection, requiring dynamic graph reconstruction rather than simple path matching \citep{he2025sentinelagent}, and environmental misconfigurations and LLM hallucinations can distract agentic systems, with poor recovery leading to goal deviation that becomes amplified in multi-agent setups with distributed subtasks \citep{deshpande2025trail, wang2023survey}.

Inter-agent dependency opacity presents additional concerns as agents may operate on inconsistent assumptions or conflicting data without explicit constraints or validation layers, necessitating anomaly detection incorporating reasoning over orchestration intent and planning coherence \citep{he2025sentinelagent}, while addressing these challenges requires comprehensive solutions such as the SagaLLM framework providing transaction support, independent validation procedures, and robust context preservation mechanisms \citep{chang2025sagallm}, and approaches like CodeAct integrating Python interpreters with LLM agents to enable code action execution and dynamic revision capabilities through multi-turn interactions \citep{wang2024executable}.

\paragraph{Applications and Performance Implications} Agent and context orchestration demonstrates practical utility across diverse application domains: healthcare applications employ context-switching mechanisms within specialized agent-based architectures performing information retrieval, question answering, and decision support, utilizing supervisory agents to interpret input features and assign subtasks to specialized agents based on clinical query type, user background, and data modality requirements \citep{li2025one, mcduff2023towards, tu2025towards}; network management applications leverage context-aware orchestration to address complexity challenges by equipping Points of Access with agents dedicated to unique contexts, enabling efficient network dynamics management through context-specific action sets including available service instances and network paths \citep{shokrnezhad2024toward}.

Business process management and simulation represent significant application areas through platforms like AgentSimulator, enabling process behavior discovery and simulation in orchestrated and autonomous settings where orchestrated behavior follows global control-flow patterns with activity selection dependent on previous activities and agent assignment based on capabilities and availability, while autonomous behavior operates through local control-flow and handover patterns acknowledging agent autonomy in collaborative work \citep{kirchdorfer2025discovering}.

Performance implications indicate that well-designed orchestration improves system effectiveness by leveraging distinct agent capabilities, with research demonstrating that human users frequently struggle with effective agent selection from available sets while automated orchestration enhances overall performance \citep{bhatt2025when}, motivating frameworks that learn agent capabilities online and orchestrate multiple agents under real-world constraints including cost, capability requirements, and operational limitations, with autonomy levels varying across implementations where some systems exhibit pronounced autonomy within designated phases, demonstrating adaptability in action management corresponding to task specificity and reaching Level 2 autonomy through contextual resource utilization \citep{hndler2023balancing}.

\section{Evaluation}
\label{sec:evaluation}

The evaluation of context-engineered systems presents unprecedented challenges that transcend traditional language model assessment paradigms. These systems exhibit complex, multi-component architectures with dynamic, context-dependent behaviors requiring comprehensive evaluation frameworks that assess component-level diagnostics, task-based performance, and overall system robustness \citep{peng2024survey, wang2024what}. 

The heterogeneous nature of context engineering components—spanning retrieval mechanisms, memory systems, reasoning chains, and multi-agent coordination—demands evaluation methodologies that can capture both individual component effectiveness and emergent system-level behaviors \citep{gao2025mcp, schick2023toolformer}.

\subsection{Evaluation Frameworks and Methodologies}
\label{subsec:evaluation_frameworks}

This subsection presents comprehensive approaches for evaluating both individual components and integrated systems in context engineering.

\subsubsection{Component-Level Assessment}
\label{subsubsec:component_assessment}

Intrinsic evaluation focuses on the performance of individual components in isolation, providing foundational insights into system capabilities and failure modes. 

For \textbf{prompt engineering} components, evaluation encompasses prompt effectiveness measurement through semantic similarity metrics, response quality assessment, and robustness testing across diverse input variations. Current approaches reveal brittleness and robustness challenges in prompt design, necessitating more sophisticated evaluation frameworks that can assess contextual calibration and adaptive prompt optimization \citep{wang2024what, lin2024training}.

\textbf{Long context processing} evaluation requires specialized metrics addressing information retention, positional bias, and reasoning coherence across extended sequences. The ``needle in a haystack'' evaluation paradigm tests models' ability to retrieve specific information embedded within long contexts, while multi-document reasoning tasks assess synthesis capabilities across multiple information sources. Position interpolation techniques and ultra-long sequence processing methods face significant computational challenges that limit practical evaluation scenarios \citep{ma2024megalodon, fu2025sliding}.

\textbf{Self-contextualization} mechanisms undergo evaluation through meta-learning assessments, adaptation speed measurements, and consistency analysis across multiple iterations. Self-refinement frameworks including Self-Refine, Reflexion, and N-CRITICS demonstrate substantial performance improvements, with GPT-4 achieving approximately 20\% improvement through iterative self-refinement processes \citep{madaan2023self, shinn2023reflexion, mousavi2023critics}. Multi-dimensional feedback mechanisms and ensemble-based evaluation approaches provide comprehensive assessment of autonomous evolution capabilities \citep{lee2024ask, lu2023self}.

\textbf{Structured and relational data integration} evaluation examines accuracy in knowledge graph traversal, table comprehension, and database query generation. However, current evaluation frameworks face significant limitations in assessing structural reasoning capabilities, with high-quality structured training data development presenting ongoing challenges. LSTM-based models demonstrate increased errors when sequential and structural information conflict, highlighting the need for more sophisticated benchmarks testing structural understanding \citep{melis2017state, linzen2016assessing, cheng2024potential}.

\subsubsection{System-Level Integration Assessment}
\label{subsubsec:system_assessment}

Extrinsic evaluation measures end-to-end performance on downstream tasks, providing holistic assessments of system utility through comprehensive benchmarks spanning question answering, reasoning, and real-world applications. 

System-level evaluation must capture emergent behaviors arising from component interactions, including synergistic effects where combined components exceed individual performance and potential interference patterns where component integration degrades overall effectiveness \citep{peng2024survey, wang2024what}.

\textbf{Retrieval-Augmented Generation} evaluation encompasses both retrieval quality and generation effectiveness through comprehensive metrics addressing precision, recall, relevance, and factual accuracy. Agentic RAG systems introduce additional complexity requiring evaluation of task decomposition accuracy, multi-plan selection effectiveness, and memory-augmented planning capabilities. Self-reflection mechanisms demonstrate iterative improvement through feedback loops, with MemoryBank implementations incorporating Ebbinghaus Forgetting Curve principles for enhanced memory evaluation \citep{huang2025survey, cheng2025survey, zhong2023memorybank, xiong2025rag, asai2023self}.

\textbf{Memory systems} evaluation encounters substantial difficulties stemming from the absence of standardized assessment frameworks and the inherently stateless characteristics of contemporary LLMs. LongMemEval offers 500 carefully curated questions that evaluate fundamental capabilities encompassing information extraction, temporal reasoning, multi-session reasoning, and knowledge updates. Commercial AI assistants exhibit 30\% accuracy degradation throughout extended interactions, underscoring significant deficiencies in memory persistence and retrieval effectiveness \citep{zhang2024memsim, xia2025minerva, huet2025episodic, pink2024assessing, he2024madial}. Dedicated benchmarks such as NarrativeQA, QMSum, QuALITY, and MEMENTO tackle episodic memory evaluation challenges \citep{kocisk2017narrativeqa, kwon2025embodied}.

\textbf{Tool-integrated reasoning systems} require comprehensive evaluation covering the entire interaction trajectory, including tool selection accuracy, parameter extraction precision, execution success rates, and error recovery capabilities. The MCP-RADAR framework provides standardized evaluation employing objective metrics for software engineering and mathematical reasoning domains. Real-world evaluation reveals significant performance gaps, with GPT-4 completing less than 50\% of tasks in the GTA benchmark, compared to human performance of 92\% \citep{gao2025mcp, wang2024gta, chakraborty2025tool, schick2023toolformer}. Advanced benchmarks including BFCL (2,000 testing cases), T-Eval (553 tool-use cases), API-Bank (73 APIs, 314 dialogues), and ToolHop (995 queries, 3,912 tools) address multi-turn interactions and nested tool calling scenarios \citep{fang2025zero, guo2024stabletoolbench, han2024nestools, ye2025toolhop, chen2023t, patil2025bfcl}.

\textbf{Multi-agent systems} evaluation captures communication effectiveness, coordination efficiency, and collective outcome quality through specialized metrics addressing protocol adherence, task decomposition accuracy, and emergent collaborative behaviors. Contemporary orchestration frameworks including LangGraph, AutoGen, and CAMEL demonstrate insufficient transaction support, with validation limitations emerging as systems rely exclusively on LLM self-validation capabilities without independent validation procedures. Context handling failures compound challenges as agents struggle with long-term context maintenance encompassing both episodic and semantic information \citep{chang2025sagallm, he2025sentinelagent, rizk2020unified}.

\subsection{Benchmark Datasets and Evaluation Paradigms}
\label{subsec:benchmark_datasets}

This subsection reviews specialized benchmarks and evaluation paradigms designed for assessing context engineering system performance.

\subsubsection{Foundational Component Benchmarks}
\label{subsubsec:foundational_benchmarks}

Long context processing evaluation employs specialized benchmark suites designed to test information retention, reasoning, and synthesis across extended sequences. Current benchmarks face significant computational complexity challenges, with O(n²) scaling limitations in attention mechanisms creating substantial memory constraints for ultra-long sequences. Position interpolation and extension techniques require sophisticated evaluation frameworks that can assess both computational efficiency and reasoning quality across varying sequence lengths \citep{ma2024megalodon, fu2025sliding, yang2025lserve}.

Advanced architectures including LongMamba and specialized position encoding methods demonstrate promising directions for long context processing, though evaluation reveals persistent challenges in maintaining coherence across extended sequences. The development of sliding attention mechanisms and memory-efficient implementations requires comprehensive benchmarks that can assess both computational tractability and task performance \citep{ye2025longmamba, gu2022parameterization}.

Structured and relational data integration benchmarks encompass diverse knowledge representation formats and reasoning patterns. However, current evaluation frameworks face limitations in assessing structural reasoning capabilities, with the development of high-quality structured training data presenting ongoing challenges. Evaluation must address the fundamental tension between sequential and structural information processing, particularly in scenarios where these information types conflict \citep{melis2017state, linzen2016assessing, cheng2024potential}.

\subsubsection{System Implementation Benchmarks}
\label{subsubsec:system_benchmarks}

Retrieval-Augmented Generation evaluation leverages comprehensive benchmark suites addressing diverse retrieval and generation challenges. Modular RAG architectures demonstrate enhanced flexibility through specialized modules for retrieval, augmentation, and generation, enabling fine-grained evaluation of individual components and their interactions. Graph-enhanced RAG systems incorporating GraphRAG and LightRAG demonstrate improved performance in complex reasoning scenarios, though evaluation frameworks must address the additional complexity of graph traversal and multi-hop reasoning assessment \citep{gao2024modular, singh2025agentic, guo2024lightrag}.

Agentic RAG systems introduce sophisticated planning and reflection mechanisms requiring evaluation of task decomposition accuracy, multi-plan selection effectiveness, and iterative refinement capabilities. Real-time and streaming RAG applications present unique evaluation challenges in assessing both latency and accuracy under dynamic information conditions \citep{huang2025survey, cheng2025survey, xiong2025rag}.

Tool-integrated reasoning system evaluation employs comprehensive benchmarks spanning diverse tool usage scenarios and complexity levels. The Berkeley Function Calling Leaderboard (BFCL) provides 2,000 testing cases with step-by-step and end-to-end assessments measuring call accuracy, pass rates, and win rates across increasingly complex scenarios. T-Eval contributes 553 tool-use cases testing multi-turn interactions and nested tool calling capabilities \citep{fang2025zero, zhu2025evolutionary, patil2025bfcl}. Advanced benchmarks including StableToolBench address API instability challenges, while NesTools evaluates nested tool scenarios and ToolHop assesses multi-hop tool usage across 995 queries and 3,912 tools \citep{guo2024stabletoolbench, han2024nestools, ye2025toolhop}.

Web agent evaluation frameworks including WebArena and Mind2Web provide comprehensive assessment across thousands of tasks spanning 137 websites, revealing significant performance gaps in current LLM capabilities for complex web interactions. VideoWebArena extends evaluation to multimodal agents, while Deep Research Bench and DeepShop address specialized evaluation for research and shopping agents respectively \citep{zhou2023webarena, deng2023towards, bosse2025deep, jang2024videowebarena}.

Multi-agent system evaluation employs specialized frameworks addressing coordination, communication, and collective intelligence. However, current frameworks face significant challenges in transactional integrity across complex workflows, with many systems lacking adequate compensation mechanisms for partial failures. Orchestration evaluation must address context management, coordination strategy effectiveness, and the ability to maintain system coherence under varying operational conditions \citep{chang2025sagallm, rizk2020unified}.

\begin{table*}[ht]
    \centering
    \scriptsize
    \renewcommand{\arraystretch}{1.2}
    \definecolor{lightgray}{RGB}{245,245,245}
     \begin{tabularx}{\textwidth}{p{2.5cm}|c|X|c|p{3cm}}
        \toprule
        \textbf{Release Date} & \textbf{Open Source} & \textbf{Method / Model} & \textbf{Success Rate (\%)} & \textbf{Source} \\
        \midrule
        \rowcolor{white}2025-02 & $\times$ & IBM CUGA & 61.7 & \citep{marreed2025towards} \\
        \rowcolor{lightgray}2025-01 & $\times$ & OpenAI Operator & 58.1 & \citep{openai2025cua} \\
        \rowcolor{white}2024-08 & $\times$ & Jace.AI & 57.1 & \citep{jace2024web} \\
        \rowcolor{lightgray}2024-12 & $\times$ & ScribeAgent + GPT-4o & 53.0 & \citep{shen2024scribeagent} \\
        \rowcolor{white}2025-01 & \checkmark & AgentSymbiotic & 52.1 & \citep{zhang2025symbiotic} \\
        \rowcolor{lightgray}2025-01 & \checkmark & Learn-by-Interact & 48.0 & \citep{su2025learn} \\
        \rowcolor{white}2024-10 & \checkmark & AgentOccam-Judge & 45.7 & \citep{yang2024agentoccam} \\
        \rowcolor{lightgray}2024-08 & $\times$ & WebPilot & 37.2 & \citep{zhang2024webpilot} \\
        \rowcolor{white}2024-10 & \checkmark & GUI-API Hybrid Agent & 35.8 & \citep{song2025browsing} \\
        \rowcolor{lightgray}2024-09 & \checkmark & Agent Workflow Memory & 35.5 & \citep{wang2024awm} \\
        \rowcolor{white}2024-04 & \checkmark & SteP & 33.5 & \citep{sodhi2024step} \\
        \rowcolor{lightgray}2025-06 & \checkmark & TTI & 26.1 & \citep{shen2025thinking} \\
        \rowcolor{white}2024-04 & \checkmark & BrowserGym + GPT-4 & 23.5 & \citep{workarena2024} \\
        \bottomrule
    \end{tabularx}
    \renewcommand{\arraystretch}{1.0}
    \caption{WebArena \citep{zhou2023webarena} Leaderboard: Top performing models with their success rates and availability status.}
    \label{tab:x_webarena_leaderboard}
\end{table*}

\subsection{Evaluation Challenges and Emerging Paradigms}
\label{subsec:evaluation_challenges}

This subsection identifies current limitations in evaluation methodologies and explores emerging approaches for more effective assessment.

\subsubsection{Methodological Limitations and Biases}
\label{subsubsec:methodological_limitations}

Traditional evaluation metrics prove fundamentally inadequate for capturing the nuanced, dynamic behaviors exhibited by context-engineered systems. Static metrics like BLEU, ROUGE, and perplexity, originally designed for simpler text generation tasks, fail to assess complex reasoning chains, multi-step interactions, and emergent system behaviors. The inherent complexity and interdependencies of multi-component systems create attribution challenges where isolating failures and identifying root causes becomes computationally and methodologically intractable. Future metrics must evolve to capture not just task success, but the quality and robustness of the underlying reasoning process, especially in scenarios requiring compositional generalization and creative problem-solving \citep{peng2024survey, wang2024what}.

Memory system evaluation faces particular challenges due to the lack of standardized benchmarks and the stateless nature of current LLMs. Automated memory testing frameworks must address the isolation problem where different memory testing stages cannot be effectively separated, leading to unreliable assessment results. Commercial AI assistants demonstrate significant performance degradation during sustained interactions, with accuracy drops of up to 30\% highlighting critical gaps in current evaluation methodologies and pointing to the need for longitudinal evaluation frameworks that track memory fidelity over time \citep{zhang2024memsim, xia2025minerva, huet2025episodic}.

Tool-integrated reasoning system evaluation reveals substantial performance gaps between current systems and human-level capabilities. The GAIA benchmark demonstrates that while humans achieve 92\% accuracy on general assistant tasks, advanced models like GPT-4 achieve only 15\% accuracy, indicating fundamental limitations in current evaluation frameworks and system capabilities \citep{mialon2023gaia, wang2024gta, chakraborty2025tool}. Evaluation frameworks must address the complexity of multi-tool coordination, error recovery, and adaptive tool selection across diverse operational contexts \citep{gao2025mcp, schick2023toolformer}.

\subsubsection{Emerging Evaluation Paradigms}
\label{subsubsec:emerging_paradigms}

Self-refinement evaluation paradigms leverage iterative improvement mechanisms to assess system capabilities across multiple refinement cycles. Frameworks including Self-Refine, Reflexion, and N-CRITICS demonstrate substantial performance improvements through multi-dimensional feedback and ensemble-based evaluation approaches. GPT-4 achieves approximately 20\% improvement through self-refinement processes, highlighting the importance of evaluating systems across multiple iteration cycles rather than single-shot assessments. However, a key future challenge lies in evaluating the meta-learning capability itself—not just whether the system improves, but how efficiently and robustly it learns to refine its strategies over time \citep{madaan2023self, shinn2023reflexion, mousavi2023critics, lee2024ask}.

Multi-aspect feedback evaluation incorporates diverse feedback dimensions including correctness, relevance, clarity, and robustness, providing comprehensive assessment of system outputs. Self-rewarding mechanisms enable autonomous evolution and meta-learning assessment, allowing systems to develop increasingly sophisticated evaluation criteria through iterative refinement \citep{lu2023self}.

Criticism-guided evaluation employs specialized critic models to provide detailed feedback on system outputs, enabling fine-grained assessment of reasoning quality, factual accuracy, and logical consistency. These approaches address the limitations of traditional metrics by providing contextual, content-aware evaluation that can adapt to diverse task requirements and output formats \citep{mousavi2023critics, lee2024ask}.

Orchestration evaluation frameworks address the unique challenges of multi-agent coordination by incorporating transactional integrity assessment, context management evaluation, and coordination strategy effectiveness measurement. Advanced frameworks including SagaLLM provide transaction support and independent validation procedures to address the limitations of systems that rely exclusively on LLM self-validation capabilities \citep{chang2025sagallm, he2025sentinelagent}.

\subsubsection{Safety and Robustness Assessment}
\label{subsubsec:safety_robustness}

Safety-oriented evaluation incorporates comprehensive robustness testing, adversarial attack resistance, and alignment assessment to ensure responsible development of context-engineered systems. Particular attention must be paid to the evaluation of agentic systems that can operate autonomously across extended periods, as these systems present unique safety challenges that traditional evaluation frameworks cannot adequately address \citep{singh2025agentic, guo2024lightrag}.

Robustness evaluation must assess system performance under distribution shifts, input perturbations, and adversarial conditions through comprehensive stress testing protocols. Multi-agent systems face additional challenges in coordination failure scenarios, where partial system failures can cascade through the entire agent network. Evaluation frameworks must address graceful degradation strategies, error recovery protocols, and the ability to maintain system functionality under adverse conditions. Beyond predefined failure modes, future evaluation must grapple with assessing resilience to ``unknown unknowns''—emergent and unpredictable failure cascades in highly complex, autonomous multi-agent systems \citep{chang2025sagallm, he2025sentinelagent}.

Alignment evaluation measures system adherence to intended behaviors, value consistency, and beneficial outcome optimization through specialized assessment frameworks. Context engineering systems present unique alignment challenges due to their dynamic adaptation capabilities and complex interaction patterns across multiple components. Long-term evaluation must assess whether systems maintain beneficial behaviors as they adapt and evolve through extended operational periods \citep{rizk2020unified}.

Looking ahead, the evaluation of context-engineered systems requires a paradigm shift from static benchmarks to dynamic, holistic assessments. Future frameworks must move beyond measuring task success to evaluating compositional generalization for novel problems and tracking long-term autonomy in interactive environments. The development of 'living' benchmarks that co-evolve with AI capabilities, alongside the integration of socio-technical and economic metrics, will be critical for ensuring these advanced systems are not only powerful but also reliable, efficient, and aligned with human values in real-world applications \citep{gao2025mcp, zhou2023webarena, zhang2024memsim}.

The evaluation landscape for context-engineered systems continues evolving rapidly as new architectures, capabilities, and applications emerge. Future evaluation paradigms must address increasing system complexity while providing reliable, comprehensive, and actionable insights for system improvement and deployment decisions. The integration of multiple evaluation approaches—from component-level assessment to system-wide robustness testing—represents a critical research priority for ensuring the reliable deployment of context-engineered systems in real-world applications \citep{peng2024survey, wang2024what}.

\section{Future Directions and Open Challenges}
\label{sec:future_directions}

Context Engineering stands at a critical inflection point where foundational advances converge with emerging application demands, creating unprecedented opportunities for innovation while revealing fundamental challenges that require sustained research efforts across multiple dimensions \citep{peng2024survey, wang2024what}. 

As the field transitions from isolated component development toward integrated system architectures, the complexity of research challenges grows exponentially, demanding interdisciplinary approaches that bridge theoretical computer science, practical system engineering, and domain-specific expertise \citep{gao2025mcp, schick2023toolformer}. 

This section systematically examines key research directions and open challenges that will define the evolution of Context Engineering over the coming decade.

\subsection{Foundational Research Challenges}
\label{subsec:foundational_challenges}

This subsection examines core theoretical and computational challenges that must be addressed to advance context engineering systems beyond current limitations.

\subsubsection{Theoretical Foundations and Unified Frameworks}
\label{subsubsec:theoretical_foundations}

Context Engineering currently operates without unified theoretical foundations that connect disparate techniques and provide principled design guidelines, representing a critical research gap that limits systematic progress and optimal system development. 

The absence of mathematical frameworks characterizing context engineering capabilities, limitations, and optimal design principles across different architectural configurations impedes both fundamental understanding and practical optimization \citep{wang2024what, lin2024training, peng2024survey, gao2025mcp}.

Information-theoretic analysis of context engineering systems requires comprehensive investigation into optimal context allocation strategies, information redundancy quantification, and fundamental compression limits within context windows. Current approaches lack principled methods for determining optimal context composition, leading to suboptimal resource utilization and performance degradation. Research must establish mathematical bounds on context efficiency, develop optimization algorithms for context selection, and create theoretical frameworks for predicting system behavior across varying context configurations \citep{ma2024megalodon, fu2025sliding}.

Compositional understanding of context engineering systems demands formal models describing how individual components interact, interfere, and synergize within integrated architectures. The emergence of complex behaviors from component interactions requires systematic investigation through both empirical studies and theoretical modeling approaches. Multi-agent orchestration presents particular challenges in developing mathematical frameworks for predicting coordination effectiveness and emergent collaborative behaviors \citep{chang2025sagallm, rizk2020unified}.

\subsubsection{Scaling Laws and Computational Efficiency}
\label{subsubsec:scaling_efficiency}

The fundamental asymmetry between LLMs' remarkable comprehension capabilities and their pronounced generation limitations represents one of the most critical challenges in Context Engineering research. 

This comprehension-generation gap manifests across multiple dimensions including long-form output coherence, factual consistency maintenance, and planning sophistication, requiring investigation into whether limitations stem from architectural constraints, training methodologies, or fundamental computational boundaries \citep{peng2024survey, wang2024what}.

Long-form generation capabilities demand systematic investigation into planning mechanisms that can maintain coherence across thousands of tokens while preserving factual accuracy and logical consistency. Current systems exhibit significant performance degradation in extended generation tasks, highlighting the need for architectural innovations beyond traditional transformer paradigms. State space models including Mamba demonstrate potential for more efficient long sequence processing through linear scaling properties, though current implementations require substantial development to match transformer performance across diverse tasks \citep{ma2024megalodon, ye2025longmamba, gu2022parameterization, ding2023longnet}.

Context scaling efficiency faces fundamental computational challenges, with current attention mechanisms scaling quadratically (O(n²)) with sequence length, creating prohibitive memory and computational requirements for ultra-long sequences. Sliding attention mechanisms and memory-efficient implementations represent promising directions, though significant research is needed to address both computational tractability and reasoning quality preservation \citep{fu2025sliding, yang2025lserve, gu2022parameterization}. Position interpolation and extension techniques require advancement to handle sequences exceeding current architectural limitations while maintaining positional understanding and coherence.

\subsubsection{Multi-Modal Integration and Representation}
\label{subsubsec:multimodal_integration}

The integration of diverse modalities within context engineering systems presents fundamental challenges in representation learning, cross-modal reasoning, and unified architectural design. Current approaches typically employ modality-specific encoders with limited cross-modal interaction, failing to capture the rich interdependencies that characterize sophisticated multi-modal understanding. VideoWebArena demonstrates the complexity of multimodal agent evaluation, revealing substantial performance gaps in current systems when processing video, audio, and text simultaneously \citep{jang2024videowebarena}.

Beyond these sensory modalities, context engineering must also handle more abstract forms of information such as graphs, whose structural semantics are not directly interpretable by language models. Capturing the high-level meaning encoded in graph structures introduces unique challenges, including aligning graph representations with language model embeddings and expressing graph topology efficiently. Recent efforts like GraphGPT~\citep{tang2024graphgpt} and GraphRAG~\citep{edge2024local} attempt to bridge this gap through cross-modal alignment strategies, while others explore converting graphs into natural language descriptions to facilitate model understanding \citep{fatemi2023talk, ge2024can}. \citet{bi2024lpnl} further propose a divide-and-conquer approach to encode text-attributed heterogeneous networks, addressing context length limitations and enabling effective link prediction. Graph reasoning thus emerges as a core difficulty in context engineering, requiring models to navigate complex relational structures beyond raw modalities.

Temporal reasoning across multi-modal contexts requires sophisticated architectures capable of tracking object persistence, causal relationships, and temporal dynamics across extended sequences. Web agent frameworks including WebArena showcase the challenges of maintaining coherent understanding across complex multi-step interactions involving diverse modalities and dynamic content. Current systems demonstrate significant limitations in coordinating multi-modal information processing with action planning and execution \citep{zhou2023webarena, deng2023towards}.

Cross-modal alignment and consistency present ongoing challenges in ensuring that information extracted from different modalities remains factually consistent and semantically coherent. Deep Research Bench evaluation reveals that current multi-modal agents struggle with complex research tasks requiring synthesis across textual, visual, and structured data sources, highlighting the need for more sophisticated alignment mechanisms \citep{bosse2025deep}.

\subsection{Technical Innovation Opportunities}
\label{subsec:technical_opportunities}

This subsection explores emerging technical approaches and architectural innovations that promise to enhance context engineering capabilities.

\subsubsection{Next-Generation Architectures}
\label{subsubsec:nextgen_architectures}

Architectural innovations beyond traditional transformer paradigms offer promising directions for addressing current limitations in context engineering systems. State space models including LongMamba demonstrate potential for more efficient long sequence processing through linear scaling properties and improved memory utilization, though current implementations require substantial development to match transformer performance across diverse tasks \citep{ye2025longmamba, ma2024megalodon}. Specialized position encoding methods and parameter-efficient architectures present opportunities for scaling to ultra-long sequences while maintaining computational tractability \citep{gu2022parameterization, fu2025sliding}.

Memory-augmented architectures require advancement beyond current external memory mechanisms to enable more sophisticated long-term memory organization, hierarchical memory structures, and adaptive memory management strategies. MemoryBank implementations incorporating Ebbinghaus Forgetting Curve principles demonstrate promising approaches to memory persistence, though significant research is needed to address the fundamental stateless nature of current LLMs \citep{zhong2023memorybank, zhang2024memsim, xia2025minerva, packer2023memgpt, xu2025mem}. The development of episodic memory systems capable of maintaining coherent long-term context across extended interactions represents a critical architectural challenge \citep{huet2025episodic, pink2024assessing, he2024camelot}.

Modular and compositional architectures enable flexible system construction through specialized component integration while maintaining overall system coherence. Modular RAG architectures demonstrate enhanced flexibility through specialized modules for retrieval, augmentation, and generation, enabling fine-grained optimization of individual components. Graph-enhanced approaches including GraphRAG and LightRAG showcase the potential for integrating structured knowledge representation with neural processing \citep{gao2024modular, singh2025agentic, guo2024lightrag}.

\subsubsection{Advanced Reasoning and Planning}
\label{subsubsec:advanced_reasoning}

Context engineering systems require enhanced reasoning capabilities spanning causal reasoning, counterfactual thinking, temporal reasoning, and analogical reasoning across extended contexts. Current systems demonstrate limited capacity for sophisticated reasoning patterns that require integration of multiple evidence sources, consideration of alternative scenarios, and maintenance of logical consistency across complex inference chains \citep{wang2024what, peng2024survey}.

Multi-step planning and execution capabilities represent critical advancement areas enabling systems to decompose complex tasks, formulate execution strategies, monitor progress, and adapt plans based on intermediate results. Agentic RAG systems demonstrate sophisticated planning and reflection mechanisms requiring integration of task decomposition, multi-plan selection, and iterative refinement capabilities. However, current implementations face significant challenges in maintaining coherence across extended planning horizons and adapting to dynamic information conditions \citep{huang2025survey, cheng2025survey, xiong2025rag}.

Tool-integrated reasoning represents a paradigmatic advancement requiring dynamic interaction with external resources during reasoning processes. The GAIA benchmark demonstrates substantial performance gaps, with human achievement of 92\% accuracy compared to advanced models achieving only 15\%, highlighting fundamental limitations in current reasoning and planning capabilities \citep{mialon2023gaia, wang2024gta, chakraborty2025tool}. Advanced tool integration must address autonomous tool selection, parameter extraction, multi-tool coordination, and error recovery across diverse operational contexts \citep{gao2025mcp, schick2023toolformer}.

\subsubsection{Complex Context Organization and Solving Graph Problems}
\label{subsubsec:complex_context_graph}

Graph reasoning represents a fundamental challenge in context engineering, requiring systems to navigate complex structural relationships while maintaining semantic understanding across interconnected elements. Recent advances in graph-language model integration demonstrate multiple paradigms: specialized architectural approaches that incorporate graph-specific components and text-based encoding strategies that transform graph structures into natural language representations \citep{wang2023can, tang2023graphgpt}.

Architectural integration approaches include GraphGPT, which employs dual-stage instruction tuning aligning graph structural information with language tokens via self-supervised graph matching \citep{tang2023graphgpt, mao2024advancing}. This framework introduces specialized GraphTokens refined through Graph Instruction Tuning and utilizes a lightweight graph-text alignment projector for transitioning between textual and structural processing modalities \citep{you2024large, feng2025hypergraph}. Building upon instruction-tuning paradigms, GraphWiz extends this approach by incorporating DPO to enhance reasoning reliability, achieving 65\% average accuracy across diverse graph tasks and significantly outperforming GPT-4's 43.8\% \citep{chen2024graphwiz}. Chain-of-thought distillation mechanisms enhance step-by-step reasoning performance \citep{wei2022chain, zhu2025graph}. RL presents another promising direction, as demonstrated by G1, which trains LLMs on synthetic graph-theoretic tasks using the Erdős dataset comprising 50 diverse tasks, achieving strong zero-shot generalization with a 3B parameter model outperforming significantly larger models \citep{guo2025g1}.

Text-based encoding approaches transform graph structures into natural language descriptions using few-shot prompting and chain-of-thought reasoning without architectural modifications \citep{fatemi2023talk, dai2024how}. These methods introduce diverse graph description templates contextualizing structural elements through multiple semantic interpretations \citep{shang2024survey, luo2024enhance}. Recent work investigates the impact of graph description ordering on LLM performance, revealing that sequential presentation significantly influences model comprehension and reasoning accuracy \citep{ge2024can}. Benchmark evaluations have expanded with GraphArena, offering both polynomial-time tasks and NP-complete challenges with a rigorous evaluation framework that classifies outputs as correct, suboptimal, hallucinatory, or missing \citep{tang2024grapharena}. Combined with existing benchmarks like NLGraph and GraphDO, these evaluations reveal substantial performance disparities between simple connectivity problems and complex tasks like maximum flow computation \citep{wang2023can, rizvi2022fimp, ge2024can}.

Current implementations face challenges in scaling to large structures, maintaining consistency across multi-hop relationships, and generalizing to novel topologies, with text-based approaches offering interpretability at reduced structural precision while specialized architectures provide enhanced performance through increased complexity \citep{ren2024survey, wang2024graphtool}. Emerging hybrid approaches including InstructGraph and GraphAdapter attempt to bridge these paradigms through structured format verbalizers and GNN-based adapters, though limitations persist in handling dynamic structures and temporal evolution of relationships \citep{fang2025graphgpt}. Looking forward, broad connection paradigms that organize information through associative networks rather than fragmented searches, spreading outward from central nodes to discover potential connections between entities, may represent the next generation of RAG systems for complex context organization \citep{chegn2022thinking}.

\subsubsection{Intelligent Context Assembly and Optimization}
\label{subsubsec:intelligent_context}

Automated context engineering systems capable of intelligently assembling contexts from available components represent a critical research frontier requiring development of context optimization algorithms, adaptive selection strategies, and learned assembly functions. Current approaches rely heavily on heuristic methods and domain-specific engineering, limiting scalability and optimality across diverse applications \citep{wang2024what, lin2024training}.

Self-refinement mechanisms demonstrate substantial potential for intelligent context optimization through iterative improvement processes. Self-Refine, Reflexion, and N-CRITICS frameworks achieve significant performance improvements, with GPT-4 demonstrating approximately 20\% improvement through iterative refinement. However, these approaches require advancement in optimization strategies for autonomous evolution and meta-learning across diverse contexts \citep{madaan2023self, shinn2023reflexion, mousavi2023critics, lee2024ask}.

Multi-dimensional feedback mechanisms incorporating diverse feedback dimensions including correctness, relevance, clarity, and robustness provide promising directions for context optimization. Self-rewarding mechanisms enable autonomous evolution capabilities, though research must address fundamental questions about optimal adaptation rates, stability-plasticity trade-offs, and preservation of beneficial adaptations across varying operational conditions \citep{lu2023self}.

\subsection{Application-Driven Research Directions}
\label{subsec:application_directions}

This subsection addresses research challenges arising from real-world deployment requirements and domain-specific applications.

\subsubsection{Domain Specialization and Adaptation}
\label{subsubsec:domain_specialization}

Context engineering systems require sophisticated specialization mechanisms for diverse domains including healthcare, legal analysis, scientific research, education, and engineering applications, each presenting unique requirements for knowledge integration, reasoning patterns, safety considerations, and regulatory compliance. Domain-specific optimization demands investigation into transfer learning strategies, domain adaptation techniques, and specialized training paradigms that preserve general capabilities while enhancing domain-specific performance \citep{wang2024what, lin2024training}.

Scientific research applications require sophisticated reasoning capabilities over complex technical content, mathematical expressions, experimental data, and theoretical frameworks while maintaining rigorous accuracy standards. Deep Research Bench evaluation reveals significant challenges in current systems' ability to conduct complex research tasks requiring synthesis across multiple information sources and reasoning over technical content. Research must address integration of symbolic reasoning with neural approaches and incorporation of domain-specific knowledge bases \citep{bosse2025deep}.

Healthcare applications demand comprehensive safety evaluation frameworks, regulatory compliance mechanisms, privacy protection protocols, and integration with existing clinical workflows while maintaining interpretability and auditability requirements. Medical context engineering must address challenges in handling sensitive information, ensuring clinical accuracy, supporting diagnostic reasoning, and maintaining patient privacy across complex healthcare ecosystems. Current evaluation frameworks reveal substantial gaps in medical reasoning capabilities and safety assessment methodologies \citep{he2024madial}.

\subsubsection{Large-Scale Multi-Agent Coordination}
\label{subsubsec:largescale_multiagent}

Scaling multi-agent context engineering systems to hundreds or thousands of participating agents requires development of distributed coordination mechanisms, efficient communication protocols, and hierarchical management structures that maintain system coherence while enabling local autonomy. Research must address fundamental challenges in distributed consensus, fault tolerance, and emergent behavior prediction in large-scale agent populations \citep{du2025survey, chen2025from}.

Communication protocol standardization represents a critical research frontier, with emerging protocols including MCP (``USB-C for AI''), A2A (Agent-to-Agent), ACP (Agent Communication Protocol), and ANP (Agent Network Protocol) demonstrating the need for unified frameworks enabling interoperability across diverse agent ecosystems. However, current implementations face security vulnerabilities and scalability limitations that must be addressed for large-scale deployment \citep{anthropic:mcp, google:a2a, ibm:acp, anp:comm, ehtesham2025survey, sarkar2025survey, li2025from}.

Orchestration challenges including transactional integrity, context management, and coordination strategy effectiveness represent significant obstacles to large-scale multi-agent deployment. Contemporary frameworks including LangGraph, AutoGen, and CAMEL demonstrate insufficient transaction support and validation limitations, requiring systems that rely exclusively on LLM self-validation capabilities. Advanced coordination frameworks must address compensation mechanisms for partial failures and maintain system coherence under varying operational conditions \citep{chang2025sagallm, he2025sentinelagent, rizk2020unified}.

\subsubsection{Human-AI Collaboration and Integration}
\label{subsubsec:human_ai_collaboration}

Sophisticated human-AI collaboration frameworks require deep understanding of human cognitive processes, communication preferences, trust dynamics, and collaboration patterns to enable effective hybrid teams that leverage complementary strengths. Research must investigate optimal task allocation strategies, communication protocols, and shared mental model development between humans and AI systems \citep{wang2024what, peng2024survey}.

Web agent evaluation frameworks reveal significant challenges in human-AI collaboration, particularly in complex task scenarios requiring sustained interaction and coordination. WebArena and Mind2Web demonstrate that current systems struggle with multi-step interactions across diverse websites, highlighting fundamental gaps in collaborative task execution. Advanced interfaces require investigation into context-aware adaptation and personalization mechanisms that enhance human-AI team performance \citep{zhou2023webarena, deng2023towards}.

Trust calibration and transparency mechanisms represent critical research areas for ensuring appropriate human reliance on AI systems while maintaining human agency and decision-making authority. Evaluation frameworks must address explanation generation, uncertainty communication, and confidence calibration to support informed human decision-making in collaborative scenarios. The substantial performance gaps revealed by benchmarks like GAIA underscore the importance of developing systems that can effectively communicate their limitations and capabilities \citep{mialon2023gaia, wang2024gta}.

\subsection{Deployment and Societal Impact Considerations}
\label{subsec:deployment_societal}

This subsection examines critical considerations for deploying context engineering systems at scale while ensuring responsible and beneficial outcomes.

\subsubsection{Scalability and Production Deployment}
\label{subsubsec:scalability_deployment}

Production deployment of context engineering systems requires addressing scalability challenges across multiple dimensions including computational resource management, latency optimization, throughput maximization, and cost efficiency while maintaining consistent performance across diverse operational conditions. The O(n²) scaling limitation of current attention mechanisms creates substantial barriers to deploying ultra-long context systems in production environments, necessitating advancement in memory-efficient architectures and sliding attention mechanisms \citep{fu2025sliding, yang2025lserve}.

Reliability and fault tolerance mechanisms become critical as context engineering systems assume increasingly important roles in decision-making processes across domains. Multi-agent orchestration frameworks face particular challenges in maintaining transactional integrity across complex workflows, with current systems lacking adequate compensation mechanisms for partial failures. Research must address graceful degradation strategies, error recovery protocols, and redundancy mechanisms that maintain system functionality under adverse conditions \citep{chang2025sagallm, he2025sentinelagent}.

Maintainability and evolution challenges require investigation into system versioning, backward compatibility, continuous integration protocols, and automated testing frameworks that support ongoing system improvement without disrupting deployed services. Memory system implementations face additional challenges due to the stateless nature of current LLMs and the lack of standardized benchmarks for long-term memory persistence and retrieval efficiency \citep{zhang2024memsim, xia2025minerva}.

\subsubsection{Safety, Security, and Robustness}
\label{subsubsec:safety_security}

Comprehensive safety evaluation requires development of assessment frameworks that can identify potential failure modes, safety violations, and unintended behaviors across the full spectrum of context engineering system capabilities. Agentic systems present particular safety challenges due to their autonomous operation capabilities and complex interaction patterns across extended operational periods \citep{singh2025agentic, guo2024lightrag}.

Security considerations encompass protection against adversarial attacks, data poisoning, prompt injection, model extraction, and privacy violations while maintaining system functionality and usability. Multi-agent communication protocols including MCP, A2A, and ACP introduce security vulnerabilities that must be addressed while preserving interoperability and functionality. Research must develop defense mechanisms and detection systems that address evolving threat landscapes across distributed agent networks \citep{ehtesham2025survey, sarkar2025survey}.

Alignment and value specification challenges require investigation into methods for ensuring context engineering systems behave according to intended objectives while avoiding specification gaming, reward hacking, and goal misalignment. Context engineering systems present unique alignment challenges due to their dynamic adaptation capabilities and complex interaction patterns across multiple components. The substantial performance gaps revealed by evaluation frameworks underscore the importance of developing robust alignment mechanisms that can maintain beneficial behaviors as systems evolve and adapt \citep{mialon2023gaia, chang2025sagallm}.

\subsubsection{Ethical Considerations and Responsible Development}
\label{subsubsec:ethical_responsible}

Bias mitigation and fairness evaluation require comprehensive assessment frameworks that can identify and address systematic biases across different demographic groups, application domains, and use cases while maintaining system performance and utility. Research must investigate bias sources in training data, model architectures, and deployment contexts while developing mitigation strategies that address root causes rather than symptoms \citep{wang2024what, peng2024survey}.

Privacy protection mechanisms must address challenges in handling sensitive information, preventing data leakage, and maintaining user privacy while enabling beneficial system capabilities. Memory systems face particular privacy challenges due to their persistent information storage and retrieval capabilities, requiring advanced frameworks for secure memory management and selective forgetting mechanisms \citep{zhang2024memsim, huet2025episodic}.

Transparency and accountability frameworks require development of explanation systems, audit mechanisms, and governance structures that enable responsible oversight of context engineering systems while supporting innovation and beneficial applications. The substantial performance gaps revealed by evaluation frameworks like GAIA (human 92\% vs AI 15\%) highlight the importance of transparent capability communication and appropriate expectation setting for deployed systems \citep{mialon2023gaia, wang2024gta}.

The future of Context Engineering will be shaped by our ability to address these interconnected challenges through sustained, collaborative research efforts that bridge technical innovation with societal considerations. 

Success will require continued investment in fundamental research, interdisciplinary collaboration, and responsible development practices that ensure context engineering systems remain beneficial, reliable, and aligned with human values as they become increasingly integrated into critical societal functions \citep{peng2024survey, wang2024what, gao2025mcp}.
\section{Conclusion}
\label{sec:conclusion}

This survey has presented the first comprehensive examination of Context Engineering as a formal discipline that systematically designs, optimizes, and manages information payloads for LLMs. Through our analysis of over 1400 research papers, we have established Context Engineering as a critical foundation for developing sophisticated AI systems that effectively integrate external knowledge, maintain persistent memory, and interact dynamically with complex environments.

Our primary contribution lies in introducing a unified taxonomic framework that organizes context engineering techniques into \textbf{Foundational Components} (Context Retrieval and Generation, Context Processing, and Context Management) and \textbf{System Implementations} (Retrieval-Augmented Generation, Memory Systems, Tool-Integrated Reasoning, and Multi-Agent Systems). This framework demonstrates how core technical capabilities integrate into sophisticated architectures addressing real-world requirements.

Through this systematic examination, we have identified several key insights. First, we observe a fundamental asymmetry between LLMs' remarkable capabilities in understanding complex contexts and their limitations in generating equally sophisticated outputs. This comprehension-generation gap represents one of the most critical challenges facing the field. Second, our analysis reveals increasingly sophisticated integration patterns where multiple techniques combine synergistically, creating capabilities that exceed their individual components. Third, we observe a clear trend toward modularity and compositionality, enabling flexible architectures adaptable to diverse applications while maintaining system coherence. The evaluation challenges we identified underscore the need for comprehensive assessment frameworks that capture the complex, dynamic behaviors exhibited by context-engineered systems. Traditional evaluation methodologies prove insufficient for systems that integrate multiple components, exhibit adaptive behaviors, and operate across extended time horizons. Our examination of future research directions reveals significant opportunities including developing next-generation architectures for efficient long context handling, creating intelligent context assembly systems, and advancing multi-agent coordination mechanisms. Key challenges span theoretical foundations, technical implementation, and practical deployment, including the lack of unified theoretical frameworks, scaling limitations, and safety considerations.

Looking toward the future, Context Engineering stands poised to play an increasingly central role in AI development as the field moves toward complex, multi-component systems. The interdisciplinary nature of Context Engineering necessitates collaborative research approaches spanning computer science, cognitive science, linguistics, and domain-specific expertise.

As LLMs continue to evolve, the fundamental insight underlying Context Engineering—that AI system performance is fundamentally determined by contextual information—will remain central to artificial intelligence development. This survey provides both a comprehensive snapshot of the current state and a roadmap for future research, establishing Context Engineering as a distinct discipline with its own principles, methodologies, and challenges to foster innovation and support responsible development of context-aware AI systems.

\section*{Acknowledgments}

This survey represents an ongoing effort to comprehensively map the rapidly evolving landscape of Context Engineering for Large Language Models. Given the dynamic nature of this field, with new developments emerging continuously, we acknowledge that despite our best efforts, some recent works or emerging trends may have been inadvertently overlooked or underrepresented. We welcome feedback from the research community to help improve future iterations of this work. We are grateful to the broader research community whose foundational contributions have made this survey possible. This work would not have been achievable without the invaluable support of both the research community and the open-source community, whose collaborative efforts in developing frameworks, tools, and resources have significantly advanced the field of Context Engineering. We extend special gratitude to the teams behind the Long Chain-of-Thought \citep{chen2025reasoning} and AI4Research \citep{chen2025ai4researchsurveyartificialintelligence} projects for their excellent template designs and visualizations, which have significantly enhanced the presentation quality of this survey. Their thoughtful contributions to the research community are deeply appreciated.

\nocite{*}
\bibliographystyle{refstyle}
\bibliography{ref}

\end{document}